\documentclass[11pt,letterpaper]{article}
\usepackage{graphicx}
\usepackage{amsmath,amssymb,amsthm}
\usepackage[margin=1in]{geometry}

\usepackage[algo2e,lined,vlined,boxed]{algorithm2e}
\usepackage{algorithm,algorithmic}
\usepackage{epstopdf}
\usepackage{subfigure}
\usepackage{hyperref}
\usepackage{color}
\usepackage{bbm}
\usepackage{multirow}
\usepackage{bibentry}
\nobibliography*

\newtheorem{definition}{Definition}
\newtheorem{lemma}{Lemma}
\newtheorem{corollary}{Corollary}
\newtheorem{theorem}{Theorem}
\newtheorem{assumption}{Assumption}

\newtheorem*{lemma*}{Lemma}

\def\real{\mathbb{R}}

\def\prob{\mathbb{P}}

\def\lambdabar{\overline{\lambda}}
\def\fhat{\hat{f}}
\def\Sigmahat{\hat{\Sigma}}

\def\P{\mathcal{P}}
\def\X{\mathcal{X}}
\def\Y{\mathcal{Y}}
\def\L{\mathcal{L}}
\def\H{\mathcal{H}}
\def\fbar{\bar{f}}

\def\Err{\text{Err}}
\def\Reg{\text{Reg}}
\def\Bias{\text{Bias}}
\def\Var{\text{Var}}
\def\CovErr{\text{CovErr}}
\def\App{\text{Approx}}
\def\poly{\text{poly}}
\def\AppErr{\text{ApproxError}}

\newcommand{\E}[1]{\mathbb{E}\left[ #1 \right]}
\newcommand{\Ed}[2]{\mathbb{E}_{#1}\left[ #2 \right]}
\newcommand{\Eemp}[1]{\widehat{\mathbb{E}}\left[ #1 \right]}
\newcommand{\indic}[1]{\mathbbm{1}\left( #1 \right)}
\newcommand{\inpdt}[2]{\left\langle #1, #2 \right\rangle}
\newcommand{\norm}[2]{\left\lVert #1 \right\rVert_{#2}}

\newcommand{\Tr}[1]{\text{Tr}\left(#1\right)}

\DeclareMathOperator*{\argmin}{arg\,min}

\title{Kernel Ridge Regression via Partitioning}

\author{Rashish Tandon\\ \texttt{rashish@cs.utexas.edu} \and Si Si\\ \texttt{ssi@cs.utexas.edu} \and Pradeep Ravikumar\\
\texttt{pradeepr@cs.utexas.edu} \and Inderjit Dhillon\\
\texttt{inderjit@cs.utexas.edu} \vspace*{0.2in}\\ Department of Computer Science\\ The University of Texas at Austin\\ Austin, TX 78712 USA}

\date{\today}

\begin{document}
\maketitle
\begin{abstract}
In this paper, we investigate a divide and conquer approach to Kernel Ridge Regression (KRR). Given $n$ samples, the division step involves separating the points based on some underlying disjoint partition of the input space (possibly via clustering), and then computing a KRR estimate for each partition. The conquering step is simple: for each partition, we only consider its own local estimate for prediction. We establish conditions under which we can give generalization bounds for this estimator, as well as achieve optimal minimax rates. We also show that the \emph{approximation error} component of the generalization error is lesser than when a single KRR estimate is fit on the data: thus providing both statistical and computational advantages over a single KRR estimate over the entire data (or an averaging over random partitions as in other recent work, \cite{zhang13}). Lastly, we provide experimental validation for our proposed estimator and our assumptions.
\end{abstract}

\section{Introduction}
Kernel methods find wide and varied applicability in machine learning. Kernelization of supervised/unsupervised learning algorithms allows an easy extension to operate them on implicitly infinite/high-dimensional feature representations. The use of kernel feature maps can also convert non-linearly separable data to be separable in the new feature space, thus resulting in good predictive performance. One such application of kernels is the problem of Kernel Ridge Regression (shortened as KRR). Given covariate-response pairs $(x,y)$, the goal is to compute a kernel-based function $f$ such that $f(x)$ approximates $y$ well on average. In this regard, several learning methods with different kernel classes have been shown to achieve good predictive performance. Despite their good generalization, kernel methods suffer from a computational drawback if the number of samples $n$ is large --- which is more so the case in modern settings. They require at least a computational cost of $O(n^2)$, which is the time required to compute the kernel matrix, and $O(n^3)$ time when the kernel matrix also has to be inverted, which is the case for KRR.\\

Several approaches have been proposed to mitigate this, including Nystr\"{o}m approximations \cite{bach13, mahoney15, rudi2015less}, approximations via random features \cite{rahimi07, rahimi08, song14, ian14}, and others \cite{raskutti14, wainwright15}. While these approaches help computationally, they typically incur an error over-and-above the error incurred by a KRR estimate on the entire data. Another class of approaches that may not incur such an error are based on what we loosely characterize as divide-and-conquer approaches, wherein the data points are \textit{divided} into smaller sets, and estimators trained on the divisions. These approaches may further be categorized into three main classes: division by uniform splitting \cite{zhang13}, division by clustering \cite{GuH13, ssi14} or division by partitioning \cite{eberts15}. The latter may also include local learning approaches, which are based on estimates using training points near a test point \cite{vapnik92,zhang06, segata10, hable13}. Given this considerable line of work, there is now an understanding that these divide-and-conquer approaches provide computational benefits, and yet have statistical performance that is either asymptotically equivalent, or at most slightly worse than that of the whole KRR estimator. Please see \cite{zhang13, ssi14, eberts15} and references therein for results reflecting this understanding for uniform splitting, clustering and partitioning respectively. However, these results have restrictive assumptions, applicability or other limitations, such as requiring the covariates/responses to be bounded \cite{eberts15}, or only being applicable to specific kernels e.g. Gaussian \cite{eberts15} or linear \cite{GuH13}, or only being targeted to classification \cite{GuH13, ssi14}, or providing error rates only on the training error \cite{ssi14}. Moreover, approaches based on uniform splitting, such as \cite{zhang13}, can suffer from worse \textit{approximation} error, as alluded to shortly.\\

In this paper, we consider a partitioning based divide-and-conquer approach to kernel ridge regression. We provide a refined analysis, applicable to general kernels, which leads us to this surprising conclusion: the partitioning based approach not only has computational benefits outlined in previous papers, but also has strong statistical benefits when compared to the whole KRR estimator. In other words, based on both a statistical and computational viewpoint, we are able to recommend the use of the partitioning based approach over the whole KRR approach.\\

The partitioning based approach is: Given $n$ sample points, we divide them into $m$ groups based on a fixed disjoint partitioning of input space $\X$ that the samples are drawn from. One way to obtain this partition is via clustering, however, in principle, any partition that satisfies certain assumptions (detailed in Section \ref{subsec:asmp}) would be acceptable. A primary intuition for considering partitioning is that the distribution within each partition may be localized with thin tails. Equivalently, the eigenspectrum of the covariance conditioned on the partition may decay sharply enough such that simply focusing on the local samples suffices to obtain a good approximation. This intuition is captured in our assumptions. So, once the samples have been divided, we learn a kernel ridge regression estimate for each partition using only its own samples. The conquering step \emph{i.e.} computing the overall estimator, $\fhat_C$, is then simple: Each individual estimator is applied to its respective partition. Thus, to perform prediction for a new point, we simply identify its partition, and use the estimator for that partition. Now, partitioning has a clear computational advantage since each estimate is trained over only a fraction of the points. Moreover, partitioning may provide statistical advantages as well if there is an inherent approximation error in the problem \emph{i.e.}, the true regressor function, $f^*$, lies outside the space of kernel-based functions. In this case, the KRR estimator on the whole data, say $\fhat_{whole}$, or the KRR estimator based on uniform splitting, say $\fhat_{avg}$, both may be viewed as estimating the best single kernel-based function that approximates $f^*$. However, if we partition, then we are estimating the best $m$-piece-wise kernel-based function to approximate $f^*$. Indeed, we can show that the approximation error for $\fhat_C$ is lesser than $\fhat_{avg}$, and corroborate this experimentally. The residual error terms on the other hand are typically of the same order, so that the overall generalization error for our method is lower. In addition, there is yet another potential computational advantage of partitioning: prediction is faster since for a new point, the kernel values must be computed w.r.t. only a fraction of the points (as opposed to all the points for $\fhat_{whole}$ or $\fhat_{avg}$).

\subsection{Related Work}\label{subsec:relatwork}
We briefly review some of the earlier mentioned work that provide theoretical analyses of divide and conquer approaches, based on clustering, uniform splitting, and partitioning. \cite{GuH13} have applied clustering to linear SVMs (instead of KRR, as in this work) with an additional global penalty to prevent over fitting, and derived simple generalization rates based on rademacher complexity estimates. \cite{ssi14} consider clustering for Kernel SVMs with a modified conquering step -- solutions of the local SVM problems are combined to produce an initialization for a solver of the global SVM problem. Under this scheme, they analyze the so-called fixed design setting \textit{i.e.} they bound the error on the training data as a function of the block diagonal approximation of the kernel. In contrast, in this work we consider the random design setting \textit{i.e.} we bound the generalization/prediction error, and also for the slightly different (but related) problem of Kernel Ridge Regression. Perhaps, the approaches most closely related to our work are \cite{zhang13, eberts15}. \cite{zhang13} analyze the uniform splitting approach where the samples are split uniformly at random, followed by an averaging of the KRR estimate of each split. The authors have derived generalization rates for this estimator, and matched optimal rates as long as the number of splits is not \textit{large}, and the true function $f^*$ lies in the specified space of kernel-based functions. However, as mentioned previously, such an estimator can have worse approximation error than our estimator, $\fhat_C$, when the true function, $f^*$, lies outside the space of kernel-based functions. \cite{eberts15} analyze a partition based approach as in our paper: their estimator works by partitioning the input space, and predicting using KRR/SVM estimates over each partition individually. For this estimator, \cite{eberts15} derive generalization rates when using Gaussian kernels, and under additional restrictions: they require bounded covariates, $\norm{x}{}\leq B$, bounded response, $\lvert y\rvert \leq M$, and that each partition be bounded by a ball of suitable radius, $R$.\footnote{One way of obtaining such partitions, as suggested by the authors, is through the Voronoi partitioning of the input space.} Given these restrictions, they show suitable choices for $R$ and the Gaussian kernel scale $\gamma$ which yield optimal rates when the true function $f^*$ lies in a smooth Sobolev/Besov space. In contrast, we provide a more general analysis that does not enforce a bound on the covariates, response, or the size of the partition. Moreover, we are able to apply it to kernels other than the Gaussian kernel, and achieve minimax optimal rates when the true function, $f^*$, lies in the space of kernel-based functions. When it doesn't, we provide an oracle inequality similar to \cite{eberts15}, which could then be specialized to obtain similar rates for their specific setting. More importantly, our analysis is also able to show that in general, the approximation component of this inequality is lesser than the approximation component of the whole KRR estimator, while the residual components can be of the same order.\\

Another line of work in the same spirit as this work is that of local learning approaches, for e.g. the early work of \cite{vapnik92}. The main idea here is to select training samples \textit{near} a given test sample, and only use those to train an SVM for the particular test sample's prediction. Several variants under this general scheme have been proposed \cite{zhang06, segata10, hable13}. However, since each test sample requires finding \textit{nearby} training points and solving its own SVM, these approaches can be inefficient for prediction.\\

From a theoretical standpoint, the generalization error for KRR has been studied extensively --- an incomplete list includes \cite{smale02, zhang05, zhou07, caponnetto07, steinwart09, mendelson2010, hsu12, eberts13}. While we shall not delve into the differences among the results derived in these articles, we refer the interested reader to \cite[Section 2.5]{hsu12}, \cite[Section 3]{eberts13}, and the references therein, for a more detailed comparison. Of particular relevance to our analysis is the approach in \cite{hsu12}, wherein the generalization error is broken down into contributions of \textit{regularization}, \textit{bias due to random design} and \textit{variance due to noise}, and consequently each of these is controlled separately w.h.p.. Moreover, the analysis of the related approach in \cite{zhang13} may be viewed as a moment version of the same strategy. We adopt a similar strategy to control the expected error of our estimator, $\fhat_C$. \\

The rest of this paper is organized as follows. Section \ref{sec:prelim} describes the KRR problem, and sets up some notation and mathematical prerequisites. Section \ref{sec:dcest} details the $DC$-estimator $\fhat_C$, our partitioning based estimator. Section \ref{sec:generr} presents the bounds on the generalization error of $\fhat_C$, and the assumptions required to achieve them. Section \ref{sec:kernel_specific} instantiates these bounds for three specific and commonly studied kernel classes. Finally, Section \ref{sec:experiments} provides empirical performance results. All proofs are provided in the appendix.

\section{Preliminaries and Problem Setup}\label{sec:prelim}
\textbf{Reproducing Kernel Hilbert Spaces}. 
Consider any set $\X$. In machine learning applications, $\X$ is typically the space of the input data. A function $K : \X\times \X \to \real$, is called a \textit{kernel function} if it is \textit{continuous, symmetric}, and \textit{positive definite}. With any kernel function $K$, one can associate a unique Hilbert space called the \textit{Reproducing Kernel Hilbert Space} of $K$ (abbreviated as RKHS henceforth). For $x\in \X$, let $\phi_x:\X\to\real$ be the function $\phi_x(\cdot):= K(x,\cdot)$. Then, the unique RKHS corresponding to kernel $K$, denoted as $\H$, is a Hilbert space of functions from $\X$ to $\real$ defined as: 
\begin{equation}
\H := \overline{\text{span}}\{\phi_x \}
\end{equation}
Thus, any $f\in \H$ has the representation $f = \sum_j \alpha_j \phi_{x_j} = \sum_j \alpha_j K(x_j,\cdot)$ with $\alpha_j \in \real, \,\forall\, j$. The inner product on $\H$ is given as: $\inpdt{\sum_{j}\alpha_j \phi_{x_j}}{\sum_k \beta_k \phi_{x_k}}_{\H} = \sum_{j}\sum_{k} \alpha_j \beta_k K(x_j, x_k)$. The inner product also induces a norm on $\H$, given as: $\norm{f}{\H} = \sqrt{\inpdt{f}{f}_{\H}}$, for any $f\in \H$.\\

\textbf{Kernel Ridge Regression}.
We are given a training set of $n$ samples, $\mathbf{D}=\left\{(x_1,y_1), \ldots, (x_n,y_n) \right\}$, of the tuple $(x,y)$ drawn i.i.d. from an unknown distribution $\overline{\P}$ on $\X\times\Y$. $x$ (and $x_i$) is a random variable in the input space $\X$, also called the covariate. $y$ (and $y_i$) is a random variable in the output space $\Y$, also called the response. We consider $\Y\subseteq \real$ and assume an additive noise model for relating the response to the covariate:
  \begin{equation}
    y = f^*(x) + \eta,
  \end{equation}
where $\eta$ is the random noise variable and $f^* :\X \to \real$ is an unknown mapping of covariates in $\X$ to responses in $\real$. The goal of regression is to compute the function (or an approximation to) $f^*$. We also assume that the noise has zero mean and bounded variance, $\E{\eta \vert x} = 0$ and $\E{\eta^2\vert x} \leq \sigma^2$, and that  $f^*$ is square integreable with respect to the measure on $\X$. Equivalently, this means that $f^*\in \L_2(\X,\P) := \{f:\X \to \real \,\vert\, \norm{f}{L_2}^2 = \mathbb{E}_{\P}\left[f(x)^2\right] <\infty \}$, where  $\P$ is the marginal of $\overline{\P}$ on the input space $\X$.  \\

A Kernel Ridge Regression (KRR) estimator approximates $f^*$ by a function in the RKHS space $\H$ (corresponding to kernel $K$). We require that the RKHS space $\H\subset L_2(\X,\P)$ --- which means $\forall x,\, \mathbb{E}_{y\sim \P}[K(x,y)^2]<\infty$ --- which is always true for several kernel classes, including Gaussian, Laplacian, or any trace class kernel w.r.t. $\P$. The KRR estimate $\fhat_{\lambda}\in \H$ is obtained by solving the following optimization problem:
\begin{equation}\label{eq:fhatopt}
 \fhat_{\lambda} = \argmin_{f\in \H} \frac{1}{n}\sum_{i=1}^{n} (y_i - f(x_i))^2 + \lambda \norm{f}{\H}^2
\end{equation}
where $\lambda > 0$ is the regularization penalty. This is tractable since, by the representer theorem, we have the relation $\fhat_{\lambda} = \sum_{i=1}^{n} \alpha_i \phi_{x_i}$, with  $\alpha\in\real^n$ being the solution of the following problem:
\begin{equation}\label{eq:alphaopt}
 \min_{\alpha\in \real^n} \frac{1}{n}\sum_{i=1}^n (y_i - (G\alpha)_i)^2 + \lambda (\alpha^T G\alpha)
\end{equation}
where $G\in\real^{n\times n}$ is the kernel matrix, with $G_{ij} = K(x_i, x_j)$ $\left(i, j\in [n]\right)$. Eq. \eqref{eq:alphaopt} has a closed form solution, given as: $\alpha = (G + n\lambda I)^{-1}y$.\\

\textbf{Generalization/Prediction Error}. For any estimator $\fhat:\X \to \real$, the generalization error provides a metric of closeness to $f^*$, by measuring the average squared error in prediction using $\fhat$. It is defined as:
\begin{align}
 \Err(\fhat) &:= \E{(\fhat(x) - f^*(x))^2} = \norm{\fhat - f^*}{L_2}^2
\end{align}
By quantifying $\Err(\fhat)$ to be small, we know that $\fhat$ is a good approximation to $f^*$. When the estimator is random, for example the KRR estimate $\fhat_{\lambda}$ in Eq. \eqref{eq:fhatopt} depends on random samples, we may quantify the average error over the randomness \emph{i.e.} bound $\mathbb{E}_{D}[\Err(\fhat_{\lambda})]$, where the expectation is taken over the samples $\mathbf{D}$.\\

In this paper, we provide bounds on the quantity $\mathbb{E}_{D}[\Err(\fhat_C)]$, where $\fhat_C$ is the \textit{Divide-and-Conquer} estimator ($DC$-estimator) described in Section \ref{sec:dcest}.\\

\textbf{Partition-specific notation}.
Since our estimator, $\fhat_C$, is based on partitioning, we setup some notation here for partition-specific quantities that play a role throughout the analysis. We say that the input space $\X$ has a disjoint partition $\{C_1, \ldots, C_m\}$ if: 
\begin{equation}
\X = \cup_{i=1}^{m} C_i, \text{ and } C_i\cap C_j = \{\phi\} \,\forall\, i, j\in [m], i\ne j
\end{equation}
Given data $\mathbf{D}=\left\{(x_1,y_1), \ldots, (x_n,y_n) \right\}$, we define a partition-based empirical covariance operator as: 
\begin{equation}\label{eq:covpartdef}
\Sigmahat_i = \frac{1}{n}\sum_{j=1}^{n} (\phi_{x_j}\otimes \phi_{x_j})\indic{x_j\in C_i}
\end{equation}
where $\indic{\cdot}$ denotes the indicator function and $\phi_x\otimes \phi_x$ denotes the operator $\phi_x \inpdt{\phi_x}{\cdot}_{\H}$. We define its population counterpart as: 
\begin{equation}\label{eq:covpopdef}
\Sigma_i = \E{(\phi_x\otimes\phi_x) \indic{x\in C_i}}
\end{equation}
Note the relation: $\Sigma = \sum_{i=1}^{m} \Sigma_i$, where $\Sigma = \E{\phi_x\otimes \phi_x}$ is the overall covariance operator.\\

We let $\{\lambda_j^i, v_j^i\}_{j=1}^{\infty}$ denote the collection of eigenvalue-eigenfunction pairs for $\Sigma_i$. For any $\lambda >0$, we define a spectral sum for $\Sigma_i$: 
\begin{equation}
S_i(\lambda) = \sum_{j} \frac{\lambda_j^i}{\lambda_j^i + \lambda}
\end{equation}
Similarly, letting $\{\lambda_j, v_j\}_{j=1}^{\infty}$ be the eigenvalue-eigenfunction pairs for the overall covariance $\Sigma$, the corresponding sum for $\Sigma$ is defined as:
\begin{equation}
S(\lambda) = \sum_{j} \frac{\lambda_j}{\lambda_j + \lambda}
\end{equation}
The quantity $S(\lambda)$ has appeared in previous work on KRR \cite{ zhang05,hsu12,zhang13}, and is called the \emph{effective dimensionality} of the kernel $K$ (at scale $\lambda$). Typically, it plays the same role as dimension does in finite dimensional ridge regression. We shall refer to the quantity $S_i(\lambda)$ as the \emph{effective dimensionality} of partition $C_i$. Finally, we let $p_i = \prob(x \in C_i)$ denote the probability mass of partition $C_i$.

\section{The $DC$-estimator: $\fhat_C$}
\label{sec:dcest}
When the number of samples $n$ is large, solving Eq. \eqref{eq:fhatopt} (through Eq. \eqref{eq:alphaopt}) may be computationally prohibitive, requiring $O(n^3)$ time in the worst case. A simple strategy to tackle this is by \textit{dividing} the samples $\mathbf{D}$ into disjoint partitions, and computing an estimate separately for each partition. In this work, we consider partitions of $\mathbf{D}$ which adhere to an underlying disjoint partition of the input space $\X$. Suppose that the input space $\X$ has a disjoint partition $\{C_1, \ldots, C_m\}$. Note that $m$ denotes the number of partitions. Also, suppose that given any point $x\in \X$, we can find the partition it belongs to from the set $\{ C_1, \ldots, C_m \}$. Note that $m$ denotes the number of partitions. Also, suppose that given any point $x\in \X$, we can find the partition it belongs to from the set $\{ C_1, \ldots, C_m \}$.\\

Now, we divide the data set $\mathbf{D}$ in agreement with this partitioning of $\X$ \emph{i.e.} we split $\mathbf{D} = \{D_1, \ldots, D_m\}$ with $D_i = \{(x_j, y_j)\,\vert\, x_j\in C_i,\, j=1,\ldots, n\}$. Let $\lvert D_i\rvert = n_i$. Then, for any partition $i\in [m]$, we compute a local estimator using only the points in its partition:
\begin{align}\label{eq:partestopt}
 \fhat_{i,\lambda} &= \argmin_{f\in \H}\, \frac{1}{n_i}\sum_{j:\,(x_j, y_j) \in D_i} (y_j - f(x_j))^2 + \lambda \norm{f}{\H}^2
\end{align}
where $\lambda > 0$ is the regularization penalty. Finally, the overall estimator, $\fhat_C$, comprises of the local estimators applied to their corresponding partitions:
\begin{align}\label{eq:clustestimator}
 \fhat_C(x) = \fhat_{i,\lambda}(x) \text{ if } x\in C_i
\end{align}

In practice, one can use a clustering algorithm to cluster the points in $\mathbf{D}$, as well as determine membership for new points $x$.
\section{Generalization Error of $\fhat_C$}
\label{sec:generr}
In this section we quantify the error $\mathbb{E}_D\left[\Err(\fhat_C)\right]$, where $\fhat_C$ is the $DC$-estimator from Eq. \ref{eq:clustestimator}. The analysis follows an integral-operator approach which has been frequently employed in deriving such bounds in learning theory, for e.g. in \cite{zhou07, hsu12}.\\

First, we observe that $\Err(\fhat_C)$ can be decomposed as a sum of errors of the local estimators, $\fhat_{i,\lambda}$, on their corresponding partitions $C_i$, $i\in [m]$. We have:
\begin{align}\label{eq:decomp1}
 \Err(\fhat_C) = \E{(f^*(x) - \fhat_C(x))^2} &= \sum_{i=1}^{m} \E{(f^*(x) - \fhat_{C}(x))^2 \indic{x \in C_i}}\nonumber\\
 &= \sum_{i=1}^{m} \E{(f^*(x) - \fhat_{i,\lambda}(x))^2 \indic{x \in C_i}}\nonumber\\
 &= \sum_{i=1}^m \Err_i(\fhat_{i,\lambda})
\end{align}
where $\indic{\cdot}$ denotes the indicator function, and we have defined the partition-wise error:
\begin{equation}
\Err_i(\fhat_{i,\lambda}) := \E{(f^*(x) - \fhat_{i,\lambda}(x))^2 \indic{x \in C_i}}
\end{equation}
By linearity of expectation, $\mathbb{E}_D\left[\Err(\fhat)\right] = \sum_{i=1}^m \mathbb{E}_D\left[\Err_i(\fhat_{i,\lambda})\right]$. Therefore, to obtain a bound on $\mathbb{E}_{D}\left[\Err(\fhat)\right]$ we need to bound $\mathbb{E}_{D}\left[\Err_i(\fhat_{i,\lambda})\right]$, for every $i\in [m]$.\\

Now, our strategy to control $\Ed{D}{\Err_i(\fhat_{i,\lambda})}$ is to bound it as a sum of intermediate error terms\footnote{Similar to the usual bias-variance decomposition; or the decomposition in \cite{hsu12, zhang13}. In contrast, loosely speaking, \cite{eberts15} analyze the error of $\fhat_C$ by viewing it as a Standard KRR with a new kernel $K1(x,x') = \sum_{i=1}^{m} K(x,x')\indic{x \in C_i}  \indic{x' \in C_i}$}, and in turn provide bounds for these intermediate error terms. For this purpose, we define the following estimates (for each $i\in [m]$):
\begin{align}
\text{For any }\lambdabar\geq 0,\quad f_{i, \overline{\lambda}} &=  \argmin_{f\in \H} \,\E{(y - f(x))^2\,\vert\, x\in C_i} + \overline{\lambda} \norm{f}{\H}^2\label{eq:fbestopt1}\\
 f_{i, \lambda} &=  \argmin_{f\in \H} \,\E{(y - f(x))^2\,\vert\, x\in C_i} + \lambda \norm{f}{\H}^2\label{eq:fbestopt2}\\
 \fbar_{i,\lambda} &= \mathbb{E}_{D}[{\fhat_{i,\lambda}}]\label{eq:fbestopt3}
\end{align}
$f_{i,\lambdabar}$ and $f_{i,\lambda}$ are the optimal \textit{population} KRR estimates for partition $C_i$, with regularization penalties $\lambdabar$ and $\lambda$ respectively. $\fbar_{i,\lambda}$ is the expected value of the \textit{empirical} KRR estimate from Eq. \eqref{eq:partestopt}, with the expectation taken over the samples $\mathbf{D}$. Note that there is no source of randomness in all of the above quantities, whereas $\fhat_{i,\lambda}$ is a random quantity due to its dependence on the random samples $\mathbf{D}$. Now, based on the above estimates, we define the following error terms:

\begin{definition}\label{def:errterms}
For any $\lambda >0$ and $\lambdabar\in [0,\lambda]$, we define
 \begin{align}
  \text{{\bf Approximation Error }}:\;\App_i(\lambdabar) &= \E{(f^*(x) - f_{i,\lambdabar}(x))^2\indic{x\in C_i}}\label{eq:defapp}\\
  \text{{\bf Regularization Error }}:\;\Reg_i(\lambdabar, \lambda) &= \E{(f_{i,\lambdabar}(x) - f_{i,\lambda}(x))^2\indic{x\in C_i}}\label{eq:defreg}\\
  \text{{\bf Bias  }}:\;\Bias_i(\lambda, n) &= \E{(f_{i,\lambda}(x) - \fbar_{i,\lambda}(x))^2\indic{x\in C_i}}\label{eq:defbias}\\
  \text{{\bf Variance }}:\;\Var_i(\lambda, D) &= \E{(\fbar_{i,\lambda}(x) - \fhat_{i,\lambda}(x))^2\indic{x\in C_i}}\label{eq:defvar}
 \end{align}
\end{definition}

The intent of $f_{i,\lambdabar}$, in the above definition, is to correspond to the \textit{best} kernel function that approximates $f^*$ in the partition $C_i$. The choice of $\lambdabar$, that determines $f_{i,\lambdabar}$, can be viewed as a \textit{small} regularization penalty that trades-off the approximation error, $\App_i(\lambdabar)$, to $\norm{f_{i,\lambdabar}}{\H}$ (which influences the remaining terms in Definition \ref{def:errterms}).  Ideally, if the unknown regression function $f^*$ lies in the RKHS space $\H$, then a choice of $\lambdabar = 0$ would suffice. In that case, we would have $f_{i,\lambdabar} = f_{i,0} = f^*$ --- which would imply zero approximation error \emph{i.e.} $\App_i(\lambdabar) = \App_i(0) = 0$, while $\norm{f_{i,\lambdabar}}{\H} (= \norm{f^*}{\H})$ would be bounded.\\

Now, the following lemma describes the decomposition of $\Ed{D}{\Err_i(\fhat_{i,\lambda})}$ in terms of the quantities in Definition \ref{def:errterms}.
\begin{lemma}[Error Decomposition]\label{lem:errdecomp}
For each partition $i\in [m]$, the error $\Ed{D}{\Err_{i}(f_{i,\lambda})}$ decomposes as (for any $\lambdabar\in[0,\lambda]$):
\begin{align}\label{eq:partdecomp}
 \Ed{D}{\Err_i(\fhat_{i,\lambda})} \leq 2\left[ \App_{i}(\lambdabar) + 2 \Reg_i(\lambdabar,\lambda) + 2 Bias_i(\lambda, n) + 2 \Ed{D}{\Var_{i}(\lambda,D)} \right]
 \end{align}
 Thus, the overall error $\Ed{D}{\Err(\fhat_C)}$ can be decomposed as (for any $\lambdabar\in[0,\lambda]$):
 \begin{align}\label{eq:wholedecomp}
 \Ed{D}{\Err(\fhat_C)} \leq 2\left[\sum_{i=1}^{m}\App_{i}(\lambdabar) + 2 \sum_{i=1}^{m}\Reg_i(\lambdabar,\lambda) + 2 \sum_{i=1}^{m} Bias_i(\lambda, n) + 2 \sum_{i=1}^{m} \Ed{D}{\Var_{i}(\lambda,D)}\right]
 \end{align}
\end{lemma}

We note that while in the above decomposition we have considered the same choice of regularization penalty, $\lambda$ (and $\lambdabar$), for all partitions $i \in [m]$, a similar decomposition would hold even if we were to choose a different $\lambda$ (and $\lambdabar$) for each partition.\\

To summarize, in Lemma \ref{lem:errdecomp}, we have decomposed the overall error of our estimator, $\Ed{D}{\Err(\fhat_C)}$, as a sum of individual errors for each partition. Furthermore, the individual errors for each partition have been decomposed into four components: \textit{Approximation, Regularization, Bias} and \textit{Variance}. The rest of this section is devoted to bounding these terms for any partition. First, however, we require certain assumptions on the partitions. These are detailed in Section \ref{subsec:asmp}. Additionally, we need a supporting bound that controls the operator norm of the sample covariance error of each partition, under a suitable \textit{whitening}. This is provided in Section \ref{subsec:covbnd}. Finally, Section \ref{sec:mainbnds} presents the bounds on the component terms for each partition.
\subsection{Assumptions}\label{subsec:asmp}
In this section, we describe three assumptions needed to bound the terms in Lemma \ref{lem:errdecomp}. It may be useful at this point to recall partition-specific definitions from Section \ref{sec:prelim}. We also remark that two of these assumptions are fairly standard (Assumption \ref{asmp:eigfn} and Assumption \ref{asmp:approx}), and analogous versions have appeared in earlier work \cite{hsu12, zhang13, eberts15}. The last assumption, Assumption \ref{asmp:goodpart}, is novel. However, we have validated it extensively on both real and synthetic data sets (see Section \ref{sec:experiments}).\\

Now, our first assumption concerns the existence of higher-order moments of the eigenfunctions, $v_j^i$.
\begin{assumption}[Eigenfunction moments]\label{asmp:eigfn}
Let $\{\lambda_j^i, v_j^i\}_{j=1}^{\infty}$ denote the eigenvalue-eigenfunction pairs for the covariance operator $\Sigma_i$. Then, $\forall i\in [m]$, $\forall j$ s.t. $\lambda_j^i\ne 0$, and for some constant $k\geq 2$, 
\begin{align}
\E{\left(\frac{v_j^i(x)^2\indic{x\in C_i}}{\lambda_j^i}\right)^{2k}}\leq a_1^k
\end{align}
where $a_1$ is a constant.
\end{assumption}
Note that we always have: $\E{\frac{v_j^i(x)^2\indic{x\in C_i}}{\lambda_j^i}} = \E{\frac{\inpdt{v_j^i}{\phi_x}_{\H}^2\indic{x\in C_i}}{\lambda_j^i}} = \frac{\inpdt{v_j^i}{\Sigma_i v_j^i}}{\lambda_j^i} = 1$. Thus, the first moment of $(v_j^i(x)^2\indic{x\in C_i}/\lambda_j^i)$ always exists. Assumption \ref{asmp:eigfn} requires \emph{sufficiently} many higher moments to exist. This assumption can also be interpreted as requiring partition-wise sub-Gaussian behaviour (up to $2k$ moments) in the RKHS space, given its primary application to the bounds (see Section \ref{subsec:asmpeigfn_interpret}, in the Appendix, for more details). Finally, we note that this assumption is similar to \cite[Assumption A]{zhang13}, but applied to each partition.\\

Our next assumption concerns the approximation variable $(f^*(x) - f_{i,\overline{\lambda}}(x))$, requiring its fourth moment to be bounded.
\begin{assumption}[Finite Approximation]\label{asmp:approx}
$\forall i\in [m]$, and any $\lambdabar\geq 0$, there exists a constant $A_{i}(\lambdabar)\geq 0$ such that
\begin{align}
\E{(f^*(x) - f_{i,\overline{\lambda}}(x))^4\,\vert\, x\in C_i}\leq A_{i}(\lambdabar)^4
\end{align}
where $f_{i,\lambdabar}$ is the solution of the optimization problem in Eq. \ref{eq:fbestopt1}.
\end{assumption}

We remark that while in the above we have specified $A_i(\lambdabar)$ to be a constant, a slow growing function of $\lambdabar$ would also work in our bounds. Also, while this assumption is stated for any $\lambdabar$, we really only care about the actual $\lambdabar$ used in Eq. \ref{eq:partdecomp}. Thus, for example, if $f^*\in \H$, then as noted earlier, a choice of $\lambdabar=0$ suffices --- consequently Assumption \ref{asmp:approx} trivially holds with $A_{i}(\lambdabar)=0$ at $\lambdabar=0$, since $f_{i,\lambdabar} = f^*$ for $\lambdabar = 0$.\\

Our final assumption enforces that the sum of \emph{effective dimensionality} over all the partitions be bounded in terms of the overall \emph{effective dimensionality}. For this purpose, we define the \textbf{goodness measure} of a partition $\{C_1, \ldots, C_m\}$ as:
\begin{align}\label{eq:goodnessdef}
g(\lambda):= \frac{\sum_{i=1}^{m} S_i(\lambda p_i)}{S(\lambda)}
\end{align}
Now, we have the following assumption.
\begin{assumption}[Goodness of Partition]
\label{asmp:goodpart}
Let $\lambda > 0$ be the regularization penalty in Eq. \eqref{eq:partestopt} for any $i\in [m]$. Then, we require:  $g(\lambda) = O(1)$.
\end{assumption}
In Section \ref{sec:kernel_specific}, we show that if we have $g(\lambda) = O(1)$ for a $\lambda$ decaying \textit{suitably} in terms of $n$, the $DC$-estimator can achieve optimal minimax rates. 
In other words, if the partitioning preserves the overall \textit{effective dimensionality}, then there is no loss in the generalization error. We validate the above assumption (at \textit{suitable} $\lambda$) by estimating $g(\lambda)$ on real and synthetic data sets (see Section \ref{sec:experiments}). From a practitioner's perspective, $g(\lambda)$ may be viewed as a surrogate for the \textit{suitability} of a partition for the DC-estimator, and can help guide the choice of partition.
\subsection{Covariance Control}\label{subsec:covbnd}
A key component to establish bounds on the terms in Lemma \ref{lem:errdecomp} involves controlling the moments of the operator norm of the sample covariance error, under a suitable \textit{whitening}. Specifically, we need a bound on the quantity (for any $i\in [m]$ and some $k\geq 2$):
\begin{equation}\label{eq:coverrterm}
\E{\norm{\Sigma_{i,\lambda p_i}^{-1/2}(\Sigmahat_i - \Sigma_i)\Sigma_{i,\lambda p_i}^{-1/2}}{}^k}^{1/k}:= \CovErr_{i}(\lambda p_i, n, k)
\end{equation}
where we use the shorthand: $\Sigma_{i,\lambda p_i} = (\Sigma_i + \lambda p_i I)$. $\Sigmahat_i$ and $\Sigma_i$ are partition-wise empirical and population covariance operators respectively, as defined in Eqs. \eqref{eq:covpartdef} and \eqref{eq:covpopdef}. The above term (in Eq. \eqref{eq:coverrterm}) appears throughout in the bounds for $\Bias_i(\lambda, n)$ and $\Var_i(\lambda, D)$, and a general bound on this quantity can be found in Lemma \ref{lem:expcovbnd} in the Appendix. While the expression in Lemma \ref{lem:expcovbnd} is complicated, it can be specialized for specific kernels to obtain meaningful expressions. We state these expressions for three cases below. Their derivation can be found in Section \ref{subsubsec:coverrcases} in the Appendix.\\

\textbf{Finite Rank Kernels}. Suppose kernel $K$ has finite rank $r$ --- examples include the linear and polynomial kernels. Then, for any $i\in[m]$ and $k>2$, we get:
\begin{align}
 \CovErr_{i}(\lambda p_i, n, k) &= O\left(\frac{\sqrt{\log r}\, S_{i}(\lambda p_i)}{\sqrt{n}}\right) = O\left(\frac{r \sqrt{\log r}}{\sqrt{n}}\right)
\end{align}

\textbf{Kernels with polynomial decay in eigenvalues}. Suppose kernel $K$ has polynomially decaying eigenvalues, $\lambda_j \leq c j^{-v}\; (\forall j$, and constants $c>0, v>2$) --- examples here include sobolev kernels with different orders. Then, for any $i\in [m]$, $k>2$, and $\lambda p_i \geq \frac{1}{n^{\alpha}}$ for some constant $\alpha < \frac{v}{2}-1$, we get:
\begin{align}
 \CovErr_{i}(\lambda p_i, n, k) &= O\left(\frac{\sqrt{\log n}}{n^{\frac{1}{2} - \frac{\alpha+1}{v}}}\right)
\end{align}

\textbf{Kernels with exponential decay in eigenvalues}. Suppose kernel $K$ has exponentially decaying eigenvalues, $\lambda_j \leq c_1 \exp(- c_2 j^2)\; (\forall j$, and constants $c_1, c_2 >0$) --- an example here is the Gaussian kernel. Then, for any $i\in[m]$, $k>2$ and $\lambda p_i \geq \poly(1/n)$, we get:
\begin{align}
 \CovErr_{i}(\lambda p_i, n, k) &= O\left(\frac{\sqrt{\log n\, (\log \log n)}}{\sqrt{n}}\right)
\end{align}

Overall, it would be useful to think of $\CovErr_{i}( \lambda p_i, n, k)$ to be scaling as $\widetilde{O}\left(n^{-1/2}\right)$. Consequently, in the bounds to follow, there are terms of the form $\CovErr_{i}(\lambda p_i, n, k)^{k}$ --- which scale as $\widetilde{O}(n^{-k/2})$, and become negligible for a sufficiently large $k$.
\subsection{Bounds on $\Reg_i$, $\Bias_i$ and $\Var_i$}\label{sec:mainbnds}
We are now ready to provide bounds on the terms involved in Lemma \ref{lem:errdecomp}. The following lemmas provide bounds on the Regularization error, Bias and Variance, for any partition $i\in[m]$, as given in Definition \ref{def:errterms}. We only state the lemmas here using the $O(\cdot)$ notation. Precise statements can be found in the appendix. Recall that $p_i = \prob(x\in C_i)$ and $f_{i,\lambdabar}$ be the solution of Eq. \ref{eq:fbestopt1}. Additionally, we use the shorthand $CE_i$ to denote $\CovErr_{i}(\lambda p_i, n, k)$.

\begin{lemma}[Regularization Loss]\label{lem:mainregbnd}
Consider any partition $i\in [m]$. Then,
 \begin{align}\label{eq:regi}
  \Reg_i(\lambda,\overline{\lambda})= O\left( p_i \frac{(\overline{\lambda}-\lambda)^2}{\lambda}  \norm{f_{i,\overline{\lambda}}}{\H}^2\right)
 \end{align}
\end{lemma}

\begin{lemma}[Bias Loss]\label{lem:mainbiasbnd}
Let $k \geq 2$ such that Assumption \ref{asmp:eigfn} holds for this $k$ (with constant $a_1$), and Assumption \ref{asmp:approx} holds (with $A_i(\lambdabar)\geq 0$). Also, suppose $p_i$ satisfies (for any $i\in [m]$): $p_i = \Omega \left(\log n/ n\right)$. Then, for any $i \in [m]$,
\begin{align}\label{eq:biasi}
\Bias_i(\lambda,n) \leq O\left((CE_i)^2\left(  T_1  + T_2  + \left(CE_i\right)^k T_3 + \left(CE_i\right)^{k/2} T_4\right) \right)
\end{align}
where we let
\begin{align}
T_1 = \frac{\sqrt{p_i}  S_{i}(\lambda p_i) A_{i}(\lambdabar)^2}{n}, &\qquad
T_2 = \frac{(\lambda - \lambdabar)^2}{\lambda}\, \frac{ p_i S_i(\lambda p_i)^2 \norm{f_{i,\lambdabar}}{\H}^2}{n} + \frac{\lambda p_i  \norm{f_{i,\lambdabar}}{\H}^2}{n} \nonumber\\
T_3 = \norm{f_{i,\lambdabar}}{\H}^2 + \frac{\sigma^2}{\lambda}, &\qquad
T_4 = \frac{A_i(\lambdabar)^2}{\lambda \sqrt{p_i}}
\end{align}
\end{lemma}

\begin{lemma}[Variance Loss]\label{lem:mainvarbnd}
Let $k \geq 2$ such that Assumption \ref{asmp:eigfn} holds for this $k$ (with constant $a_1$), and Assumption \ref{asmp:approx} holds (with $A_i(\lambdabar)\geq 0$). Also, suppose $p_i$ satisfies (for any $i\in [m]$): $p_i =\Omega\left(\log n/ n\right)$. Then, for any $i \in [m]$,
\begin{align}\label{eq:vari}
\Ed{D}{\Var_i(\lambda,D)}&\leq O\left(W_1 + W_2 +  \left(CE_i\right)^k W_3 + \left(CE_i\right)^{k/2} W_4\right)
\end{align}
where we let
\begin{align}
W_1 = \frac{(\sigma^2 + \sqrt{p_i}A_i(\lambdabar)^2) S_i(\lambda p_i)}{n}, &\qquad
W_2 = \frac{(\lambdabar - \lambda)^2 p_i}{\lambda}\norm{f_{i,\lambdabar}}{\H}^2\nonumber\\
W_3 = \norm{f_{i,\overline{\lambda}}}{\H}^2 + \frac{\sigma^2}{\lambda}, &\qquad
W_4 = \frac{A_{i}(\lambdabar)^2}{\lambda \sqrt{p_i}}
\end{align}
\end{lemma}

Note that Lemma \ref{lem:mainbiasbnd} and Lemma \ref{lem:mainvarbnd} have a minimum requirement on $p_i$, namely: $p_i = \Omega\left(\log n/ n\right)$. However, this is minor since this essentially corresponds to each partition having $\Omega(\log n)$ samples. We also remark that this requirement can be avoided under certain restrictions for e.g. if the unknown regression function $f^*$ is uniformly bounded \emph{i.e.} $\lvert f^*(x)\rvert\leq M\, \forall\, x$. Now, to interpret the above bounds, recall from Section \ref{subsec:covbnd} that $CE_i = \CovErr_{i}(d, \lambda p_i, n)$ can scale as $\widetilde{O}\left(n^{-1/2}\right)$. Therefore, terms of the form $\CovErr_{i}(d, \lambda p_i, n)^{k}$ --- which scale as $\widetilde{O}(n^{-k/2})$ --- will be of lower order for a large enough $k$. Also note that the overall bias term gets multiplied with an $\widetilde{O}(n^{-1})$ factor. Indeed, in most cases, the bias term (Eq \eqref{eq:biasi}) turns out to be of a much lower order than the variance term (Eq \eqref{eq:vari}). Moreover, the first two terms in the variance bound (Eq \eqref{eq:vari}), and the bound for $\Reg_i$ (Eq \eqref{eq:regi}), become the overall dominating terms. Consequently, using Lemma \ref{lem:errdecomp}, we have an overall scaling of: $\Ed{D}{\Err_i(\fhat_{i,\lambda})} \approx O\left(\App_i(\lambdabar) + \frac{(\lambdabar - \lambda)^2 p_i}{\lambda}\norm{f_{i,\lambdabar}}{\H}^2 + \frac{\sigma^2 S_i(\lambda p_i)}{n} \right)$.
\section{Bounds under Specific Cases}
\label{sec:kernel_specific}
In this section, using the bounds on \textit{regularization error}, \textit{bias} and \textit{variance} from Section \ref{sec:mainbnds}, we instantiate the overall error bounds for the kernel classes discussed in Section \ref{subsec:covbnd}. We do this under the assumption that $f^*\in \H$. When $f^*\notin \H$, we provide an oracle inequality for the error term and contrast this with a similar inequality derived in \cite{zhang13}. Throughout this section, we assume that the conditions of Lemma \ref{lem:mainbiasbnd} and Lemma \ref{lem:mainvarbnd} are satisfied.

\subsection{$f^*\in \H$ --- Zero approximation error}
As mentioned earlier, in this case a choice of $\lambdabar = 0$ suffices. With $\lambdabar = 0$, we have $f_{i,\lambdabar} = f^*$ (from Eq. \eqref{eq:fbestopt1}). Thus, $\App_i(\lambdabar)=0$ at $\lambdabar=0$. Also, Assumption \ref{asmp:approx} trivially holds with $A_i(\lambdabar) = 0$ at $\lambdabar=0$.

\begin{theorem}[Finite Rank Kernels]\label{thm:finrank}
Let $f^*\in \H$ and suppose kernel $K$ has a finite rank $r$. Let $m$ denote the number of partitions, and let $k\geq 2$ such that Assumption \ref{asmp:eigfn} holds for this $k$. Then, the overall error for the $DC$-estimator $\fhat_C$ is given as:
\begin{align}
\Ed{D}{\Err(\fhat_C)}& = O\bigg(\lambda  \norm{f^*}{\H}^2 + \frac{\sigma^2}{n}g(\lambda) S(\lambda) +  m\left(\frac{r^2 \log r}{n}\right)^{k/2}\left(\norm{f^*}{\H}^2 + \frac{\sigma^2}{\lambda}\right)\bigg)
\end{align}
Now, if $m = O\left(\sqrt{\frac{n^{(k-4)}}{(r^2 \log r)^{k}}}\right)$ and Assumption \ref{asmp:goodpart} holds at $\lambda = r/n$, then the $DC$-estimator $\fhat_C$ achieves the optimal rate: $\Ed{D}{\Err(\fhat_C)} = O\left(\frac{r}{n}\right)$ at $\lambda = r/n$.
\end{theorem}

Note that the requirement of $m = O\left(\sqrt{\frac{n^{(k-4)}}{(r^2 \log r)^{k}}}\right)$ in the above theorem is only meaningful for $k\geq 4$ \emph{i.e.} we require at least 4 moments of the quantity in Assumption \ref{asmp:eigfn} to exist. If this is true, and if Assumption \ref{asmp:goodpart} holds, Theorem \ref{thm:finrank} gives the rate $\Ed{D}{\Err(\fhat_C)} = O\left(\frac{r}{n}\right)$, which is known to be minimax optimal \cite{zhang13, raskutti12}.

\begin{theorem}[Kernels with polynomial eigenvalue decay]\label{thm:polydec}
Let $f^*\in \H$ and suppose kernel $K$ has polynomially decaying eigenvalues : $\lambda_j \leq c j^{-v}\; (\forall j$, and constants $c>0, v>2$). Let $m$ denote the number of partitions, and let $k\geq 2$ such that Assumption \ref{asmp:eigfn} holds for this $k$. Also, suppose $\lambda p_i \geq \frac{1}{n^{\alpha}}$ for some constant $0 < \alpha < \frac{v}{2} - 1$, and $\forall i\in[m]$. Then, the overall error for the $DC$-estimator $\fhat_C$ is given as:
\begin{align}
\Ed{D}{\Err(\fhat_C)}& = O\bigg(\lambda  \norm{f^*}{\H}^2 + \frac{\sigma^2}{n}g(\lambda) S(\lambda) + m\left(\frac{\log n}{n^{1 - \frac{2(\alpha + 1)}{v}}}\right)^{k/2}\left(\norm{f^*}{\H}^2 + \frac{\sigma^2}{\lambda}\right)\bigg)
\end{align}
Now, if $m = O\left(\sqrt{\frac{n^{k - \frac{2k(\alpha + 1)}{v} - \frac{4v}{v+1}}}{(\log n)^k}}\right)$ and Assumption \ref{asmp:goodpart} holds at $\lambda = 1/n^{\frac{v}{v+1}}$ and $p_i \geq \frac{1}{n^{\alpha - \frac{v}{v+1}}}$ ($\forall i\in [m]$), then the $DC$-estimator $\fhat_C$ achieves the optimal rate: $\Ed{D}{\Err(\fhat_C)} = O\left(\frac{1}{n^{\frac{v}{v+1}}}\right)$ at $\lambda = 1/n^{\frac{v}{v+1}}$.
\end{theorem}

Note that the requirement of $p_i \geq \frac{1}{n^{\alpha - \frac{v}{v+1}}}$ in the latter part of the above theorem implicitly entails: $\alpha > \frac{v}{v+1}$. This, when coupled with the requirement $\alpha < \frac{v}{2} - 1$ from the former part of the above theorem, can only be meaningful for $v > 1+\sqrt{2} \approx 2.44$. Therefore, the latter part of Theorem \ref{thm:polydec} is only applicable to slightly stronger polynomial decays than the former part (which holds for $v>2$). Now, assuming $v>1+\sqrt{2}$, the additional requirement of $m = O\left(\sqrt{\frac{n^{k - \frac{2k(\alpha + 1)}{v} - \frac{4v}{v+1}}}{(\log n)^k}}\right)$ is only meaningful for a sufficiently large $k$. In particular, for $k\geq \frac{4v^2}{(v+1)(v - 2(\alpha+1))}$. When this happens, Theorem \ref{thm:polydec} guarantees the optimal rate $\Ed{D}{\Err(\fhat_C)} = O\left(\frac{1}{n^{\frac{v}{v+1}}}\right)$.

\begin{theorem}[Kernels with exponential eigenvalue decay]\label{thm:eigdec}
Let $f^*\in \H$ and suppose kernel $K$ has eigenvalues that decay as: $\lambda_j \leq c_1\exp(-c_2 j^2)$. Let $m$ denote the number of partitions, and let $k\geq 2$ such that Assumption \ref{asmp:eigfn} holds for this $k$. Then, the overall error for the $DC$-estimator $\fhat_C$ is given as:
\begin{align}
\Ed{D}{\Err(\fhat_C)}& = O\bigg(\lambda  \norm{f^*}{\H}^2 + \frac{\sigma^2}{n}g(\lambda) S(\lambda) + m\left(\frac{\log n (\log \log n)}{n}\right)^{k/2}\left(\norm{f^*}{\H}^2 + \frac{\sigma^2}{\lambda}\right)\bigg)
\end{align}
Now, if $m = O\left(\sqrt{\frac{n^{(k-4)}}{(\log n \log \log n)^{k}}}\right)$ and Assumption \ref{asmp:goodpart} holds at $\lambda = 1/n$, then the $DC$-estimator $\fhat_C$ achieves the optimal rate: $\Ed{D}{\Err(\fhat_C)} = O\left(\frac{\sqrt{\log n}}{n}\right)$ at $\lambda = 1/n$.
\end{theorem}

Here, as in the earlier two cases, the requirement on $m$ above is only meaningful for a sufficiently large $k$, in particular $k\geq 4$. In this case, Theorem \ref{thm:eigdec} gives the minimax optimal rate $\Ed{D}{\Err(\fhat_C)} = O(\frac{\sqrt{\log n}}{n})$.

\subsection{$f^*\notin \H$ --- With approximation error}
For the case when $f^*\notin \H$, note that we need not necessarily have $\App_i(\lambdabar)=0$ for any $\lambdabar>0$, $i\in [m]$. At $\lambdabar=0$ we will always have $\App_i(\lambdabar)=0$, however $f_{i,\lambdabar}$ may not be bounded (in other words, no element in $\H$ would achieve this approximation). One situation where we can still have $\App_i(\lambdabar)=0$ with $\lambdabar=0$, while having $f_{i,\lambdabar}$ to be bounded, is if $f^*$ is a piece-wise kernel function over our chosen partitions i.e. $f^*(x) = f^*_i(x)$ if $x\in C_i$, with $f^*_i \in \H$. This would then be analogous to the previous section. In general, however, without enforcing further assumptions on $f^*$, it is hard to give meaningful bounds on $\App_i(\cdot)$. While we can still proceed as in the previous section to obtain exact expressions for the \textit{regularization, bias} and \textit{variance} terms for $\Ed{D}{\Err(\fhat_C)}$, in this situation it may be more instructive to compare our bounds with the bounds for the averaging estimator in \cite{zhang13}. Let us denote this estimator as $\fhat_{avg}$. To compute $\fhat_{avg}$, the samples $n$ are randomly split into $m$ groups, and a KRR estimate is computed for each group. $\fhat_{avg}$ is then simply the average of the estimates over all groups. In this case, we have from \cite{zhang13} (for any $\lambdabar\in [0,\lambda]$):
\begin{align}\label{eq:zhangdecomp}
\Ed{D}{\Err(\fhat_{avg})} \leq 2\left(\App(\lambdabar) + \mathcal{E}(n,m,\lambda, \lambdabar)\right)
\end{align}
where $\App(\lambdabar)$ corresponds to the overall approximation term and $\mathcal{E}(R,n,m,\lambda)$ is the residual error term. In particular, $\App(\lambdabar)= \E{(f^*(x) - f_{\lambdabar}(x))^2}$ with $f_{\lambdabar}$ being the overall \textit{population} KRR estimate:
\begin{align}\label{eq:overallbest}
f_{\lambdabar} = \argmin_{f\in \H} \E{(f^*(x) - f(x))^2} + \lambdabar \norm{f}{\H}^2
\end{align}
Also, under certain conditions and restrictions on the number of partitions $m$, \cite{zhang13} can establish the scaling:
\begin{align}
\mathcal{E}(N,m,\lambda, \lambdabar) = O\left(\lambda \norm{f_{\lambdabar}}{\H}^2 + \frac{S(\lambda)}{n} \right)
\end{align}

In comparison, for our $DC$-estimator, we can have a (potentially) different $\lambdabar_i$ for each partition, and get the decomposition (similar to Eq. \eqref{eq:wholedecomp}):
\begin{align}\label{eq:difflamdecomp}
&\Ed{D}{\Err(\fhat_C)}\leq  2\bigg(\sum_{i=1}^{m}\App_{i}(\lambdabar_i) +\underbrace{\sum_{i=1}^{m}\Reg_i(\lambdabar_i,\lambda) + \sum_{i=1}^{m} \Bias_i(\lambda, n) + \sum_{i=1}^{m} \Ed{D}{\Var_{i}(\lambda,D)}}_{\mathcal{E}_C}\bigg)
\end{align}

Before comparing the bounds in Eq. \eqref{eq:zhangdecomp} with Eq. \ref{eq:difflamdecomp}, we require an additional definition. For any partition $C_i$, $i\in[m]$, let us define
\begin{align}
\AppErr_i(f_{\lambdabar}) = \E{(f^*(x) - f_{\lambdabar}(x))^2\indic{x\in C_i}}
\end{align} 
\text{i.e.} the error incurred by the global estimate $f_{\lambdabar}$ (Eq. \eqref{eq:overallbest}) in the $i^{th}$ partition. Note that, $\sum_{i=1}^{m} \AppErr_i(f_{\lambdabar}) = \App(\lambdabar)$. To avoid confusion, we would like to emphasize the distinction between $\AppErr_i(f_{\lambdabar})$ and $\App_i(\lambdabar_i)$. While the former is the \textit{local} error (in the $i^{th}$ partition) incurred by solving a global problem with regularization $\lambdabar$, the latter is the \textit{local} error incurred by solving a \textit{local} problem with regularization $\lambdabar_i$ (as defined in Eq \eqref{eq:defapp}).

Now, to simplify presentation in the sequel, let us assume that $CE_i = \CovErr_{i}(\lambda p_i, n, k) = \widetilde{O}\left(n^{-1/2}\right)$ --- which was the case for the kernels discussed in Section \ref{subsec:covbnd}. Also, suppose the quintuplet $(n,m,k,p_i,\lambda,\lambdabar)$, for any $i\in[m]$, satisfies:
\begin{align}\label{eq:quintcond}
m = O\left(\max\left(\lambda n^{\frac{k-2}{2}}, \frac{n^{\frac{k-2}{2}}}{\norm{f_{\lambdabar}}{\H}^2} \right)\right), \quad p_i = \Omega\left(\min\left(\frac{m^2}{\lambda^2 n^{k-2}}, \frac{\App(f_{\lambdabar})}{n^{k/2}\lambdabar} \right)\right)
\end{align}
The above restrictions essentially guarantee that all terms involving $CE_i^k$ in Lemma \ref{lem:mainbiasbnd} and Lemma \ref{lem:mainvarbnd} are of a lower order. Then, we have the following theorem: 
\begin{theorem}\label{thm:thm3}
Consider any $\lambdabar > 0$. Let $f_{\lambdabar}$ be the solution of Eq. \eqref{eq:overallbest}. Then, $\exists \lambdabar_1, \ldots, \lambdabar_m$ with $\lambdabar_i \in[0, \lambdabar], i\in[m]$, such that, 
\begin{align}
\App_i(\lambdabar_i) &\leq \AppErr_i(f_{\lambdabar})\quad\text{and,}\\
\norm{f_{i,\lambdabar_i}}{\H} &= O\left( \norm{f_{\lambdabar}}{\H} + \sqrt{\frac{\AppErr_i(f_{\lambdabar})}{(\lambdabar p_i)}}\right)
\end{align}
Thus, $\sum_{i=1}^{m}\App_i(\lambdabar_i) \leq \App(\lambdabar)$

Moreover, if $\App(\lambdabar) = O\left(\lambdabar \norm{f_{\lambdabar}}{\H}^2\right)$, Assumption \ref{asmp:eigfn} holds, Assumption \ref{asmp:approx} holds ($\forall \lambdabar_i$), Assumption \ref{asmp:goodpart} holds, and the quintuplet $(n,m,k,p_i, \lambda, \lambdabar)$ satisfies the restriction in Eq. \ref{eq:quintcond}, then
\begin{align}
\mathcal{E}_C = O\left(\lambda \norm{f_{\lambdabar}}{\H}^2 + \frac{\sigma^2 S(\lambda)}{n} \right)
\end{align}
\end{theorem}

The above theorem shows that the \textbf{approximation error term} in each partition of our estimator in $\Ed{D}{\Err(\fhat_C)}$ is lower than its counterpart in $\Ed{D}{\Err(\fhat_{avg})}$. Consequently, the overall approximation term is also lower. On the other hand, the residual \textbf{estimation error} terms can be of the same order. Intuitively, this makes sense since by partitioning the space, we are fitting piece-wise kernel functions, as opposed to just a single kernel function in the averaging case. We demonstrate this through experiments in the next section.

\section{Experiments}\label{sec:experiments}
\begin{figure*}[t]
  \centering
  \begin{tabular}{ccc}
    \subfigure[Piece-wise constant]{
    \label{fig:toy1}
    \includegraphics[width=0.3\linewidth]{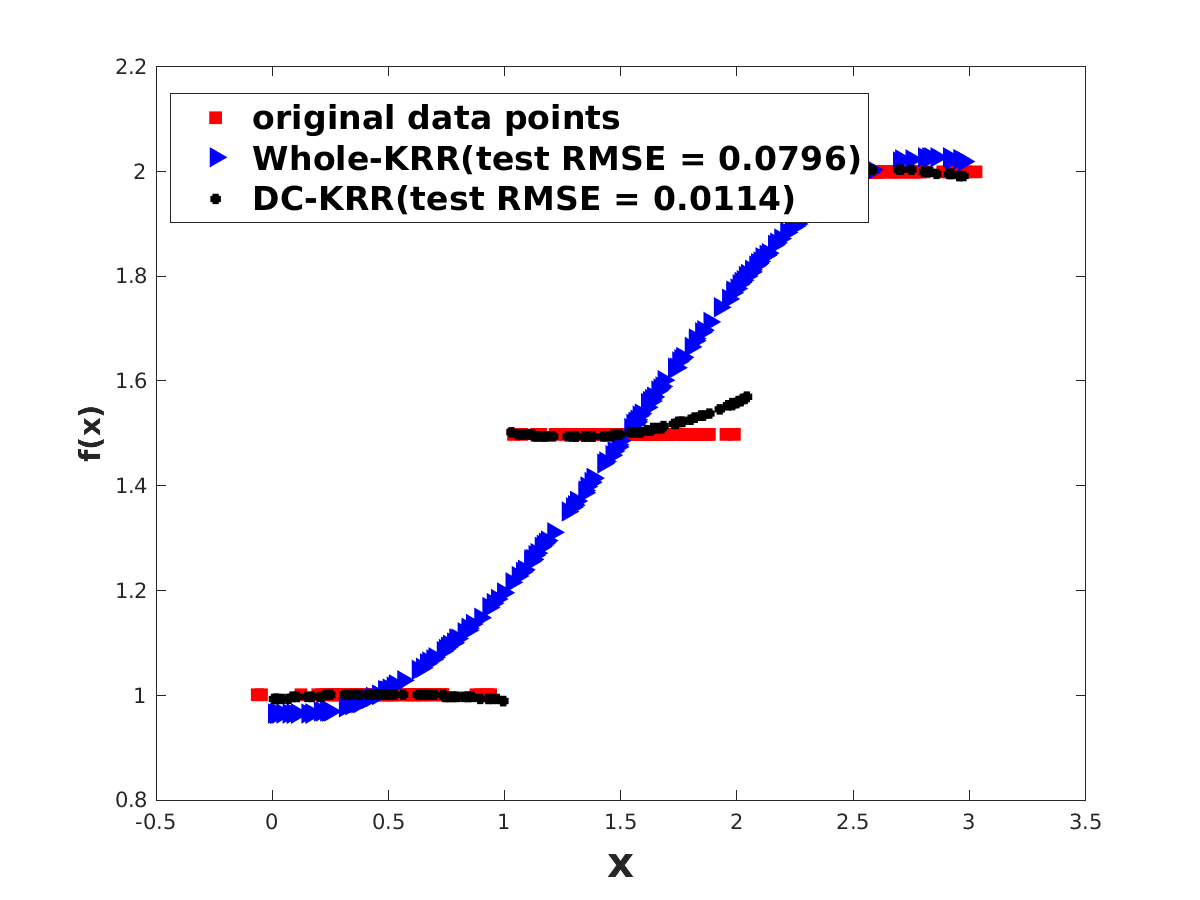}
    }&\hspace{-15pt} 
    \subfigure[Piece-wise Gaussian]{
    \label{fig:toy2}
    \includegraphics[width=0.3\linewidth]{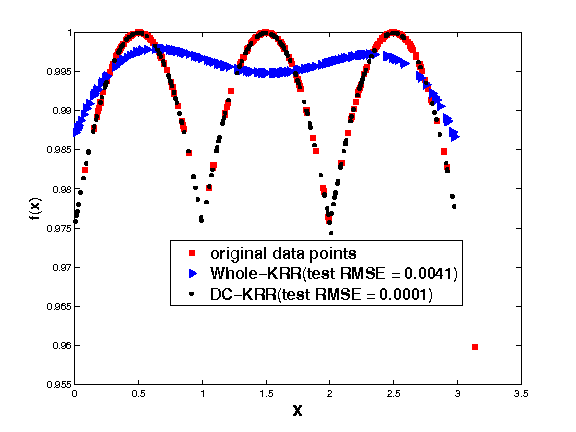}
    } &
\hspace{-15pt} 
    \subfigure[Sine]{
    \label{fig:toy3}
    \includegraphics[width=0.3\linewidth]{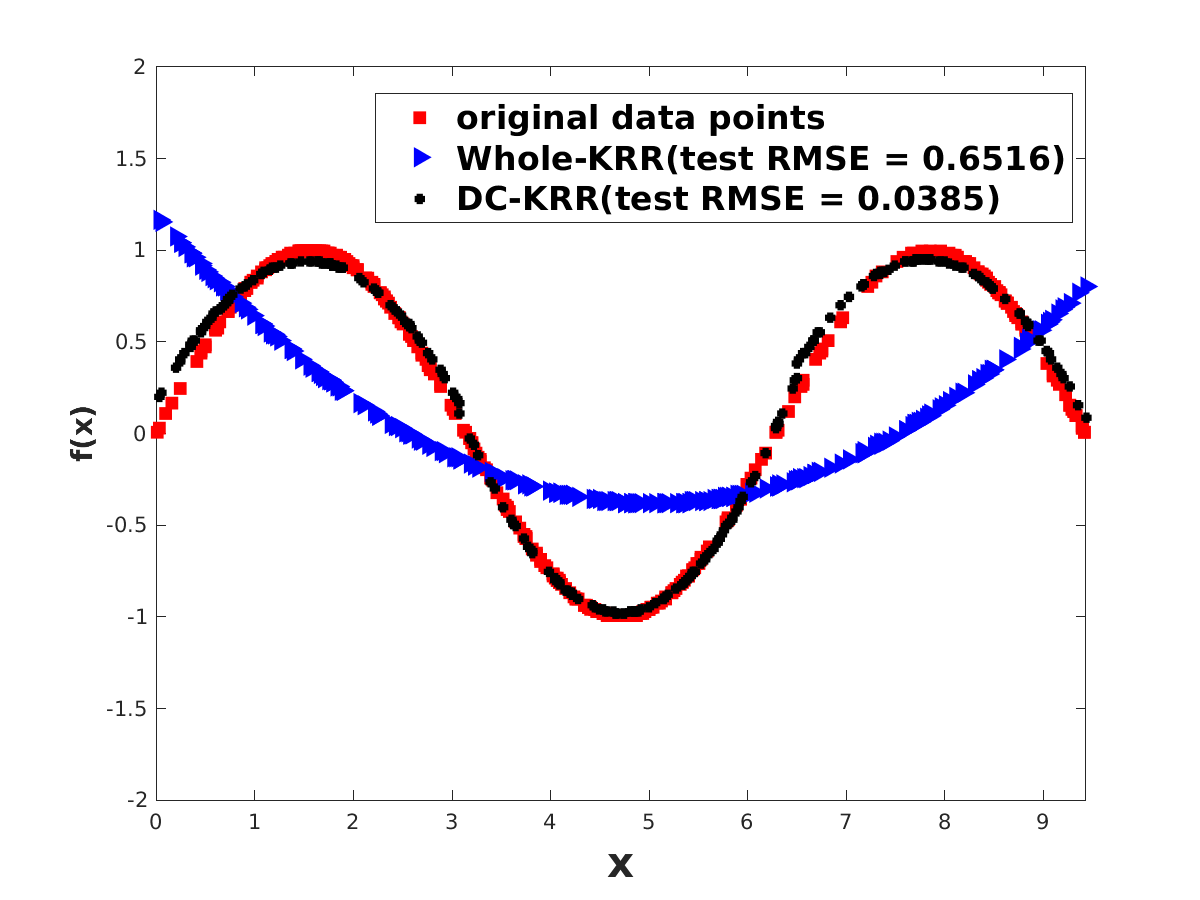}
    }
  \end{tabular}
  \caption{Plots of functions obtained via Whole-KRR and DC-KRR (with 3 partitions).}
  \label{fig:toy_function}
\end{figure*}

\begin{figure*}[t]
  \centering
  \begin{tabular}{ccc}
    \subfigure[Piece-wise constant]{
    \label{fig:toy1_rmse}
    \includegraphics[width=0.33\linewidth]{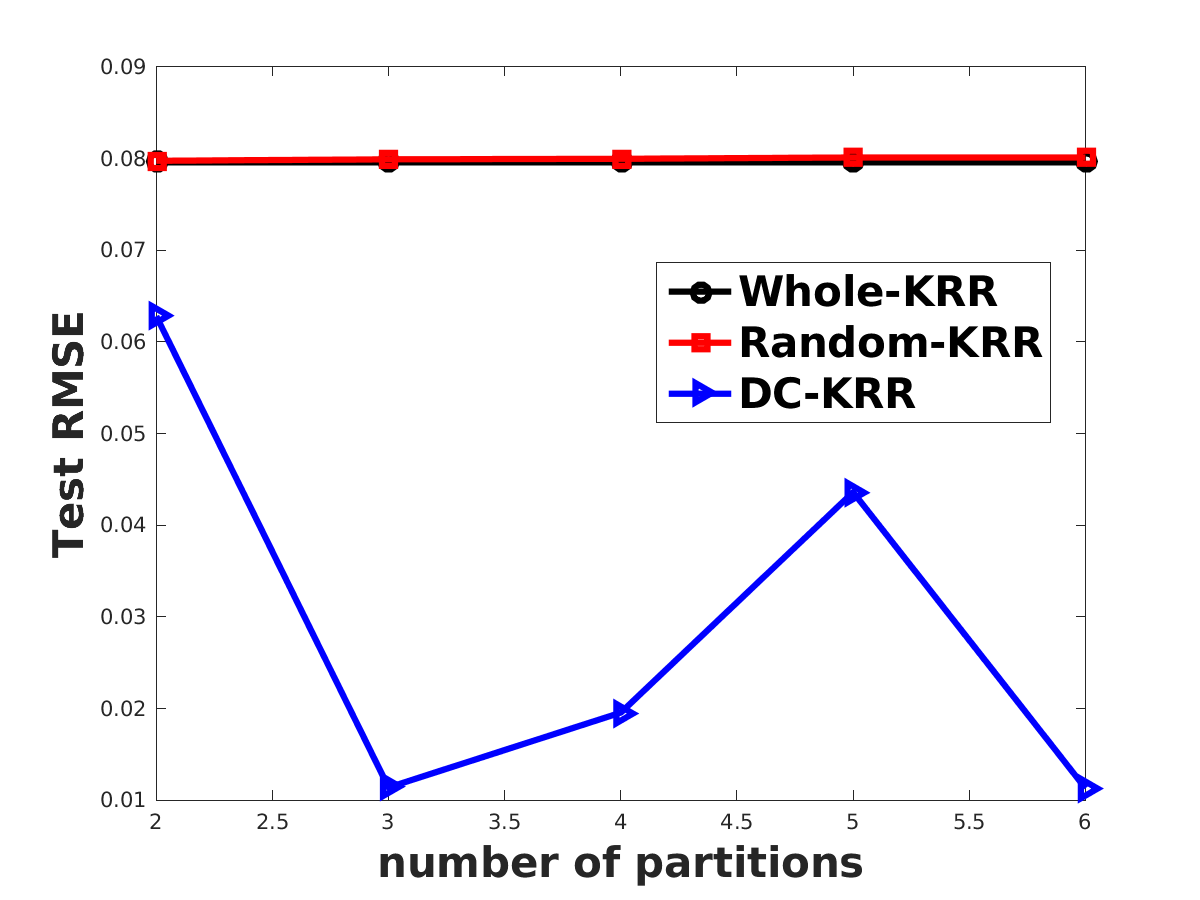}
    }&\hspace{-15pt} 
    \subfigure[Piece-wise Gaussian]{
    \label{fig:toy2_rmse}
    \includegraphics[width=0.33\linewidth]{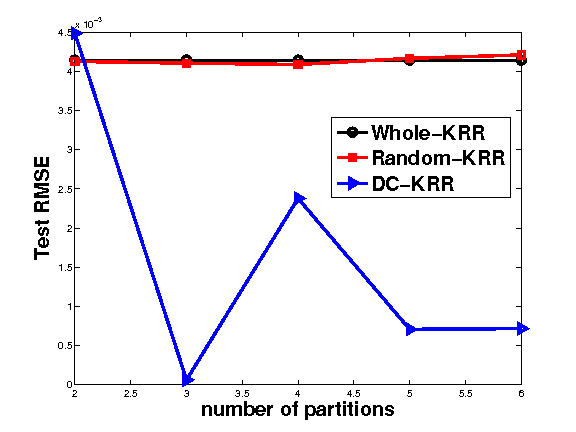}
    } &
\hspace{-15pt} 
    \subfigure[Sine]{
    \label{fig:toy3_rmse}
    \includegraphics[width=0.33\linewidth]{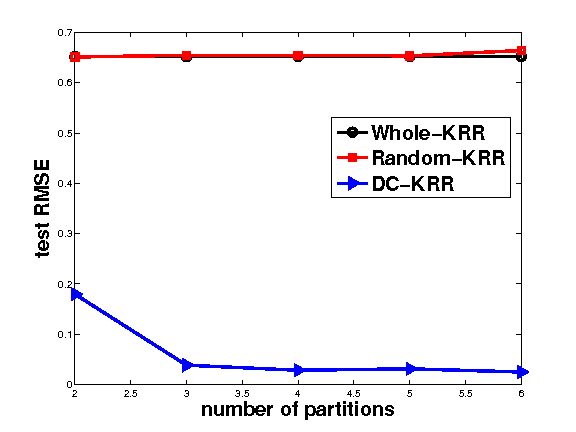}
    }
  \end{tabular}
  \caption{Plots of Test RMSE vs. Number of partitions on Three Toy data sets.}
  \label{fig:rmse_toy}
\end{figure*}

\begin{table*}[t]
  \centering
  \caption{Data set statistics for real data sets used in our experiments. $\gamma$ was chosen using cross-validation on the entire data set, or a sub-sample of size $10,000$ for larger data sets.}
  \label{tab:multilabel_datasets}
\resizebox{12cm}{!}{
  \begin{tabular}{|c|r|r|r|r|}
    \hline
    Data set &\# training samples & \# testing samples & \# features & $\gamma$\\ 
    \hline
   house  &404 & 102 & 13 & $10^{-4}$  \\
   air &1,202 & 301  & 5 & $10^{-3}$   \\
   cpusmall  &6,553 & 1,639  & 12  & $10^{-1}$  \\
   Pole  & 12,000 & 3,000 & 26& 1\\
   CT Slice & 42,800 & 10,700 & 385& $10^{-2}$\\
 Road  & 347,899  & 86,974& 3&$0.1$ \\ 
   \hline
  \end{tabular}
  }
\end{table*}

\begin{table*}[t]
  \centering
  \caption{Test RMSE and Training Time on real data sets used in our experiments. \# partitions is only applicable to the Random-KRR and DC-KRR columns.}
  \label{tab:realdat}
\resizebox{18cm}{!}{
  \begin{tabular}{|c|r|r|r|r|r|r|r|r|r|}
    \hline
     Data set &\# partitions & \multicolumn{2}{|c}{Whole-KRR} & \multicolumn{2}{|c|}{Random-KRR}& \multicolumn{2}{|c}{DC-KRR(kernel k-means)}& \multicolumn{2}{|c|}{DC-KRR(k-means)}  \\ 
   \hline
      & 		& Test RMSE & Time(s)& Test RMSE & Time(s)& Test RMSE & Time(s) & Test RMSE & Time(s)  \\
\hline
    house  &4 &4.4822 & 0.08&4.5609&0.02 & {\bf 3.3849} & 0.18 & 3.8244& 0.06  \\
    air &8 & 4.3537 & 2.46 & 4.6604 & 0.07 &{\bf 4.2577} & 0.79 & 4.4782 & 0.23      \\
    cpusmall  &8 & 5.8853&118.98 & 7.1757&4.04&{\bf 5.7947}&30.86 &6.4616&7.86 \\
    Pole  & 16 & {\bf 14.7256} & 1088.9 & 21.5768 &6.15 &15.0005 &277.80& 15.1167 & 11.88  \\
    CT Slice & 32 & {\bf 2.1165} & 3840.7 & 10.0318& 43.81 & 3.6100&405.38 & 2.4302&64.06  \\
  Road & 256 & - & -  &13.6444 & 43.48& 11.0550& 1081.3& {\bf 8.6358}& 78.16  \\ 
    \hline
  \end{tabular}
  }
\end{table*}

\begin{figure*}[t]
  \centering
  \begin{tabular}{ccc}
    \subfigure[house]{
    \label{fig:real1_rmse}
    \includegraphics[width=0.33\linewidth]{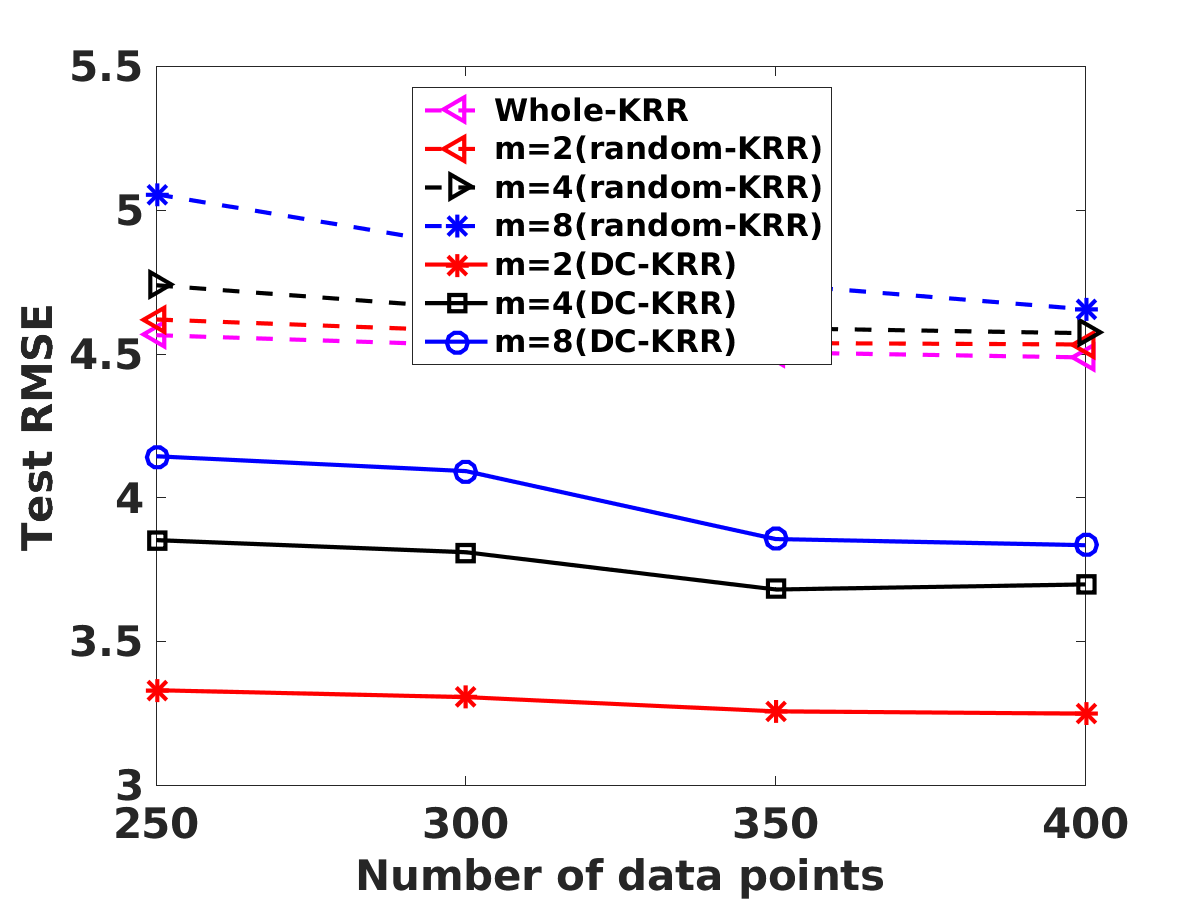}
    }&\hspace{-15pt} 
    \subfigure[air]{
    \label{fig:real2_rmse}
    \includegraphics[width=0.33\linewidth]{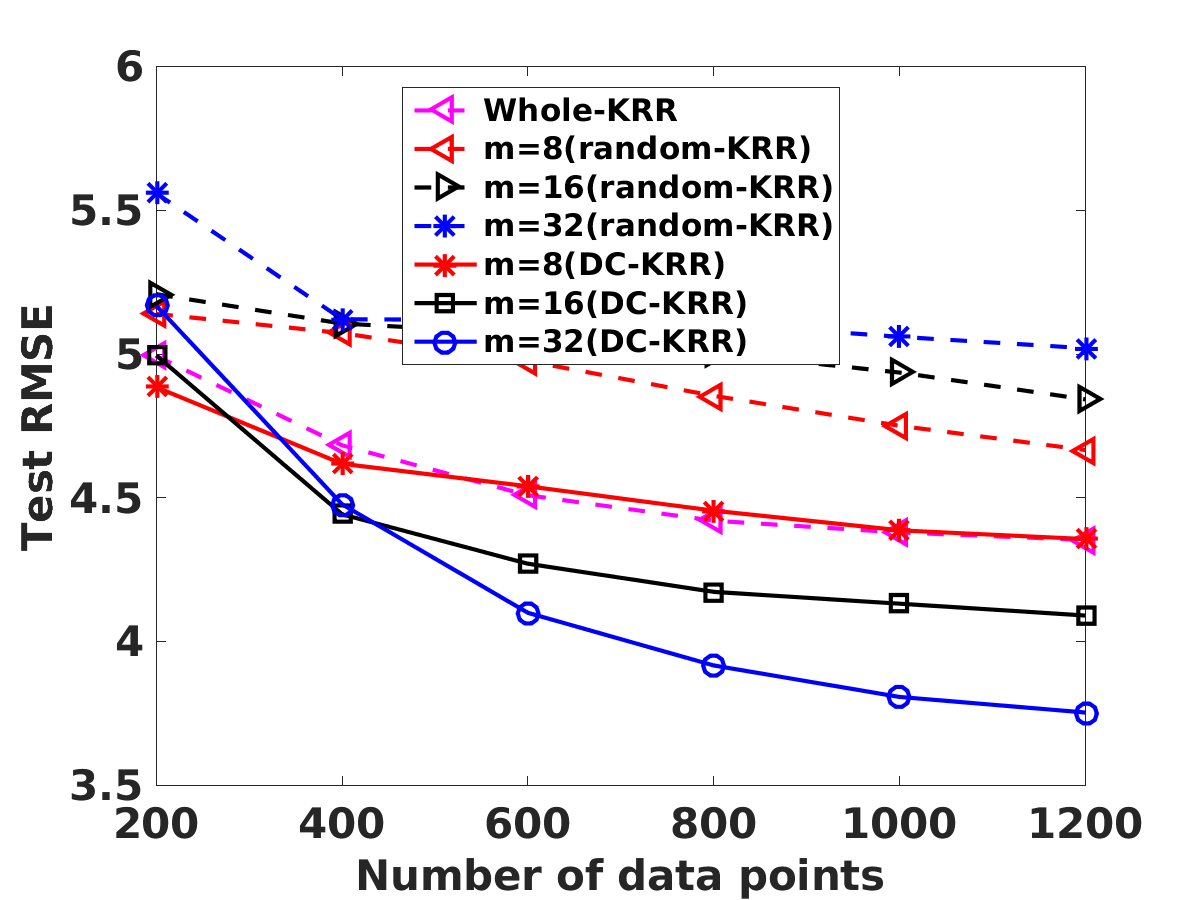}
    } &\hspace{-15pt} 
    \subfigure[cpusmall]{
    \label{fig:real3_rmse}
    \includegraphics[width=0.33\linewidth]{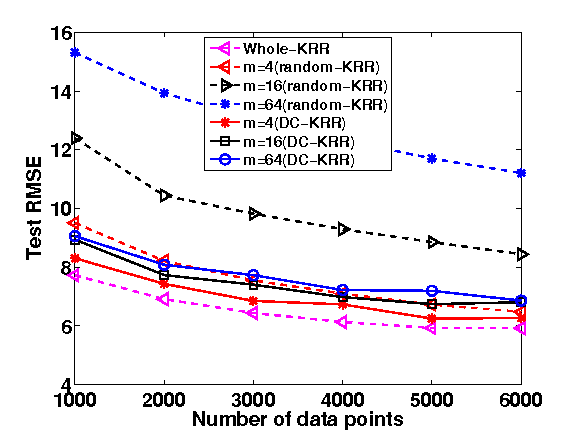}
    } \\
    \subfigure[Pole]{
    \label{fig:real4_rmse}
    \includegraphics[width=0.33\linewidth]{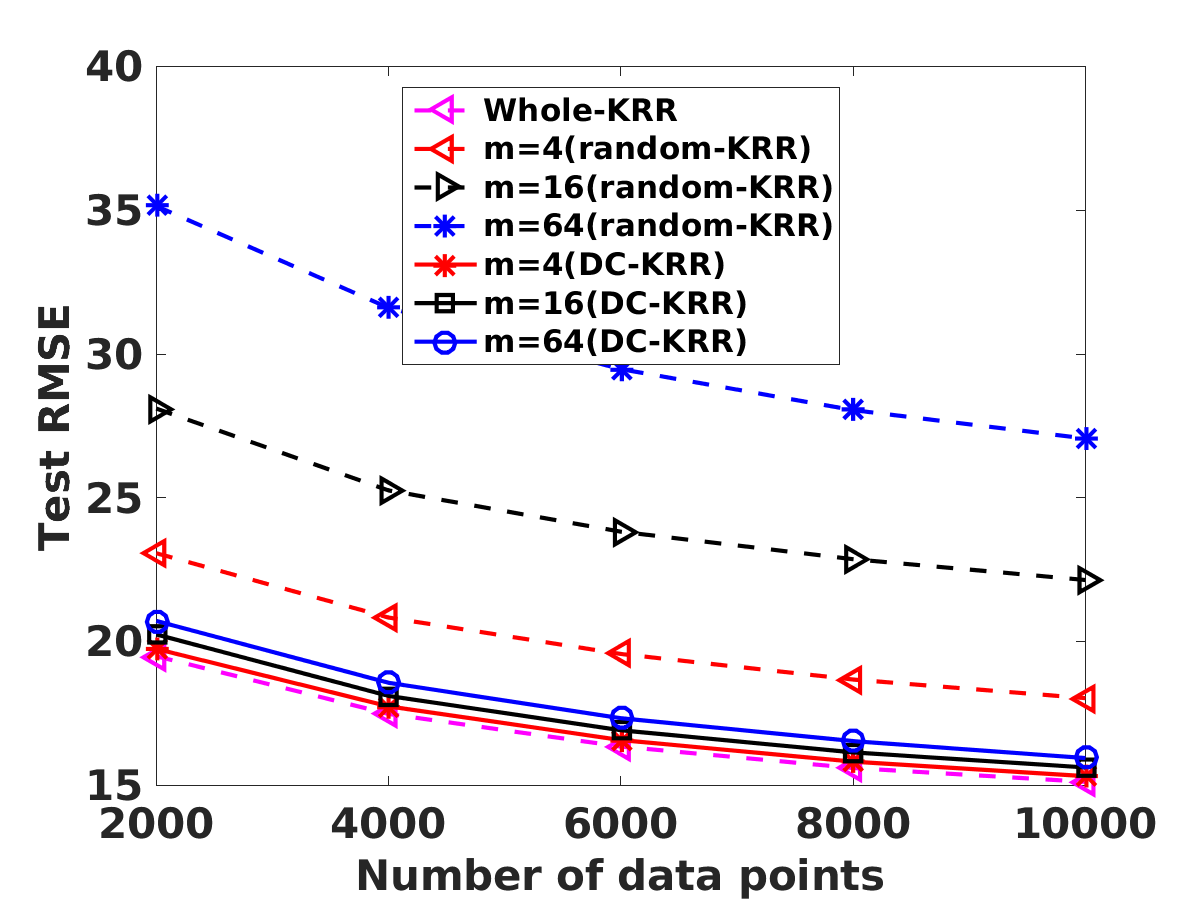}
    }&\hspace{-15pt} 
    \subfigure[CT Slice]{
    \label{fig:real5_rmse}
    \includegraphics[width=0.33\linewidth]{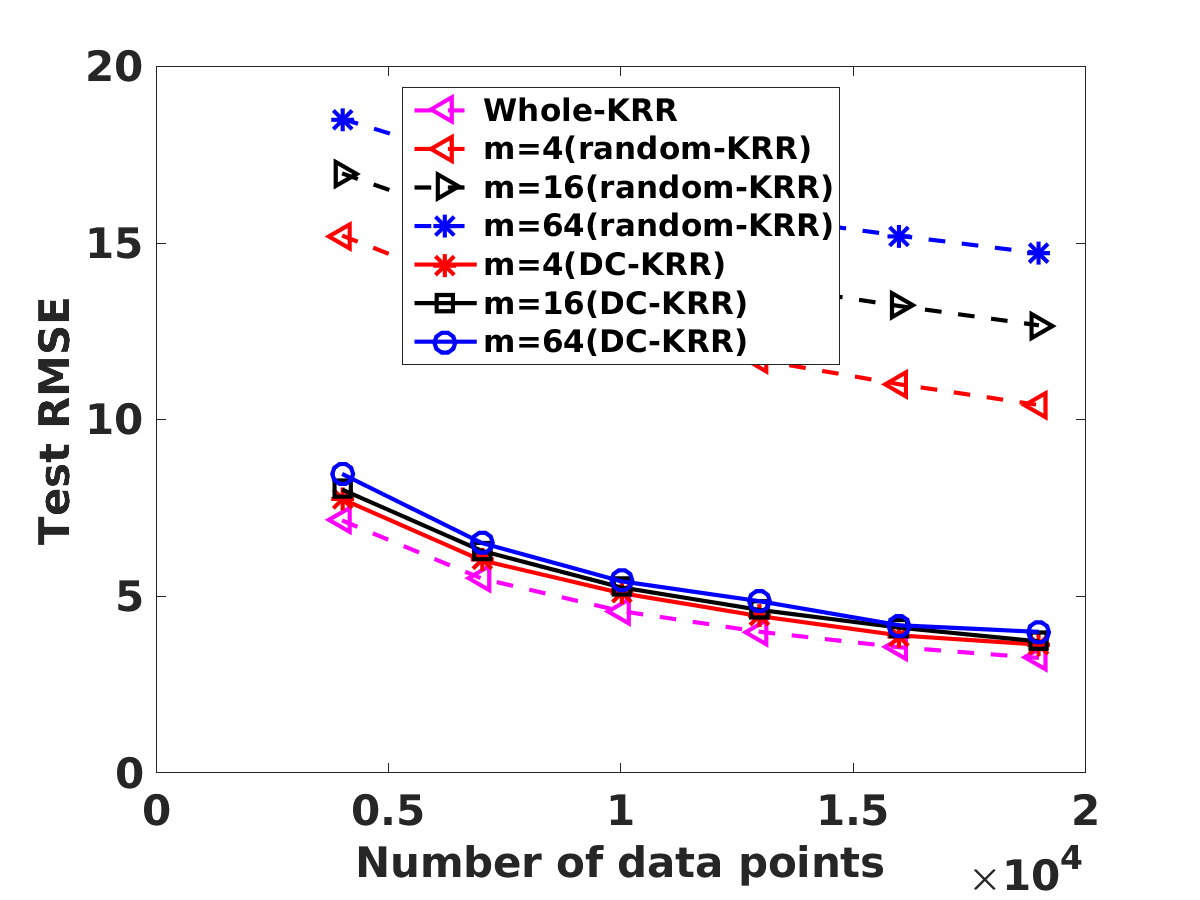}
    } &\hspace{-15pt} 
    \subfigure[Road]{
    \label{fig:real6_rmse}
    \includegraphics[width=0.33\linewidth]{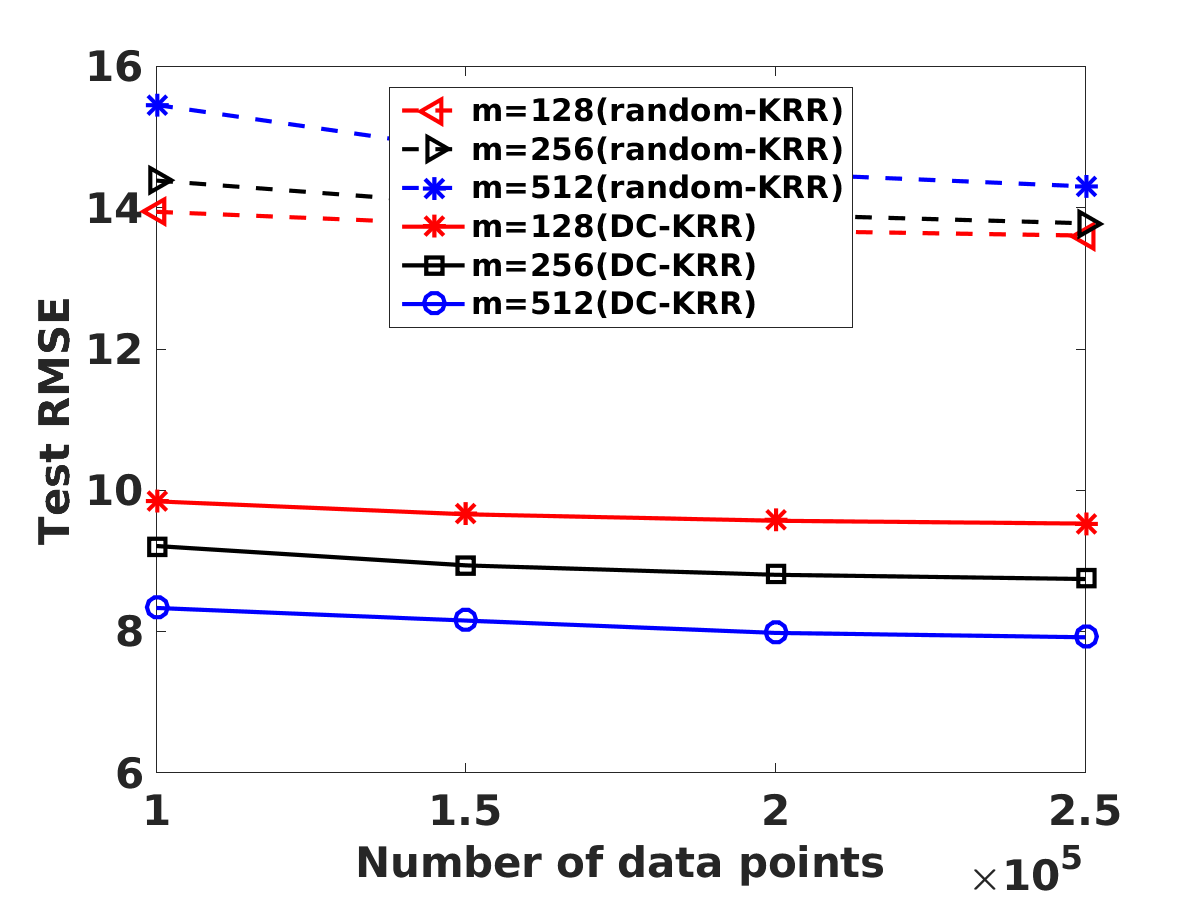}
    }
  \end{tabular}
  \caption{Plots of Test error vs. Training size, $n$ on real data sets. $m$ denotes the number of partitions. The DC-KRR plots correspond to a k-means clustering partition.}
  \label{fig:realdat}
\end{figure*}

\begin{figure}[ht]
  \centering
  \begin{tabular}{ccc}
    \subfigure[house]{
    \label{fig:real1}
    \includegraphics[width=0.33\linewidth]{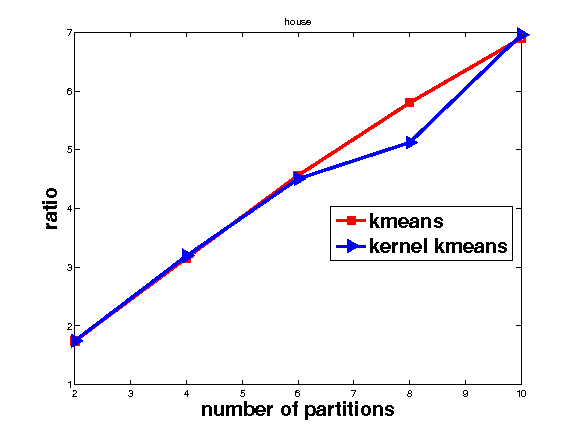}
    }&\hspace{-15pt} 
    \subfigure[air]{
    \label{fig:real2}
    \includegraphics[width=0.33\linewidth]{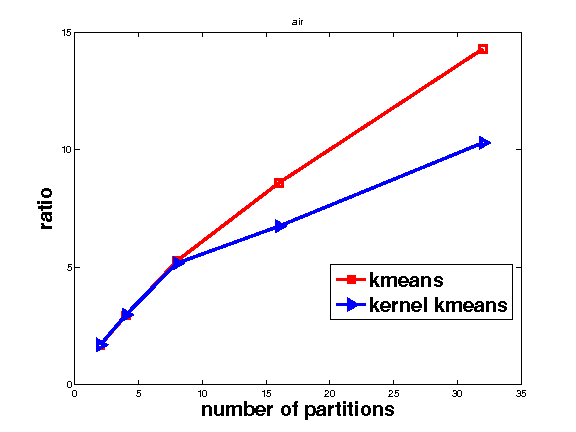}
    } &\hspace{-15pt} 
    \subfigure[cpusmall]{
    \label{fig:real3}
    \includegraphics[width=0.33\linewidth]{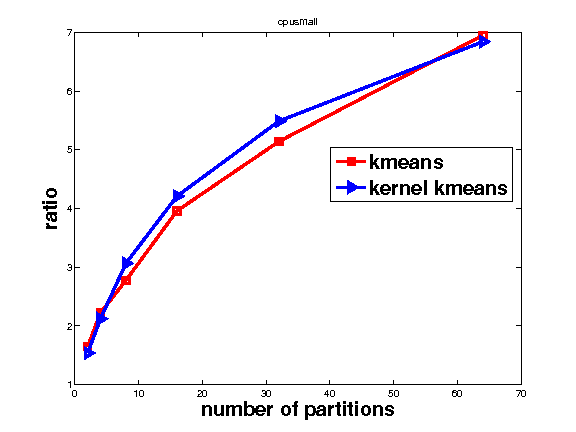}
    } \\
    \subfigure[Pole]{
    \label{fig:real1}
    \includegraphics[width=0.33\linewidth]{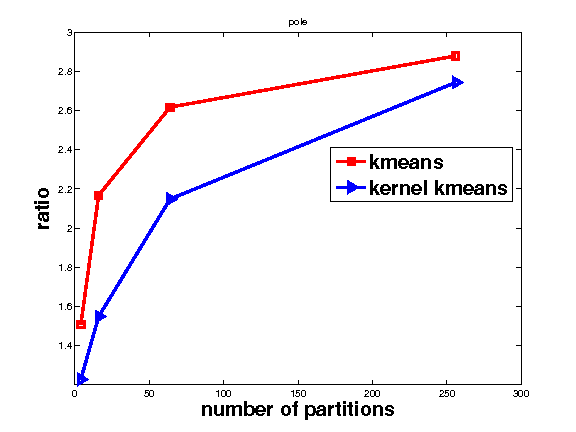}
    }&\hspace{-15pt} 
    \subfigure[CT Slice]{
    \label{fig:real2}
    \includegraphics[width=0.33\linewidth]{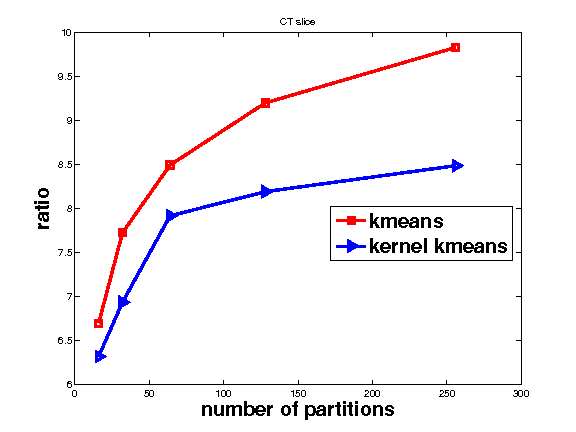}
    } &\hspace{-15pt} 
    \subfigure[Road]{
    \label{fig:real3}
    \includegraphics[width=0.33\linewidth]{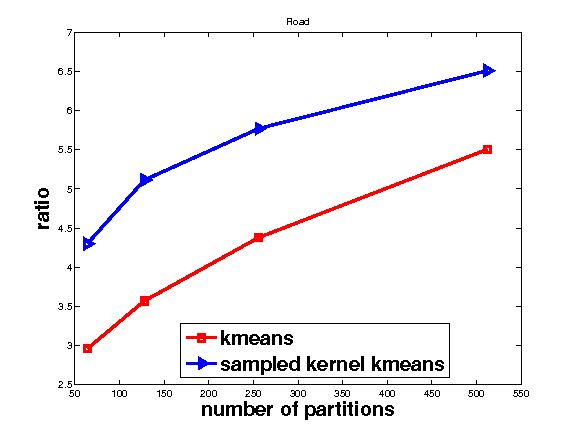}
    }
  \end{tabular}
  \caption{Plots of $g(\lambda)$ vs. number of partitions on real data sets.
}
  \label{fig:goodnessrealdat}
\end{figure}
\begin{figure}[hb]
  \centering
  \begin{tabular}{ccc}
    \subfigure[Piece-wise constant]{
    \label{fig:real1}
    \includegraphics[width=0.33\linewidth]{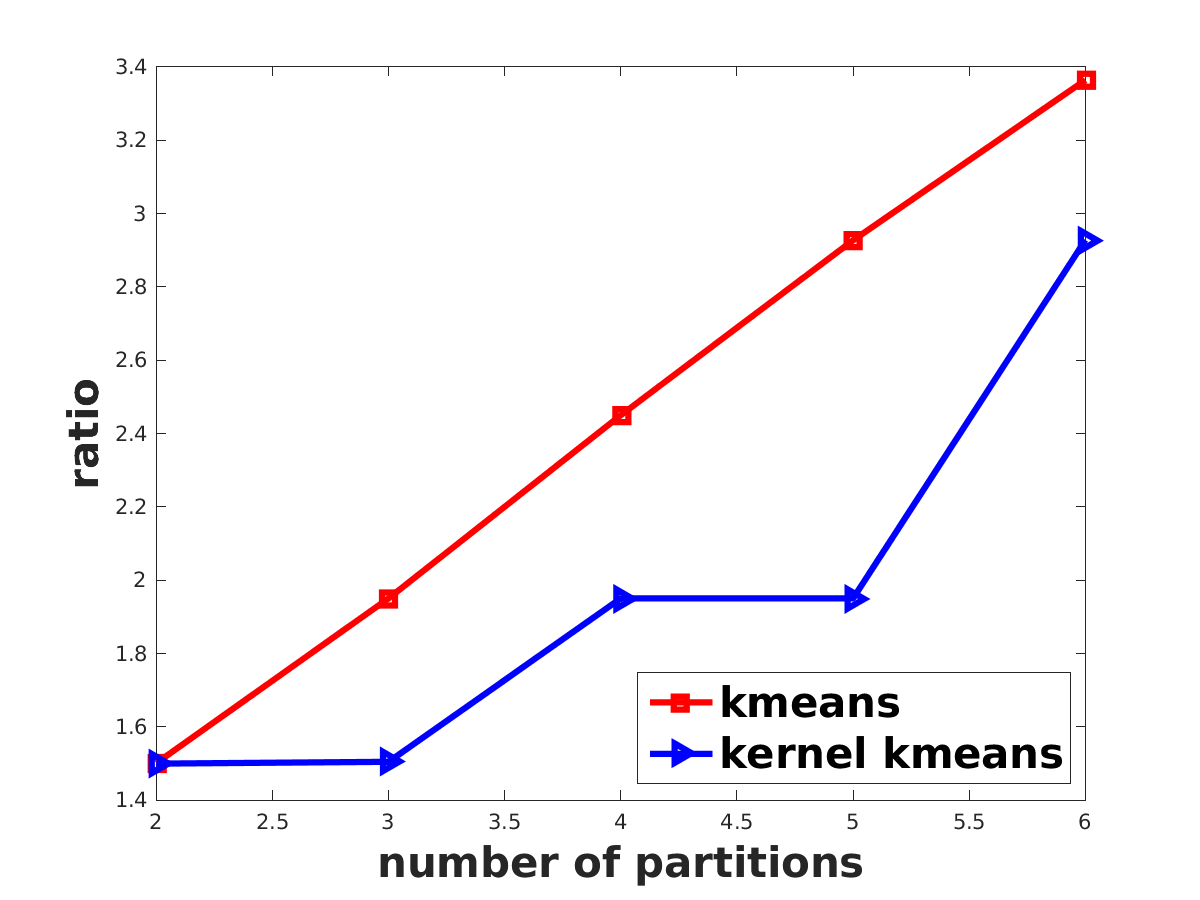}
    }&\hspace{-15pt} 
    \subfigure[Piece-wise Gaussian]{
    \label{fig:real2}
    \includegraphics[width=0.33\linewidth]{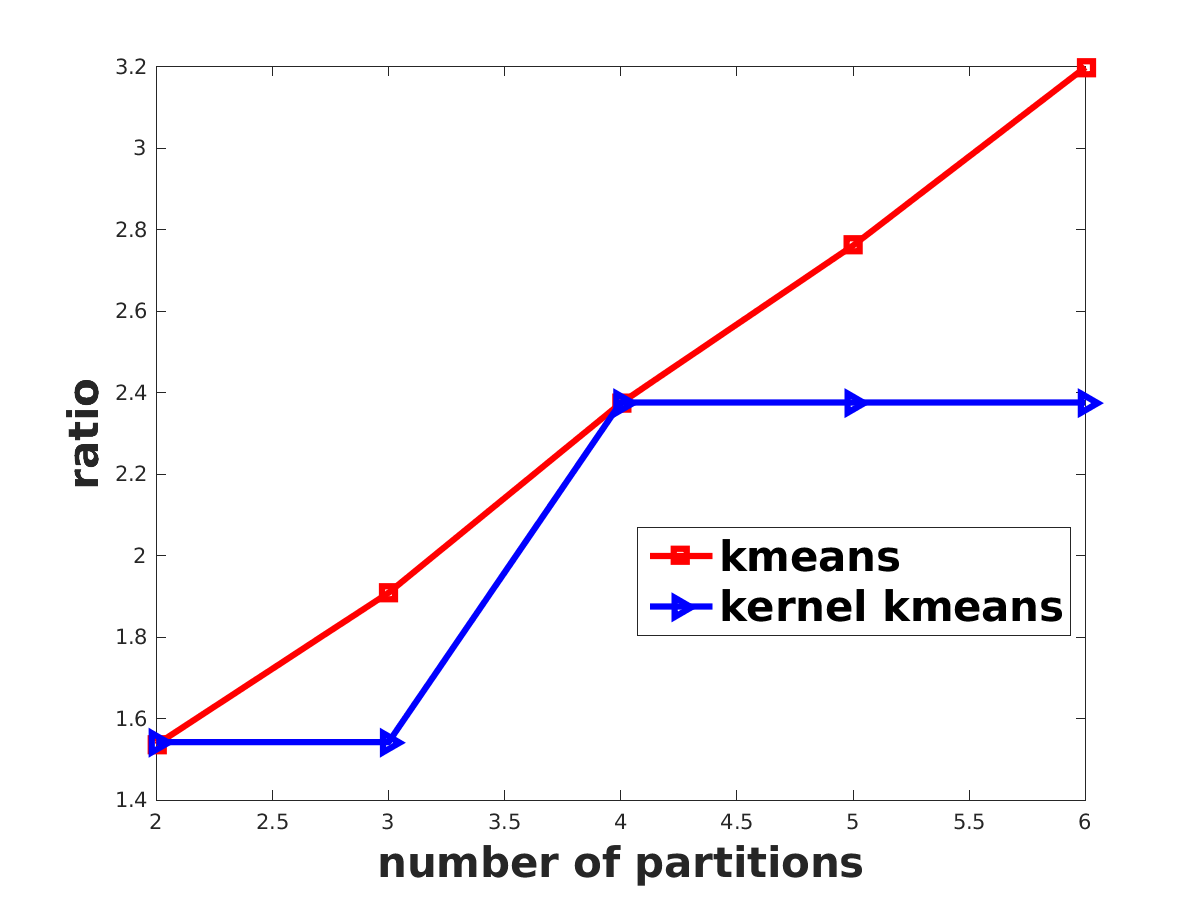}
    } &
\hspace{-15pt} 
    \subfigure[Sin]{
    \label{fig:real2}
    \includegraphics[width=0.33\linewidth]{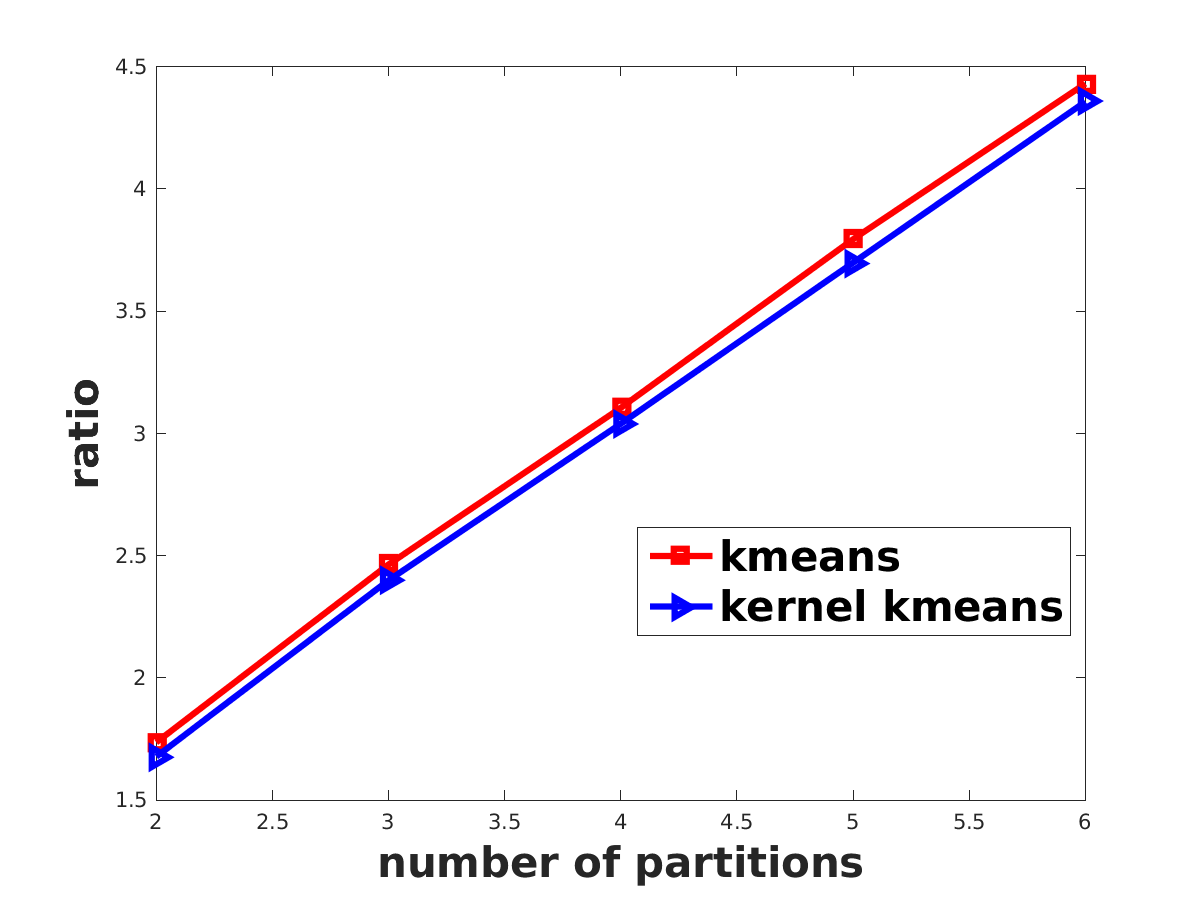}
    }
  \end{tabular}
  \caption{Plots of $g(\lambda)$ vs. Number of partitions on synthetic data sets.}
  \label{fig:goodness_toy}
\end{figure}
In this section, we present experimental results of our $DC$-estimator (denoted DC-KRR), on both real and toy data sets. For comparison, we tested against the random splitting approach of \cite{zhang13} (denoted Random-KRR), and KRR estimate on the entire training set (denoted Whole-KRR). In addition, we also present tests on $g(\lambda)$ (to validate Assumption \ref{asmp:goodpart}), and empirical comparisons with the approach of \cite{eberts15} (denoted VP-KRR).\\

\textbf{Toy Data sets}: We performed experiments on 3 toy data sets, shown in Fig~\ref{fig:toy_function}. In each case, the covariate $x$ was generated from a mixture of 3 Gaussians: $x \sim \frac{1}{3}\mathcal{N}(\mu_1, \sigma) + \frac{1}{3}\mathcal{N}(\mu_2,\sigma) + \frac{1}{3}\mathcal{N}(\mu_3,\sigma)$. For the first two toy examples, $(\mu_1 = 0.5,\mu_2 = 1.5,\mu_3 = 2.5)$ and $\sigma = 0.2$, and for the third one, $(\mu_1 = \pi/2,\mu_2 = 3 \pi/2,\mu_3 = 3 \pi)$ and $\sigma=1$. The response $y$ was $y = f^*(x) + \eta $, for different choices of $f^*$, and with $\eta\sim \mathcal{N}(0,0.05)$. For each data set, we generated a training set of size $600$, and a test set of size $100$.\\
 
We chose $f^*$ as: $(i)$ a piece-wise constant function, $f^*(x) = \indic{x\leq1} + 1.5\times\indic{1<x<2} + 2\times\indic{x\geq 2}$, in Fig~\ref{fig:toy1}, $(ii)$ a piece-wise Gaussian kernel function, $f^*(x) = \exp(-\gamma(x-0.5)^2)\times\indic{x\leq1} + \exp(-\gamma(x-1.5)^2)\times\indic{1<x<2} + \exp(-\gamma(x-2.5)^2)\times\indic{x\geq 2}$, with $\gamma=0.1$, in Fig~\ref{fig:toy2}, and $(iii)$ a sine function, $f^*(x) = \sin(x)$, in Fig~\ref{fig:toy3}. To obtain a KRR estimate, we used a Gaussian kernel ($K(x,y) = \exp(-\gamma(x-y)^2)$) with $\gamma= 0.1$ for the first two toy data sets, and degree 2 polynomial kernel ($K(x,y) = (1+xy)^2$) for the third one. When running DC-KRR, we obtained the partition of the data points using k-means. A regularization penalty of $\lambda = 1/n$ was used, where $n=$ Total number of training points.\\

Fig~\ref{fig:toy_function} shows a comparison of the functions obtained using DC-KRR (run with 3 partitions) and Whole-KRR. We see that DC-KRR could approximate the true underlying function better than Whole-KRR, while still being computationally more efficient. Fig~\ref{fig:rmse_toy} shows the Test-RMSE with varying number of partitions for DC-KRR, Whole-KRR and Random-KRR. We observe that while Random-KRR had a similar performance to Whole-KRR, DC-KRR achieved lower error than both. This can be attributed to lower approximation error of piece-wise estimates.\\

\textbf{Real Data sets}: We performed experiments on 6 real data sets from the UCI repository~\cite{uci}. Data sets statistics are presented in Table \ref{tab:multilabel_datasets}. The data was normalized to have standard deviation 1. In all cases, we utilized a Gaussian kernel with kernel parameter $\gamma$ chosen using cross-validation, as shown in Table \ref{tab:multilabel_datasets}. We varied the number of partitions, $m$, and the number of training points, $n$. When running DC-KRR, the partitions were determined using clustering, and we tested with k-means and Kernel k-means. Kernel k-means was run on a sub-sampled set of points for larger data sets. The regularization penalty for KRR was chosen as $\lambda = 1/n$. Results of these experiments are presented in Table \ref{tab:realdat} and Fig \ref{fig:realdat}.\\

In all cases, DC-KRR achieved lower test error than Random-KRR, while being comparable to Whole-KRR. Moreover, the training time for DC-KRR, when running via k-means, was similar to Random-KRR (due to the small overhead of clustering), but much faster than Whole-KRR. Interestingly, in two cases (Fig \ref{fig:real1_rmse} and Fig\ref{fig:real2_rmse}), we found that DC-KRR also achieved lower test error than Whole-KRR. This can also be attributed to a lower approximation error due to piece-wise estimates.\\

\textbf{Testing Goodness of Partitioning}: We also estimated $g(\lambda)$ (Eq. \eqref{eq:goodnessdef}) vs. a varying number of partitions, on both our real and toy data sets (shown in Fig \ref{fig:goodness_toy} and Fig \ref{fig:goodnessrealdat} respectively) to verify the validity of Assumption \ref{asmp:goodpart}.\\

To estimate $S(\lambda)$ and $S_i(\lambda p_i)$, $i\in [m]$ (which comprise $g(\lambda)$), we used an SVD to compute the eigenvalues of the kernel matrix on the training samples (respectively, the kernel matrix of the training samples in partition $i$) and normalized this with $n$, the training size (respectively, $n_i$, the training size in partition $i$). In case of larger data sets, we did this on a sub-sampled version of the data set. It is known that the eigenvalues of $K_D/n$, with $K_D$ being the kernel matrix on randomly sampled points $D$, converge to the eigenvalues of the covariance in the associated RKHS \cite{rosasco2010}. We used a Gaussian kernel and set $\lambda = 1/n$, the same as in our experiments, with $n=$ total training size/sub-sample size.\\

On real data sets, we found that while $g(\lambda)$ increases as the number of partitions increases, it continues to be a constant even for a large number of partitions in several cases, thereby justifying Assumption \ref{asmp:goodpart}. On synthetic data sets, it seemed to grow at a somewhat faster rate. However, this could be attributed to lesser clustering structure, since the true number of clusters was only $3$ --- at which point $g(\lambda)$ is still a small constant.
\begin{table*}[t]
  \centering
  \caption{Test RMSE and Training Times on cpusmall for  VP-KRR(\cite{eberts15}), Random-KRR and DC-KRR(with k-means and kernel k-means). \# partitions is only applicable to the Random-KRR and DC-KRR columns. For VP-KRR(\cite{eberts15}), we choose the radius for obtaining voronoi partitions, $r$, to be $\alpha$ times the maximum distance between any two points in the data set, with $\alpha$ chosen as 0.01, 0.04, 0.07 and 0.12. After we know the number of partitions for a specific $r$, we generate the same number of partitions using k-means and kernel k-means (for DC-KRR), and random partitioning (for Random-KRR).} 
  \label{tab:compare_eberts}
\resizebox{14cm}{!}{
  \begin{tabular}{|c|r|r|r|r|r|r|r|r|}
    \hline
      \# partitions  & \multicolumn{2}{|c}{6} & \multicolumn{2}{|c|}{9}& \multicolumn{2}{|c}{13}& \multicolumn{2}{|c|}{40}  \\ 
   \hline
      		& Test RMSE & Time(s)& Test RMSE & Time(s)& Test RMSE & Time(s) & Test RMSE & Time(s)  \\
\hline
     VP-KRR   &5.8914&129.9600&5.8653&119.9500 & 6.1331 &113.0400 & 6.3026& 49.69  \\
    Random-KRR &  6.6232 & 4.49 &7.3143 & 2.2400  & 7.9986  &1.2100 &10.1980 & 0.2500       \\
   DC-KRR(k-means) &6.4246&24.72 &6.4610&8.6800& 6.6415 &4.1700 &7.2206 &0.9400 \\
   DC-KRR(kernel k-means) &5.7819&17.0900 &5.8338&14.4700& 5.8069 &13.00 &6.01 &12.09 \\
    \hline
  \end{tabular}
}
\end{table*}

\textbf{Comparison with \cite{eberts15}}: We also performed additional empirical comparisons between the approach in \cite{eberts15} (denoted as VP-KRR), DC-KRR (with k-means and kernel k-means) and Random-KRR, on the cpusmall data set (see Table \ref{tab:multilabel_datasets}). The main algorithmic difference between DC-KRR and VP-KRR is that the latter proposes to obtain bounded partitions using a Voronoi partitioning of the input space, while in DC-KRR we use a clustering algorithm to obtain the partitions. The results of our tests are shown in Table\ref{tab:compare_eberts}. We see that DC-KRR(with kernel k-means) was slightly better than VP-KRR in terms of Test RMSE, but also DC-KRR required much lesser training time than VP-KRR. A reason for this is that Voronoi partitioning tends to produce a very unbalanced clustering. For example, when using Voronoi partitioning to generate 9 clusters, we found that the first cluster had 6484 data points out of total 6553 data points in the dataset, and the remaining clusters had very few data points. Consequently, the training time for the one cluster was almost as huge as the time it would take to train Whole-KRR.

\section{Conclusion}
In this paper, we have provided conditions under which we can give generalization rates (and match minimax rates) for a partitioning based approach to Kernel Ridge Regression. Moreover, we have demonstrated potential statistical advantages as well for such an approach, as it allows for lower approximation error. We hope that this would encourage further investigation into partitioning based extensions of other kernel methods, both from a computational and statistical perspective.

\clearpage
\bibliographystyle{plain}
\bibliography{dc_ridge}

\newpage
\section{Appendix}

This section contains the proofs of all theorems, lemmas and corollaries presented in this paper, as well as some figures and tables. First, we summarize some definitions and notations in the following subsection.

\subsection{Definitions and Notation}

We are given $n$ samples $\mathbf{D}=\left\{(x_1,y_1), \ldots, (x_n,y_n) \right\}$, of the tuple $(x,y)$ drawn i.i.d. from a distribution, $\overline{\P}$, on $\X\times\Y$. $x$ (and $x_i$) is a random vector in the input space $\X$, also called the covariate. $y$ (and $y_i$) is a random variable in the output space $\Y$, also called the response. The collection of sets $\{C_1, \ldots, C_m\}$ is used to denote a disjoint partition of the covariate space: 
\begin{equation}
\X = \cup_{i=1}^{m} C_i \;\text{ and }\; C_i \cap C_j = \{\phi\}, \forall\, i,j \in [m]
\end{equation}

Additionally, we restrict $\Y\subseteq \real$ and assume an additive noise model relating the response to the covariate \emph{i.e.} for each $i\in [n]$:
  \begin{equation}
    y_i = f^*(x_i) + \eta_i.
  \end{equation}
where $f^* :\X \to \real$ is an \textit{unknown} mapping of covariates in $\X$ to responses in $\real$, and $\eta_i$ is the random noise corresponding to sample $i$. We assume that $f^*$ is square integreable with respect to the marginal of $\overline{\P}$ on $\X$. Equivalently, we can say $f^*$ lies in the space $\L_2(\X,\P) = \{f:\X \to \real \,\vert\, \norm{f}{L_2}^2 = \E{f(x)^2} <\infty \}$, where  $\P$ denotes the marginal of $\overline{\P}$ on the input space $\X$. The random noise is assumed to be zero mean with bounded variance \emph{i.e.} $\E{\eta_i \vert x_i} = 0$ and $\E{\eta_i^2\vert x_i} \leq \sigma^2$, $\forall\, i\in [n]$.\\

We are given a \textit{continuous, symmetric, positive definite} kernel $K : \X\times \X \to \real$. For any $x\in \X$, we define $\phi_x := K(x,\cdot)$. Then, the Reproducing Kernel Hilbert Space (RKHS) corresponding to kernel $K$ is given as $\H = \overline{\text{span}}\{\phi_x, \, x\in \X\}$, with inner product defined as 
\begin{align}
\inpdt{\sum_{j}\alpha_j \phi_{x_j}}{\sum_k \beta_k \phi_{x_k}}_{\H} = \sum_{j}\sum_{k} \alpha_j \beta_k K(x_j, x_k)
\end{align}
We require that the RKHS space $\H\subset L_2(\X,\P)$ --- which means $\forall x,\, \mathbb{E}_{y\sim \P}[K(x,y)^2]<\infty$ --- a condition which is always true for several kernel classes, including Gaussian, Laplacian, or any trace class kernel w.r.t. $\P$.\\

The partition based empirical and population covariance operators are defined as (for partition $C_i$):
\begin{align}
\Sigmahat_i &= \frac{1}{n}\sum_{j=1}^{n} (\phi_{x_j}\otimes \phi_{x_j})\indic{x_j\in C_i} \label{eq:sigmahatdef}\\
\Sigma_i &= \E{(\phi_x\otimes\phi_x) \indic{x\in C_i}},
\end{align}
where $\phi_x\otimes \phi_x$ denotes the operator $\phi_x \inpdt{\phi_x}{\cdot}_{\H}$, and $\indic{\cdot}$ denotes the indicator function. Note that we have the relation: \begin{equation}
\Sigma = \sum_{i=1}^{m} \Sigma_i
\end{equation}
where $\Sigma = \E{\phi_x\otimes \phi_x}$, the overall covariance operator.\\

We let $\{\lambda_j^i, v_j^i\}_{j=1, \ldots,\infty}$ be the collection of eigenvalue-eigenfunction pairs for $\Sigma_i$. Then,
\begin{align}
 \Sigma_i = \sum_j \lambda_j^i (v_j^i\otimes v_j^i)
\end{align}

For any $d\in\mathbb{N}, d\geq 1$, we define $P_d$ as the projection operator onto the first $d$ eigenfunctions of $\Sigma_i$. Thus,
\begin{align}
 P_d = \sum_{j=1}^{d} v_j^i\otimes v_j^i
\end{align}
We denote by $\Sigmahat_i^d$ and $\Sigma_i^d$, the projected low-rank empirical and population covariances (with rank $ = d$), obtained using the operator $P_d$. Thus, 
\begin{align}
 \Sigmahat_i^d &= \frac{1}{n}\sum_{j=1}^{n} (P_d\phi_{x_j}\otimes P_d\phi_{x_j})\indic{x_j\in C_i} \label{eq:sigmahatddef}\\
 \Sigma_i^d &= \sum_{j=1}^d \lambda_j^i (v_j^i\otimes v_j^i) \label{eq:sigmaddef}
\end{align}

For any $\lambda > 0$, we define the following spectral sums:
\begin{align}\label{eq:specpartsum}
S_i(\lambda) = \sum_{j=1}^{\infty} \frac{\lambda_j^i}{\lambda_j^i + \lambda},\; U_i(d,\lambda) = \sum_{j=1}^{d} \frac{\lambda_j^i}{\lambda_j^i + \lambda},\; L_i(d,\lambda) = \sum_{j>d} \frac{\lambda_j^i}{\lambda_j^i + \lambda}
\end{align}
Thus, $S_i(\lambda) = U_i(d,\lambda) + L_i(d,\lambda)$, for any $d\in \mathbb{N}$.\\

Finally, we also introduce the shorthand: $\Sigma_{i,\lambda} = (\Sigma_i + \lambda I)$, $\phi_x' = \Sigma_{i,\lambda}^{-1/2}\phi_x$ and $P_d^{\perp} = \sum_{j>d} (v_j^i\otimes v_j^i)$. 

\subsection{Bound on $\E{\norm{(\Sigma_i + \lambda I)^{-1/2}\phi_x}{\H}^{2k}\indic{x\in C_i}}$}\label{subsec:asmpeigfn_interpret}
In this section, we show how Assumption \ref{asmp:eigfn} guarantees a bound on $\E{\norm{(\Sigma_i + \lambda I)^{-1/2}\phi_x}{\H}^{2k}\indic{x\in C_i}}$. Consider any $i\in[m]$. Let us assume that Assumption \ref{asmp:eigfn} holds with parameters $a_1$ and $k (\geq 2)$.

Now, note that for any $x\in \X$, we have:

\begin{align}
\phi_x &= \sum_{j} v_j^i(x) v_j^i\nonumber\\
\Rightarrow\, (\Sigma_i + \lambda I)^{-1/2}\phi_x &= \sum_j \frac{\sqrt{\lambda_j^i}}{\sqrt{\lambda_j^i + \lambda}} \frac{v_j^i(x)}{\sqrt{\lambda_j^i}} v_j^i\nonumber\\
\Rightarrow\, \norm{(\Sigma_i + \lambda I)^{-1/2}\phi_x}{\H}^{2k} &= \left(\sum_j \frac{\lambda_j^i}{\lambda_j^i + \lambda} \frac{v_j^i(x)^2}{\lambda_j^i} v_j^i\right)^k\nonumber\\
&= \left(\sum_k \lambda_k^i/(\lambda_k^i + \lambda)\right)^k\left(\sum_j \frac{\lambda_j^i/(\lambda_j^i + \lambda)}{\sum_k \lambda_k^i/(\lambda_k^i + \lambda)}\frac{v_j^i(x)^2}{\lambda_j^i}\right)^k\nonumber\\
&= S_i(\lambda)^k \left(\sum_j \frac{\lambda_j^i/(\lambda_j^i + \lambda)}{S_i(\lambda)}\frac{v_j^i(x)^2}{\lambda_j^i}\right)^k\nonumber\\
&\overset{(a)}{\leq} S_i(\lambda)^k \left(\sum_j \frac{\lambda_j^i/(\lambda_j^i + \lambda)}{S_i(\lambda)}\left(\frac{v_j^i(x)^2}{\lambda_j^i}\right)^k\right)
\end{align}
where we have $(a)$ using Jensen's inequality.

Thus, we have
\begin{align}
\E{\norm{(\Sigma_i + \lambda I)^{-1/2}\phi_x}{\H}^{2k}\indic{x\in C_i}}&\leq S_i(\lambda)^k \E{\sum_j \frac{\lambda_j^i/(\lambda_j^i + \lambda)}{S_i(\lambda)}\left(\frac{v_j^i(x)^2}{\lambda_j^i}\right)^k}\nonumber\\
&\leq S_i(\lambda)^k a_1^k
\end{align}
where we have used Assumption \ref{asmp:eigfn} in the last step.

\subsection{Moments of the operator norm for Covariance operators}

In this section, we state a lemma providing a bound on the quantity $\E{\norm{\Sigma_{i,\lambda}^{-1/2}(\Sigmahat_i - \Sigma_i)\Sigma_{i,\lambda}^{-1/2}}{}^k}^{1/k}$, for some constant $k\geq 2$. Note that the norm here, $\norm{\cdot}{}$, corresponds to the operator norm. This quantity appears repeatedly in other bounds, and therefore it is useful to have a lemma recording its bound, as stated below. The proof can be found in Section \ref{sec:covbndproof}. First, we introduce the following notion of truncated spectral sums for $\Sigma_i$. For any $d\geq 1$, we let: 
\begin{align}
L_i(d,\lambda) &= \sum_{j=d+1}^{\infty}\frac{\lambda_j^i}{\lambda_j^i + \lambda}\\
U_i(d,\lambda) &= \sum_{j=1}^d \frac{\lambda_j^i}{\lambda_j^i + \lambda}
\end{align}
Note that for any $d\geq 1$, we have: $L_i(d,\lambda) + U_i(d,\lambda) = S_i(\lambda)$, where $S_i(\lambda)$ is defined in Eq. \eqref{eq:specpartsum}.\\

Now, we have the following lemma providing the required bound.

\begin{lemma}\label{lem:expcovbnd}
Consider any $d\in \mathbb{N}, d\geq 1$. Also, let $k \geq 2$ such that Assumption \ref{asmp:eigfn} holds for this $k$ (with constant $a_1$). Then, we have
\begin{align}
 \E{\norm{\Sigma_{i,\lambda}^{-1/2}(\Sigmahat_i - \Sigma_i)\Sigma_{i,\lambda}^{-1/2}}{}^k}^{1/k} \leq \CovErr_{i}(d, \lambda, n, k)
\end{align}
where we have the following expression for $\CovErr_{i}(d,\lambda, n, k)$:
\begin{align}\label{eq:coverrexp}
 \CovErr_{i}(d,\lambda, n, k) &=  a_1 L_i(d,\lambda) + a_1 \sqrt{L_i(d,\lambda) U_i(d,\lambda)} + \frac{a_1\, \sqrt{e \log d} \,\, U_i(d,\lambda)}{\sqrt{n}} \nonumber\\
 & \quad + \frac{4e \log d \,\, \left(a_1 U_i(d,\lambda) + \frac{\lambda_1^i}{\lambda_1^i + \lambda}\right)}{n^{1-1/k}} + \frac{\lambda_{d+1}^i}{\lambda_{d+1}^i + \lambda}
\end{align}
\end{lemma}

Using the above lemma and applying Markov's inequality, we get the following simple corollary.
\begin{corollary}\label{corr:covcorr}
 Consider any $d\in \mathbb{N}, d\geq 1$, and let $k \geq 2$ such that Assumption \ref{asmp:eigfn} holds for this $k$ (with constant $a_1$). Then, we have
 \begin{align}
  \prob\left(\norm{\Sigma_{i,\lambda}^{-1/2}(\Sigmahat_i - \Sigma_i)\Sigma_{i,\lambda}^{-1/2}}{} \geq \frac{1}{2} \right))\leq 2^{k} \left[\CovErr_{i}(d,\lambda, n)\right]^k
 \end{align}

\end{corollary}

\subsubsection{Bounds on $\CovErr_{i}(d, \lambda p_i, n, k)$ for specific cases}
\label{subsubsec:coverrcases}
While the expression in Eq. \ref{eq:coverrexp} may seem complicated, it is possible to obtain concrete expressions for specific kernels through an appropriate choice of $d$, similar to the approach in \cite{zhang13}. The idea is to choose a $d$ which makes  the $L_i(d, \lambda p_i)$ terms negligible in Eq. \ref{eq:coverrexp}. We do this for a few cases below.\\

\textbf{Finite Rank Kernels}. Suppose kernel $K$ has finite rank $r$ --- examples include the linear and polynomial kernels. Then, for any $i\in[m]$, the partitionwise covariance operator $\Sigma_i$ is also finite rank. Thus, we can pick $d=r$ (in Eq. \ref{eq:coverrexp}), which gives $L_i(d,\lambda p_i) = 0$ and $\lambda_{d+1}^i = 0$. Also, $U_i(d, \lambda p_i) = S_i(\lambda p_i)\leq r$. Plugging these into Eq. \ref{eq:coverrexp}, we get:
\begin{align}
 \CovErr_{i}(r,\lambda p_i, n) &= O\left(\frac{\sqrt{\log r}\, S_{i}(\lambda p_i)}{\sqrt{n}}\right) = O\left(\frac{r \sqrt{\log r}}{\sqrt{n}}\right)
\end{align}

\textbf{Kernels with polynomial decay in eigenvalues}. Suppose kernel $K$ has polynomially decaying eigenvalues, $\lambda_j \leq c j^{-v}\; (\forall j$, and constants $c>0, v>2$) --- examples here include sobolev kernels with different orders. Now, since we have $\Sigma = \sum_{i=1}^m \Sigma_i$ being a sum of psd operators, the minimax characterization of eigenvalues yields: $\lambda_j^i \leq \lambda_j$ $\forall j$ and any $i\in [m]$. As a consequence, we have: $L_i(d,\lambda)\leq \sum_{j>d}\frac{\lambda_j}{\lambda_j + \lambda}$ and $S_{i}(\lambda)\leq S(\lambda)$. Then, following the same approach as \cite{zhang13} i.e. choosing $d = n^{C/(v-1)}$ for some constant $C > 0$, we get:
\begin{align}
L_i(d, \lambda p_i) \leq \int_{d}^{\infty} \frac{c j^{-v}}{c j^{-v} + \lambda p_i}dj \leq \frac{c}{\lambda p_i} \int_{d}^{\infty} j^{-v}dj \leq \frac{c (v-1)}{\lambda p_i} d^{-(v-1)} \leq \frac{c (v-1)}{\lambda p_i n^C}
\end{align} and, $U_i(d, \lambda p_i) \leq d = n^{C/(v-1)}$. Consequently, for $v>2$ and $\lambda p_i \geq \frac{1}{n^{C\frac{v}{v-1} - 1}}$, we get:
\begin{align}
 \CovErr_{i}(n^{C/(v-1)},\lambda p_i, n) &= O\left(\frac{\sqrt{\log n}}{n^{\frac{1}{2} - \frac{C}{v-1}}}\right)
\end{align}

\textbf{Kernels with exponential decay in eigenvalues}. Suppose kernel $K$ has exponentially decaying eigenvalues, $\lambda_j \leq c_1 \exp(- c_2 j^2)\; (\forall j$, and constants $c_1, c_2 >0$) --- an example here is the Gaussian kernel. Again, since $\Sigma = \sum_{i=1}^m \Sigma_i$, the minimax characterization of eigenvalues yields: $\lambda_j^i \leq \lambda_j$ $\forall j$ and any $i\in [m]$. Thus: $L_i(d,\lambda)\leq \sum_{j>d}\frac{\lambda_j}{\lambda_j + \lambda}$ and $S_{i}(\lambda)\leq S(\lambda)$. Choosing $d = C\sqrt{\log n}/\sqrt{c_2}$ for some constant $C$, we get:
\begin{align}
L_i(d, \lambda p_i) \leq \int_{d}^{\infty} \frac{c_1 \exp(- c_2 j^2)}{c_1 \exp(- c_2 j^2) + \lambda p_i}dj \leq \frac{c_1}{\lambda p_i} \int_{d}^{\infty}\exp(- c_2 j^2)dj \leq \frac{c_1}{\lambda p_i} \exp(- c_2 d^2) \leq \frac{c_1}{\lambda p_i n^C}
\end{align} and, $U_i(d, \lambda p_i) \leq d = C\sqrt{\log n}/\sqrt{c_2}$. Consequently, as long as $\lambda p_i \geq \poly(1/n)$, we can choose a sufficiently large $C$ to make the terms involving $\lambda_{d+1}^i$ and $L_i(d, \lambda p_i)$ negligible. Thus, we get:
\begin{align}
 \CovErr_{i}(C\sqrt{\log n},\lambda p_i, n) &= O\left(\frac{\sqrt{\log n\, (\log \log n)}}{\sqrt{n}}\right)
\end{align}

\subsection{Proof of Lemma \ref{lem:errdecomp}}
The proof is as follows:
\begin{align}
\Err_{i}(\fhat_{i,\lambda}) = \E{(f^*(x) - \fhat_{i,\lambda}(x))^2} &= \E{\left(f^*(x) - f_{i,\lambdabar}(x) + f_{i,\lambdabar}(x) - f_{i,\lambda}(x) + f_{i,\lambda}(x) - \fhat_{i,\lambda}(x)\right)^2}\nonumber\\
& \overset{(a)}{\leq} 2\left(\App_{i}(\lambdabar) + 2 \Reg_i(\lambdabar,\lambda) + 2 \E{\left(f_{i,\lambda}(x) - \fhat_{i,\lambda}(x)\right)^2}\right)
\end{align}
where we have $(a)$ since $(a + b + c)^2 \leq 2(a^2 + 2 b^2 + 2 c^2)$.\\

Now, following a standard bias-variance decomposition, we have:
\begin{align}
\Ed{D}{\E{\left(f_{i,\lambda}(x) - \fhat_{i,\lambda}(x)\right)^2}} &=  \E{\left(f_{i,\lambda}(x) - \fbar_{i,\lambda}(x)\right)^2} + \Ed{D}{\E{\left(f_{i,\lambda}(x) - \fhat_{i,\lambda}(x)\right)^2}}\nonumber \\
&= \Bias_i(\lambda, n) + \Ed{D}{\Var_{i}(\lambda,D)}
\end{align}

Combining the above expressions, we get:
\begin{align}
\Ed{D}{\Err_i(\fhat_{i,\lambda})} \leq 2\left[ \App_{i}(\lambdabar) + 2 \Reg_i(\lambdabar,\lambda) + 2 \Bias_i(\lambda, n) + 2 \Ed{D}{\Var_{i}(\lambda,D)} \right]
\end{align}

\subsection{Proof of Theorems \ref{thm:finrank} and \ref{thm:eigdec}}

The theorems are a simple consequence of combining Lemmas \ref{lem:mainregbnd}, \ref{lem:mainbiasbnd}, and \ref{lem:mainvarbnd} via Lemma \ref{lem:errdecomp}, plugging $A_i(\lambdabar)=0$ with $\lambdabar=0$, ignoring the bias terms which are of a lower order, and using the expressions for $CE_i = \CovErr_i(\lambda p_i, n, k)$ discussed in Section \ref{subsec:covbnd}.  

\subsection{Proof of Theorem \ref{thm:thm3}}
Consider any $\lambdabar > 0$, and let $f_{\lambdabar}$ be the solution of Eq. \eqref{eq:overallbest}. Now, for any partition $i\in [m]$, consider the following optimization problem:
\begin{align}
\fhat_i = \argmin_{f\in \H, \norm{f}{\H} \leq \norm{f_{\lambdabar}}{\H}} \E{(f^*(x) - f(x))^2\indic{x\in C_i}}
\end{align}
By duality, $\exists$ $\lambdabar_i'\geq 0$ s.t. $\fhat_i = f_{i, \lambdabar_i'}$, with $f_{i, \lambdabar_i'}$ being the solution of Eq. \eqref{eq:fbestopt1}. Now, by the optimality of $f_{i, \lambdabar_i'}$, we have:
\begin{align}
\App_i(\lambdabar_i') = \E{(f^*(x) - f_{i, \lambdabar_i'}(x))^2\indic{x\in C_i}} \leq \E{(f^*(x) - f_{\lambdabar}(x))^2\indic{x\in C_i}} = \AppErr_i(f_{\lambdabar})
\end{align}
and $\norm{f_{i,\lambdabar_i'}}{\H}\leq \norm{f_{\lambdabar}}{\H}$. 

Now, if $\lambdabar_i'\leq \lambdabar$, we are done. Suppose $\lambdabar_i' > \lambdabar$. Then, we know that $\App_i(\lambdabar)\leq \AppErr(f_{\lambdabar})$, since decreasing the regularization penalty from $\lambdabar_i'$ to $\lambdabar$ would only decrease the approximation error. Moreover, using the fact that the following function
\begin{align}
T(\lambda) := \min_{f\in \H} \E{(f^*(x) - f(x))^2\indic{x\in C_i}} + \lambda p_i \norm{f}{\H}^2
\end{align}
is a monotonically increasing function of $\lambda$\cite{steinwartbook}, we have:
\begin{align}
\App_i(\lambdabar) + \lambda p_i \norm{f_{i, \lambdabar}}{\H}^2 &\leq \App_i(\lambdabar_i') + \lambda p_i \norm{f_{i, \lambdabar_i'}}{\H}^2\\
&\leq \AppErr_i(f_{\lambdabar}) + \lambda p_i \norm{f_{\lambdabar}}{\H}^2
\end{align}

Thus, $\norm{f_{i, \lambdabar}}{\H} = O\left(\norm{f_{\lambdabar}}{\H} + \sqrt{\frac{\AppErr_i(f_{\lambdabar})}{\lambdabar p_i}}\right)$. Therefore, the result holds with $\lambdabar_i = min(\lambdabar, \lambdabar_i')$.

The bound on the estimation error $\mathcal{E}_C$ is a simple consequence of the fact that, under the conditions assumed, all terms in Lemma \ref{lem:mainbiasbnd} and Lemma \ref{lem:mainvarbnd} involving $(CE)_i^k$ are of a lower order, and that the condition $\App(\lambdabar) = O(\lambdabar\norm{f_{\lambdabar}}{\H}^2)$ guarantees that:
\begin{align}
\sum_{i} p_i \norm{f_{i, \lambdabar_i}}{\H}^2 = O( \norm{f_{\lambdabar}}{\H}^2)
\end{align}

Then, combining Lemma \ref{lem:mainregbnd}, Lemma \ref{lem:mainbiasbnd} and Lemma \ref{lem:mainvarbnd} via Lemma \ref{lem:errdecomp} gives us the required scaling.

\subsection{Regularization Bound}

In this section we provide a proof of Lemma \ref{lem:mainregbnd}. The lemma is restated below for convenience.
\begin{lemma*}
For any $\lambda > 0, \overline{\lambda} > 0$ and partition $i\in [m]$,
 \begin{align}
  \Reg_i(\lambda,\overline{\lambda}) = \E{(f_{i, \overline{\lambda}}(x) - f_{i,\lambda}(x))^2\indic{x\in C_i}} \leq p_i \frac{(\overline{\lambda}-\lambda)^2}{\lambda}  \norm{f_{i,\overline{\lambda}}}{\H}^2
  \end{align}
\end{lemma*}

\subsubsection{Proof of Lemma \ref{lem:mainregbnd}}
\begin{proof}
We want to bound
\begin{align}
\E{(f_{i, \overline{\lambda}}(x) - f_{i,\lambda}(x))^2\indic{x\in C_i}}  &= \norm{f_{i,\lambdabar} - f_{i,\lambda}}{\Sigma_i}^2\nonumber\\
& = \norm{\Sigma_i^{1/2}(f_{i,\lambdabar} - f_{i,\lambda})}{\H}^2
\end{align}
Using first order conditions for the optimality of $f_{i,\lambda}$ and $f_{i,\overline{\lambda}}$, we have
\begin{align}
(\Sigma_i + \lambda p_i I)f_{i,\lambda} = \E{y\phi_x \indic{x\in C_i}}\nonumber\\
(\Sigma_i + \lambdabar p_i I)f_{i,\lambdabar} = \E{y\phi_x \indic{x\in C_i}}
\end{align}
Thus, $f_{i,\lambda} = (\Sigma_i + \lambda p_i I)^{-1} (\Sigma_i + \lambdabar p_i I)f_{i,\lambdabar}$. 

Letting $f_{i,\lambdabar} = \sum_j \alpha_j v_j^i$, we get
\begin{align}
\Sigma_i^{1/2}(f_{i,\lambdabar} - f_{i,\lambda}) &= p_i(\lambda - \lambdabar) \sum_{j}  \frac{\sqrt{\lambda_j^i}}{\lambda_j^i + \lambda p_i} \alpha_j v_j^i\nonumber\\
\Rightarrow\; \norm{\Sigma_i^{1/2}(f_{i,\lambdabar} - f_{i,\lambda})}{\H}^2 &= p_i^2(\lambda - \lambdabar)^2 \sum_{j} \frac{\lambda_j^i}{(\lambda_j^i+\lambda p_i)^2} \alpha_j^2\nonumber\\
&\leq p_i \frac{(\lambda - \lambdabar)^2}{\lambda} \sum_{j} \alpha_j^2 = p_i \frac{(\lambda - \lambdabar)^2}{\lambda} \norm{f_{i,\lambdabar}}{\H}^2
\end{align}
\end{proof}

\subsection{Bias Bound}
In this section we provide a proof of Lemma \ref{lem:mainbiasbnd}. The lemma is restated below.

\begin{lemma*}
Consider any $d\in \mathbb{N}, d\geq 1$, and $k\geq 2$. Suppose Assumption \ref{asmp:eigfn} holds for this $k$ (with constant $a_1$), and Assumption \ref{asmp:approx} holds. Also, suppose $\forall i\in [m]$, $p_i$ satisfies: $p_i \geq \frac{16\log (np_i)}{n-1}$. Then we have
\begin{align}
\Bias_i(\lambda,n) \leq (CovErr_{i}(d,\lambda p_i, n))^2\left(  T_1  + T_2  + 2^{k+1}\left[\CovErr_{i}(d,\lambda p_i, n)\right]^k T_3 + 2^{k/2 + 3}\left[\CovErr_{i}(d,\lambda p_i, n)\right]^{k/2} T_4\right) 
\end{align}
where we let
\begin{align}
T_1 &= \frac{16 a_1 \sqrt{p_i}  S_{i}(\lambda p_i) A_{i}(\lambdabar)^2}{n}\nonumber\\
T_2 & = \left(\frac{16 a_1^2 (\lambda - \lambdabar)^2}{\lambda}\, \frac{ p_i S_i(\lambda p_i)^2 \norm{f_{i,\lambdabar}}{\H}^2}{n} + \frac{8 \lambda p_i  \norm{f_{i,\lambdabar}}{\H}^2}{n} \right)\nonumber\\
T_3 & = \left(2\norm{f_{i,\lambdabar}}{\H}^2 + \frac{\sigma^2}{\lambda}\right)(\lambda_1^i + \lambda p_i)\nonumber\\
T_4 & = \frac{(\lambda_1^i + \lambda p_i) A_i(\lambdabar)^2}{\lambda \sqrt{p_i}}
\end{align}	
\end{lemma*}
\subsubsection{Proof of Lemma \ref{lem:mainbiasbnd}}
\begin{proof}
We want to bound $\Bias_i(\lambda,n)$, where
\begin{align}
\Bias_i(\lambda,n) &= \E{(f_{i,\lambda}(x) - \fbar_{i,\lambda}(x))^2\indic{x\in C_i}}\nonumber\\
&= \E{\left(\inpdt{f_{i,\lambda} - \fbar_{i,\lambda}}{\phi_x}_{\H}\right)^2\indic{x\in C_i}}\nonumber\\
&= \norm{f_{i,\lambda} - \fbar_{i,\lambda}}{\Sigma_i}^2
\end{align}
Let $\Delta_b = f_{i,\lambda} - \fhat_{i, \lambda}$. Then, equivalently, we want to bound $\norm{\E{\Delta_b}}{\Sigma_i}^2$

Now, from first order conditions of optimality for Eqs. \eqref{eq:fhatopt} and \eqref{eq:fbestopt2}, we have
\begin{align}\label{eq:biasfirstorder}
(\Sigmahat_i + \lambda p_i I)\fhat_{i,\lambda} &= \Eemp{y \phi_x \indic{x\in C_i}}\nonumber\\
(\Sigma_i + \lambda p_i I)f_{i,\lambda} &= \E{y\phi_x \indic{x \in C_i}}
\end{align}
Combining the above, we get
\begin{align}
\E{(\Sigmahat_i + \lambda p_i I)\Delta_b} &= \E{(\Sigmahat_i + \lambda p_i I)f_{i,\lambda}} - \E{(\Sigmahat_i + \lambda p_i I)\fhat_{i,\lambda}}\nonumber\\ 
&= (\Sigma_i + \lambda p_i I)f_{i,\lambda} - \E{y \phi_x \indic{x\in C_i}}\nonumber\\
& = 0
\end{align}
Rearranging and multiplying $\Sigma_{i,\lambda p_i}^{-1/2}$, we get
\begin{align}
\Sigma_{i, \lambda p_i}^{1/2}\E{\Delta_b} &= -\E{\Sigma_{i, \lambda p_i}^{-1/2}(\Sigmahat_i - \Sigma_i)\Sigma_{i,\lambda p_i}^{-1/2} \Sigma_{i,\lambda p_i}^{1/2}\Delta_b}\nonumber\\
& = -\E{\Sigma_{i, \lambda p_i}^{-1/2}(\Sigmahat_i - \Sigma_i)\Sigma_{i,\lambda p_i}^{-1/2} \Sigma_{i,\lambda p_i}^{1/2}\E{\Delta_b\,\vert\,X}}\nonumber\\
\end{align}
where we let $X$ denote the set $\{x_1, \ldots, x_n\}$ \emph{i.e.} the covariates in the data $\mathbf{D}$.\\

So, 
\begin{align}
\norm{\Sigma_{i, \lambda p_i}^{1/2}\E{\Delta_b}}{\H}^2 &= \norm{\E{\Sigma_{i, \lambda p_i}^{-1/2}(\Sigmahat_i - \Sigma_i)\Sigma_{i,\lambda p_i}^{-1/2} \Sigma_{i,\lambda p_i}^{1/2}\E{\Delta_b\,\vert\,X}}}{\H}^2\nonumber\\
\Rightarrow \norm{\Sigma_{i}^{1/2}\E{\Delta_b}}{\H}^2 \overset{(a)}{\leq} \norm{\Sigma_{i, \lambda p_i}^{1/2}\E{\Delta_b}}{\H}^2 &= \norm{\E{\Sigma_{i, \lambda p_i}^{-1/2}(\Sigmahat_i - \Sigma_i)\Sigma_{i,\lambda p_i}^{-1/2} \Sigma_{i,\lambda p_i}^{1/2}\E{\Delta_b\,\vert\,X}}}{\H}^2\nonumber\\
& \overset{(b)}{\leq} \left(\E{\norm{\Sigma_{i, \lambda p_i}^{-1/2}(\Sigmahat_i - \Sigma_i)\Sigma_{i,\lambda p_i}^{-1/2} \Sigma_{i,\lambda p_i}^{1/2}\E{\Delta_b\,\vert\,X}}{\H}}\right)^2\nonumber\\
& \overset{(c)}{\leq} \left(\E{\norm{\Sigma_{i, \lambda p_i}^{-1/2}(\Sigmahat_i - \Sigma_i)\Sigma_{i,\lambda p_i}^{-1/2}}{\H}\norm{\Sigma_{i,\lambda p_i}^{1/2}\E{\Delta_b\,\vert\,X}}{\H}}\right)^2\nonumber\\
& \overset{(d)}{\leq} \E{\norm{\Sigma_{i, \lambda p_i}^{-1/2}(\Sigmahat_i - \Sigma_i)\Sigma_{i,\lambda p_i}^{-1/2}}{\H}^2}\E{\norm{\Sigma_{i,\lambda p_i}^{1/2}\E{\Delta_b\,\vert\,X}}{\H}^2}
\end{align}
where we have $(a)$ using the fact that $\inpdt{u}{\Sigma_i u}_{\H} < \inpdt{u}{(\Sigma_i + \lambda p_i I) u}_{\H}$ $\forall u\in \H$, $(b)$ by Jensen's inequality, $(c)$ by the definition of the operator norm, $(d)$ by the Cauchy-Schwarz inequality.\\

Thus,
\begin{align}\label{eq:biasbndorig}
\norm{\E{\Delta_b}}{\Sigma_i}^2 \leq \E{\norm{\Sigma_{i, \lambda p_i}^{-1/2}(\Sigmahat_i - \Sigma_i)\Sigma_{i,\lambda p_i}^{-1/2}}{\H}^2}\E{\norm{\Sigma_{i,\lambda p_i}^{1/2}\E{\Delta_b\,\vert\,X}}{\H}^2}
\end{align}
Now, Lemma \ref{lem:expcovbnd} provides a bound for $\E{\norm{\Sigma_{i, \lambda p_i}^{-1/2}(\Sigmahat_i - \Sigma_i)\Sigma_{i,\lambda p_i}^{-1/2}}{\H}^2}$. For the remainder of the proof, we provide the bound for $\E{\norm{\Sigma_{i,\lambda p_i}^{1/2}\E{\Delta_b\,\vert\,X}}{\H}^2}$. Combining these bounds will yield the main statement of the lemma.\\

From first order conditions again (Eq. \eqref{eq:biasfirstorder}), we have
\begin{align}
(\Sigmahat_i  +\lambda p_i I)\E{\Delta_b\,\vert\,X} = \Eemp{f^*(x)\phi_x\indic{x\in C_i}} - (\Sigmahat_i + \lambda p_i)f_{i,\lambda}
\end{align}

Multiplying by $\Sigma_{i, \lambda p_i}^{-1/2}$ on both sides and rewriting differently, we get
\begin{align}
\left(\Sigma_{i,\lambda p_i}^{-1/2}(\Sigmahat_i - \Sigma_i)\Sigma_{i,\lambda p_i}^{-1/2} + I\right)\Sigma_{i,\lambda p_i}^{1/2}\E{\Delta_b\,\vert\,X} &= \Sigma_{i,\lambda p_i}^{-1/2}\left(\Eemp{f^*(x)\phi_x\indic{x\in C_i}} - (\Sigmahat_i + \lambda p_i)f_{i,\lambda}\right)\nonumber\\
& = \left(\Eemp{(f^*(x)-f_{i,\lambda}(x))\Sigma_{i,\lambda p_i}^{-1/2} \phi_x\indic{x\in C_i}} - \lambda p_i \Sigma_{i,\lambda p_i}^{-1/2} f_{i,\lambda}\right)\nonumber\\
\Rightarrow\; \norm{\left(\Sigma_{i,\lambda p_i}^{-1/2}(\Sigmahat_i - \Sigma_i)\Sigma_{i,\lambda p_i}^{-1/2} + I\right)\Sigma_{i,\lambda p_i}^{1/2}\E{\Delta_b\,\vert\,X}}{\H}^2 &= \norm{\Eemp{(f^*(x)-f_{i,\lambda}(x))\Sigma_{i,\lambda p_i}^{-1/2} \phi_x\indic{x\in C_i}}- \lambda p_i \Sigma_{i,\lambda p_i}^{-1/2} f_{i,\lambda}}{\H}^2\nonumber\\
& = \norm{\frac{1}{n}\sum_{j=1}^{n} w_j}{\H}^2
\end{align}
where we define $w_j := (f^*(x_j) -f_{i,\lambda}(x_j))\Sigma_{i,\lambda p_i}^{-1/2} \phi_{x_j}\indic{x_j\in C_i}- \lambda p_i \Sigma_{i,\lambda p_i}^{-1/2} f_{i,\lambda}$. Note that $\E{w_j} = 0$.\\

Let us define the event $\mathcal{E}_{cov} = \left\{\norm{\Sigma_{i,\lambda p_i}^{-1/2}(\Sigmahat_i - \Sigma_i)\Sigma_{i,\lambda p_i}^{-1/2}}{}\leq 1/2\right\}$. Note that from Corollary \ref{corr:covcorr}, we have $\prob\left(\mathcal{E}_{cov}^c\right)\leq 2^k \left[\CovErr_{i}(d,\lambda p_i, n)\right]^k$.  Now, under the event $\mathcal{E}_{cov}$,
\begin{align}
\E{\norm{\Sigma_{i,\lambda p_i}^{1/2} \E{\Delta_b\,\vert\,X}}{\H}^2} &\leq 4 \E{\norm{\frac{1}{n}\sum_{j=1}^{n} w_j}{\H}^2}\nonumber\\
&= \frac{4}{n^2}\sum_{j=1}^n \E{\norm{w_j}{\H}^2}
\end{align}

To control $\E{\norm{\Sigma_{i,\lambda p_i}^{1/2} \E{\Delta_b\,\vert\,X}}{\H}^2}$ overall, we have
\begin{align}\label{eq:biasoverallb1}
\E{\norm{\Sigma_{i,\lambda p_i}^{1/2} \E{\Delta_b\,\vert\,X}}{\H}^2} & = \E{\norm{\Sigma_{i,\lambda p_i}^{1/2} \E{\Delta_b\,\vert\,X}}{\H}^2 \indic{\mathcal{E}_{cov}}} + \E{\norm{\Sigma_{i,\lambda p_i}^{1/2} \E{\Delta_b\,\vert\,X}}{\H}^2 \indic{\mathcal{E}_{cov}^c}}\nonumber\\
& \leq \frac{4}{n^2}\sum_{j=1}^n \E{\norm{w_j}{\H}^2} + (\lambda_1^i + \lambda p_i)\E{\E{\norm{\Delta_b}{\H}^2\,\vert\, X}\indic{\mathcal{E}_{cov}^c}}
\end{align}

\textbf{Bound on $\E{\norm{w_j}{\H}^2}$}.
We have
\begin{align}
\E{\norm{w_j}{\H}^2} &\overset{(a)}{\leq} 2 \E{(f^*(x_j) -f_{i,\lambda}(x_j))^2\norm{\Sigma_{i,\lambda p_i}^{-1/2} \phi_{x_j}}{\H}^2\indic{x_j\in C_i})} + 2(\lambda p_i)^2\norm{\Sigma_{i,\lambda p_i}^{-1/2} f_{i,\lambda}}{\H}^2\nonumber\\
& \overset{(b)}{\leq} 4 \E{(f^*(x_j) -f_{i,\overline{\lambda}}(x_j))^2\norm{\Sigma_{i,\lambda p_i}^{-1/2} \phi_{x_j}}{\H}^2\indic{x_j\in C_i})}\nonumber\\ &\quad + 4 \E{(f_{i,\overline{\lambda}}(x_j) -f_{i,\lambda}(x_j))^2\norm{\Sigma_{i,\lambda p_i}^{-1/2} \phi_{x_j}}{\H}^2\indic{x_j\in C_i})} + 2(\lambda p_i)^2\norm{\Sigma_{i,\lambda p_i}^{-1/2} f_{i,\lambda}}{\H}^2\nonumber\\
& \overset{(c)}{\leq} 4\sqrt{\E{(f^*(x_j) -f_{i,\overline{\lambda}}(x_j))^4\indic{x_j\in C_i}}}\sqrt{\E{\norm{\Sigma_{i,\lambda p_i}^{-1/2} \phi_{x_j}}{\H}^4\indic{x_j\in C_i})}}\nonumber\\
&\quad + 4 \norm{f_{i,\overline{\lambda}} -f_{i,\lambda}}{\Sigma_{i,\lambda p_i}}^2\E{\norm{\Sigma_{i,\lambda p_i}^{-1/2} \phi_{x_j}}{\H}^4\indic{x_j\in C_i})} + 2(\lambda p_i)^2\norm{\Sigma_{i,\lambda p_i}^{-1/2} f_{i,\lambda}}{\H}^2\nonumber\\
&\overset{(d)}{\leq} 4 a_1 \sqrt{p_i} A_{i}(\lambdabar)^2 S_i(\lambda p_i) + 4 a_1^2 S_i(\lambda p_i)^2 \norm{f_{i,\overline{\lambda}} -f_{i,\lambda}}{\Sigma_{i,\lambda p_i}}^2  + 2(\lambda p_i)^2\norm{\Sigma_{i,\lambda p_i}^{-1/2} f_{i,\lambda}}{\H}^2\nonumber\\
&\overset{(e)}{\leq} 4 a_1 \sqrt{p_i} A_{i}(\lambdabar)^2 S_i(\lambda p_i) + 4 a_1^2 p_i \frac{(\lambda - \lambdabar)^2}{\lambda} S_i(\lambda p_i)^2 \norm{f_{i,\lambdabar}}{\H}^2 + 2 \lambda p_i \norm{f_{i,\overline{\lambda}}}{\H}^2\nonumber\\
& = \left[4 a_1 \sqrt{p_i} A_{i}(\lambdabar)^2\right] S_{i}(\lambda p_i) + \left[4 a_1^2 p_i \frac{(\lambda - \lambdabar)^2}{\lambda} S_i(\lambda p_i)^2 + 2\lambda p_i \right]\norm{f_{i,\lambdabar}}{\H}^2
\end{align}
where we have $(a)$ using $\norm{x + y}{\H}^2\leq 2\norm{x}{\H}^2 + 2\norm{y}{\H}^2$, $(b)$ since $(f^*(x_j) -f_{i,\lambda}(x_j))^2\leq 2(f^*(x_j) -f_{i,\overline{\lambda}}(x_j))^2 + 2(f_{i,\overline{\lambda}}(x_j) -f_{i,\lambda}(x_j))^2$, $(c)$ using Cauchy-Schwarz inequality in two different ways, namely, $\E{XY}\leq \sqrt{\E{X^2}}\sqrt{\E{Y^2}}$ and $(f_{i,\overline{\lambda}}(x_j) -f_{i,\lambda}(x_j))^2 = \left(\inpdt{f_{i,\overline{\lambda}} - f_{i,\lambda}}{\phi_{x_j}}_{\H}\right)^2 \leq \norm{f_{i,\overline{\lambda}} - f_{i,\lambda}}{\Sigma_{i,\lambda p_i}}^2 \norm{\Sigma_{i,\lambda p_i}^{-1/2}\phi_{x_j}}{\H}^2$, $(d)$ using Assumption \ref{asmp:approx}, and via Jensen's inequality and Assumption \ref{asmp:eigfn}
\begin{align}
\E{\norm{\Sigma_{i,\lambda p_i}^{-1/2} \phi_{x_j}}{\H}^4\indic{x_j\in C_i})} &= \E{\left(\sum_j \frac{\lambda_j^i}{\lambda_j^i + \lambda p_i} \frac{v_j^i(x)^2}{\lambda_j^i} \indic{x_j\in C_i}\right)^2}\nonumber\\
&\leq S_i(\lambda p_i)^2 \sum_j \frac{\lambda_j^i/(\lambda_j^i + \lambda p_i)}{\sum_k \lambda_k^i/(\lambda_k^i + \lambda p_i)} \E{\frac{v_j^i(x)^4}{{\lambda_j^i}^2}\indic{x_j\in C_i}}\nonumber\\
&= a_1^2 S_i(\lambda p_i)^2,
\end{align} $(e)$ using the relation $f_{i, \lambda} = \Sigma_{i,\lambda p_i}^{-1}\Sigma_{i,\lambdabar p_i} f_{i,\lambdabar}$.\\

\textbf{Bound on $\E{\norm{\Delta_b}{\H}^2\,\vert\, \{x_1,\ldots x_n\}}$}.
We have
\begin{align}
\E{\norm{\Delta_b}{\H}^2\,\vert\, \{x_1,\ldots x_n\}} &\overset{(a)}{\leq} 2\norm{f_{i,\lambda}}{\H}^2 + 2\E{\norm{\fhat_{i,\lambda}}{\H}^2\,\vert\, \{x_1\ldots x_n\}}\nonumber\\
&\overset{(b)}{\leq} 4\norm{f_{i,\lambdabar}}{\H}^2 + \frac{2}{\lambda}\frac{1}{n_i}\sum_{j=1}^n (f^*(x_j) - f_{i,\lambdabar}(x_j))^2\indic{x_j\in C_i} + \frac{2\sigma^2}{\lambda}
\end{align}
where we have $(a)$ using $\norm{x + y}{\H}^2 \leq 2\norm{x}{\H}^2 + 2\norm{y}{\H}^2$, $(b)$ using optimality of $\fhat_{i,\lambda}$ for the loss function in Eq. \eqref{eq:partestopt}.

\textbf{Overall Bound}. Combining the above bounds with the terms in Eq. \eqref{eq:biasoverallb1}, we have
\begin{align}\label{eq:biasp1}
\frac{4}{n^2}\sum_{j=1}^n \E{\norm{w_j}{\H}^2} \leq \frac{4 S_{i}(\lambda p_i)}{n}\left[4 a_1 \sqrt{p_i} A_{i}(\lambdabar)^2\right]  + \frac{4 \norm{f_{i,\lambdabar}}{\H}^2}{n}\left[4 a_1^2 p_i \frac{(\lambda - \lambdabar)^2}{\lambda} S_i(\lambda p_i)^2 + 2\lambda p_i \right]
\end{align}
and
\begin{align}
\E{\E{\norm{\Delta_b}{\H}^2\,\vert\, x_1,\ldots x_n}\indic{\mathcal{E}_{cov}^c}} &\leq \left(4\norm{f_{i,\lambdabar}}{\H}^2 + \frac{2\sigma^2}{\lambda}\right)\prob(\mathcal{E}_{cov}^c)\nonumber\\
&\quad + \frac{2}{\lambda}\E{\frac{1}{n_i}\sum_{j=1}^n (f^*(x_j) - f_{i,\lambdabar}(x_j))^2\indic{x_j\in C_i}  \indic{\mathcal{E}_{cov}^c}}
\end{align}

Now,
\begin{align}\label{eq:biasbinseq}
\E{\frac{1}{n_i}(f^*(x_j) - f_{i,\lambdabar}(x_j))^2\indic{x_j\in C_i}  \indic{\mathcal{E}_{cov}^c}} &\overset{(a)}{\leq} \sqrt{\E{\frac{1}{n_i^2}(f^*(x_j) - f_{i,\lambdabar}(x_j))^4\indic{x_j\in C_i}}}\sqrt{\prob(\mathcal{E}_{cov}^c)}\nonumber\\
&\overset{(b)}{=} \sqrt{\prob(\mathcal{E}_{cov}^c)}\sqrt{p_i}\sqrt{\E{(f^*(x_j) - f_{i,\lambdabar}(x_j))^4\,\vert\, x_j \in C_i} \E{\frac{1}{(1 + Y)^2}}}\nonumber\\
&\overset{(c)}{\leq} \sqrt{\prob(\mathcal{E}_{cov}^c)}\sqrt{p_i}A_i(\lambdabar)^2\sqrt{\left(\exp(-(n-1)p_i/8) + \frac{4}{((n-1)p_i)^2}\right)}\nonumber\\
&\overset{(d)}{\leq} \frac{4}{n\sqrt{p_i}}\sqrt{\prob(\mathcal{E}_{cov}^c)}A_i(\lambdabar)^2
\end{align}
where we have $(a)$ using Cauchy-Schwarz, $(b)$ using $n_i = \sum_{j=1}^n \indic{x_j\in C_i}$, independence of $x_1, \ldots, x_n$, and letting $Y\sim Bin(n-1,p_i)$, $(c)$ using Assumption \ref{asmp:approx} and $\E{\frac{1}{(1+Y)^2}}\leq \exp(-np/8) + \frac{4}{(np)^2}$ for $Y\sim Bin(n,p)$ with $p\leq 1/2$, $(d)$ using $p_i\geq \frac{16 \log(n p_i/2)}{n-1}$.\\

Consequently, we have
\begin{align}\label{eq:biasp2}
\E{\E{\norm{\Delta_b}{\H}^2\,\vert\, x_1,\ldots x_n}\indic{\mathcal{E}_{cov}^c}} \leq \left(4\norm{f_{i,\lambdabar}}{\H}^2 + \frac{2\sigma^2}{\lambda}\right)\prob(\mathcal{E}_{cov}^c) + 8 \sqrt{\prob(\mathcal{E}_{cov}^c)}\frac{A_i(\lambdabar)^2}{\lambda \sqrt{p_i}}
\end{align}

Finally, plugging Eqs. \eqref{eq:biasp1} and \eqref{eq:biasp2} into Eq. \eqref{eq:biasoverallb1} followed by Eq. \eqref{eq:biasbndorig}, we have the bias bound
\begin{align}
&\norm{\E{\Delta_b}}{\Sigma_i}^2 \leq \nonumber\\
&(CovErr_{i}(d,\lambda p_i, n))^2\left(  T_1  + T_2  + 2^{k+1}\left[\CovErr_{i}(d,\lambda p_i, n)\right]^k T_3 + 2^{k/2 + 3}\left[\CovErr_{i}(d,\lambda p_i, n)\right]^{k/2} T_4\right) 
\end{align}
where we let
\begin{align}
T_1 &= \frac{16 a_1 \sqrt{p_i}  S_{i}(\lambda p_i) A_{i}(\lambdabar)^2}{n}\nonumber\\
T_2 & = \left(\frac{16 a_1^2 (\lambda - \lambdabar)^2}{\lambda}\, \frac{ p_i S_i(\lambda p_i)^2 \norm{f_{i,\lambdabar}}{\H}^2}{n} + \frac{8 \lambda p_i  \norm{f_{i,\lambdabar}}{\H}^2}{n} \right)\nonumber\\
T_3 & = \left(2\norm{f_{i,\lambdabar}}{\H}^2 + \frac{\sigma^2}{\lambda}\right)(\lambda_1^i + \lambda p_i)\nonumber\\
T_4 & = \frac{(\lambda_1^i + \lambda p_i) A_i(\lambdabar)^2}{\lambda \sqrt{p_i}}
\end{align}	
\end{proof}

\subsection{Variance Bound}
In this section we provide a proof of Lemma \ref{lem:mainvarbnd}. First, we restate the lemma below.
\begin{lemma*}
 Consider any $d\in \mathbb{N}, d\geq 1$, and $k\geq 2$. Suppose Assumption \ref{asmp:eigfn} holds for this $k$ (with constant $a_1$), and Assumption \ref{asmp:approx} holds. Also, suppose $\forall i\in [m]$, $p_i$ satisfies: $p_i =\Omega\left(\log n/ n\right)$. Then we have
 \begin{align}
\E{\Var_i(\lambda,D)}&\leq \frac{4(\sigma^2 + a_1 \sqrt{p_i}A_i(\lambdabar)^2) S_i(\lambda p_i)}{n} + 4\frac{(\lambdabar - \lambda)^2 p_i}{\lambda}\norm{f_{i,\lambdabar}}{\H}^2  \nonumber\\
&\qquad+  2^{k+2} \left[\CovErr_{i}(d,\lambda p_i, n)\right]^k W_1 + 2^{\frac{k}{2} + 4} \left[\CovErr_{i}(d,\lambda p_i, n)\right]^{k/2} W_2
\end{align}
where we let
\begin{align}
W_1 &= \lambda_{1}^i \left(\norm{f_{i,\overline{\lambda}}}{\H}^2 + \frac{\sigma^2}{2\lambda}\right) \nonumber\\
W_2 &= \lambda_{1}^i \frac{A_{i}(\lambdabar)^2}{\lambda \sqrt{p_i}}
\end{align}
\end{lemma*}
\subsubsection{Proof of Lemma \ref{lem:mainvarbnd}}
\begin{proof}
We want to bound the quantity $\E{\Var_i(\lambda,D)}$, where
\begin{align}
\Var_i(\lambda, D) &= \E{(\fbar_{i,\lambda}(x) - \fhat_{i,\lambda}(x))^2\indic{x\in C_i}}\\
& = \norm{\fbar_{i,\lambda} - \fhat_{i,\lambda}}{\Sigma_i}^2
\end{align}

Since $\fbar_{i,\lambda} = \E{\fhat_{i,\lambda}}$ minimizes $\E{\norm{\fhat_{i,\lambda} - f}{\Sigma_i}^2}$ for $f\in \H$, we can get:
\begin{align}
 \E{Var_i(\lambda,D)} = \E{\norm{\fbar_{i,\lambda} - \fhat_{i,\lambda}}{\Sigma_i}^2} \leq \E{\norm{f_{i,\lambdabar} - \fhat_{i,\lambda}}{\Sigma_i}^2}
\end{align}
where $f_{i,\lambdabar}$ is the solution of \eqref{eq:fbestopt1}. Let $\Delta_v = \fhat_{i,\lambda} - f_{i,\lambdabar}$.

Now, from first order optimality conditions for Eq \eqref{eq:fhatopt}, we have
\begin{align}
 (\Sigmahat_i + \lambda p_i I)\fhat_{i,\lambda} &= \Eemp{y\phi_x \indic{x\in C_i}}\\
 &= \Eemp{f^*(x)\phi_x \indic{x\in C_i}} + \Eemp{\eta\phi_x \indic{x\in C_i}}
\end{align}
Subtracting $(\Sigmahat + \lambda p_i I)f_{i,\lambdabar}$ from the above, we get,
\begin{align}
(\Sigmahat_i + \lambda p_i I)\Delta_v &= \Eemp{(f^*(x) - f_{i,\lambdabar}(x))\phi_x \indic{x\in C_i} - \lambda p_i f_{i,\lambdabar}} + \Eemp{\eta\phi_x \indic{x\in C_i}}\\
&= \Eemp{(f^*(x) - f_{i,\lambdabar}(x))\phi_x \indic{x\in C_i} - \lambdabar p_i f_{i,\lambdabar}} + \Eemp{\eta\phi_x \indic{x\in C_i}} + (\lambdabar - \lambda)p_i f_{i,\lambdabar}
\end{align}
Thus,
\begin{align}
\left(\Sigma_{i,\lambda p_i}^{-1/2}(\Sigmahat_i - \Sigma_i)\Sigma_{i,\lambda p_i}^{-1/2} + I \right)\Sigma_{i,\lambda p_i}^{1/2} \Delta_v &= \Eemp{(f^*(x) - f_{i,\lambdabar}(x))\Sigma_{i,\lambda p_i}^{-1/2} \phi_x \indic{x\in C_i} - \lambdabar p_i \Sigma_{i,\lambda p_i}^{-1/2}f_{i,\lambdabar}}\nonumber\\ 
&\qquad + \Eemp{\eta \Sigma_{i,\lambda p_i}^{-1/2}\phi _x \indic{x\in C_i}} + (\lambdabar - \lambda)p_i \Sigma_{i,\lambda p_i}^{-1/2} f_{i,\lambdabar}
\end{align}
Let us define the event $\mathcal{E}_{cov} = \left\{\norm{\Sigma_{i,\lambda p_i}^{-1/2}(\Sigmahat_i - \Sigma_i)\Sigma_{i,\lambda p_i}^{-1/2}}{}\leq 1/2\right\}$. Note that from Corollary \ref{corr:covcorr}, we have $\prob\left(\mathcal{E}_{cov}^c\right)\leq 2^k \left[\CovErr_{i}(d,\lambda p_i, n)\right]^k$. 
Now, under the event $\mathcal{E}_{cov}$,
\begin{align}
 \E{\norm{\Sigma_{i,\lambda p_i}^{1/2} \Delta_v}{\H}^2} &\leq 4\E{\norm{\Eemp{(f^*(x) - f_{i,\lambdabar}(x))\Sigma_{i,\lambda p_i}^{-1/2} \phi_x \indic{x\in C_i} - \lambdabar p_i \Sigma_{i,\lambda p_i}^{-1/2}f_{i,\lambdabar}}}{\H}^2}\nonumber\\
 &\qquad + 4\E{\norm{\Eemp{\eta \Sigma_{i,\lambda p_i}^{-1/2}\phi _x \indic{x\in C_i}}}{\H}^2} + 4(\lambdabar - \lambda)^2 p_i^2 \norm{\Sigma_{i,\lambda p_i}^{-1/2} f_{i,\lambdabar}}{\H}^2 \nonumber\\
 \end{align}
Now, we can control each of the component terms in the above inequality as follows:
\begin{align}
& 4\E{\norm{\Eemp{(f^*(x) - f_{i,\lambdabar}(x))\Sigma_{i,\lambda p_i}^{-1/2} \phi_x \indic{x\in C_i} - \lambdabar p_i \Sigma_{i,\lambda p_i}^{-1/2}f_{i,\lambdabar}}}{\H}^2}\nonumber\\
&\overset{(a)}{=}\frac{4}{n}\E{ (f^*(x) - f_{i,\lambdabar}(x))^2\norm{\Sigma_{i,\lambda p_i}^{-1/2} \phi_x}{\H}^2 \indic{x\in C_i}} - \frac{4}{n}\lambdabar^2 p_i^2 \norm{\Sigma_{i,\lambda p_i}^{-1/2}f_{i,\lambdabar}}{\H}^2\nonumber\\
&\overset{(b)}{\leq} \frac{4}{n}\sqrt{\E{(f^*(x) - f_{i,\lambdabar}(x))^4\indic{x\in C_i}}}\sqrt{\E{\norm{\Sigma_{i,\lambda p_i}^{-1/2} \phi_x}{\H}^4 \indic{x\in C_i}}}\nonumber\\
&\overset{(c)}{\leq} \frac{4}{n}a_1 \sqrt{p_i}A_i(\lambdabar)^2 S_i(\lambda p_i)
\end{align}
where we have $(a)$ using independence of $\{x_1, \ldots, x_n\}$ and $\E{(f^*(x) - f_{i,\lambdabar}(x)) \phi_x \indic{x\in C_i} - \lambdabar p_i f_{i,\lambdabar}} = 0$ (via first order optimality conditions for $f_{i,\lambdabar}$) , $(b)$using Cauchy-Schwarz and ignoring the negative quantity, $(c)$ using Assumption \ref{asmp:approx} and $\E{\norm{\Sigma_{i,\lambda p_i}^{-1/2} \phi_x}{\H}^4 \indic{x\in C_i}}\leq a_1^2 S_i(\lambda p_i)^2$ (via Assumption \ref{asmp:eigfn}),

And,
\begin{align}
 4\E{\norm{\Eemp{\eta \Sigma_{i,\lambda p_i}^{-1/2}\phi _x \indic{x\in C_i}}}{\H}^2} &= 4\E{\frac{1}{n^2}\sum_{j=1}^n\sum_{k=1}^n \eta_j \eta_k \inpdt{\Sigma_{i,\lambda p_i}^{-1/2}\phi_{x_j}\indic{x_j\in C_i}}{\Sigma_{i,\lambda p_i}^{-1/2}\phi_{x_k}\indic{x_k\in C_i}}_{\H}}\nonumber\\
& \overset{(a)}{=}  4\E{\frac{1}{n^2}\sum_{j=1}^n \eta_j^2 \inpdt{\Sigma_{i,\lambda p_i}^{-1/2}\phi_{x_j}}{\Sigma_{i,\lambda p_i}^{-1/2}\phi_{x_j}}_{\H}\indic{x_j\in C_i}}\nonumber\\
 &\overset{(b)}{\leq} \frac{4\sigma^2 S_i(\lambda p_i)}{n}
\end{align}
where we have $(a)$ since $\E{\eta_j \eta_k}=0$ for $j\ne k$, $(b)$ using $\E{\eta_j^2}\leq \sigma^2$, $\E{\norm{\Sigma_{i,\lambda p_i}^{-1/2}\phi_{x_j}}{\H}^2} = S_i(\lambda p_i)$ and the independence of $\eta_j$ and $x_j$,

And,
\begin{align}
4(\lambdabar - \lambda)^2 p_i^2 \norm{\Sigma_{i,\lambda p_i}^{-1/2} f_{i,\lambdabar}}{\H}^2 &\leq 4(\lambdabar - \lambda)^2 p_i^2\frac{\norm{f_{i,\lambdabar}}{\H}^2}{\lambda_1^i + \lambda p_i}\nonumber\\
&\leq 4\frac{(\lambdabar - \lambda)^2 p_i}{\lambda}\norm{f_{i,\lambdabar}}{\H}^2
\end{align}

Thus, overall, we have
\begin{align}\label{eq:overvar1}
 \E{\norm{\Sigma_i^{1/2} \Delta_v}{\H}^2} &= \E{\norm{\Sigma_i^{1/2} \Delta_v}{\H}^2\indic{\mathcal{E}_{cov}}} + \E{\norm{\Sigma_i^{1/2} \Delta_v}{\H}^2\indic{\mathcal{E}_{cov}^c}}\nonumber\\
 & \leq \E{\norm{\Sigma_{i,\lambda p_i}^{1/2} \Delta_v}{\H}^2 \indic{\mathcal{E}_{cov}}} + \E{\norm{\Sigma_i^{1/2} \Delta_v}{\H}^2\indic{\mathcal{E}_{cov}^c}}\nonumber\\
 & \leq \frac{4(\sigma^2 + a_1 \sqrt{p_i}A_i(\lambdabar)^2) S_i(\lambda p_i)}{n} + 4\frac{(\lambdabar - \lambda)^2 p_i}{\lambda}\norm{f_{i,\lambdabar}}{\H}^2 + \E{\norm{\Sigma_i^{1/2} \Delta_v}{\H}^2\indic{\mathcal{E}_{cov}^c}}\nonumber\\
 & \leq \frac{4(\sigma^2 + a_1 \sqrt{p_i}A_i(\lambdabar)^2) S_i(\lambda p_i)}{n} + 4\frac{(\lambdabar - \lambda)^2 p_i}{\lambda}\norm{f_{i,\lambdabar}}{\H}^2 + \lambda_{1}^i\E{\E{\norm{\Delta_v}{\H}^2\,\vert\, x_1\ldots x_n}\indic{\mathcal{E}_{cov}^c}}
\end{align}
where in the last step, we use the fact that $\mathcal{E}_{cov}$ only depends on $\{x_1,\ldots x_n\}$.

Now, we have the following bound on $\E{\norm{\Delta_v}{\H}^2\,\vert\, x_1\ldots x_n}$.
\begin{align}
 \E{\norm{\Delta_v}{\H}^2\,\vert\, x_1\ldots x_n} &= \E{\norm{\fhat_{i,\lambda} - f_{i,\lambdabar}}{\H}^2\,\vert\, x_1\ldots x_n}\nonumber\\
 & \leq 2\E{\norm{\fhat_{i,\lambda}}{\H}^2\,\vert\, x_1\ldots x_n} + 2\norm{f_{i,\lambdabar}}{\H}^2\nonumber\\
 &\overset{(a)}{\leq} 4\norm{f_{i,\overline{\lambda}}}{\H}^2 + 2\frac{\sigma^2}{\lambda} + \frac{2}{\lambda}\frac{1}{n_i} \sum_{j=1}^n (f^*(x_j) - f_{i,\overline{\lambda}}(x_j))^2\indic{x_j\in C_i}
\end{align}
where we have $(a)$ using the optimality of $\fhat_{i,\lambda}$ in Eq. \eqref{eq:partestopt}

Plugging the above back into Eq. \eqref{eq:overvar1}, we get
\begin{align}
\E{\Var_i(\lambda,D)} &\leq \frac{4(\sigma^2 + a_1 \sqrt{p_i}A_i(\lambdabar)^2) S_i(\lambda p_i)}{n} + 4\frac{(\lambdabar - \lambda)^2 p_i}{\lambda}\norm{f_{i,\lambdabar}}{\H}^2 + 4\lambda_{1}^i \prob(\mathcal{E}_{cov}^c)\left(\norm{f_{i,\overline{\lambda}}}{\H}^2 + \frac{\sigma^2}{2\lambda}\right)\nonumber\\
&\qquad + 4\frac{\lambda_{1}^i}{\lambda} \E{\frac{1}{n_i} \sum_{j=1}^n (f^*(x_j) - f_{i,\overline{\lambda}}(x_j))^2\indic{x_j\in C_i} \indic{\mathcal{E}_{cov}^c}}\nonumber\\
&\overset{(a)}{\leq} \frac{4(\sigma^2 + a_1 \sqrt{p_i}A_i(\lambdabar)^2) S_i(\lambda p_i)}{n} + 4\frac{(\lambdabar - \lambda)^2 p_i}{\lambda}\norm{f_{i,\lambdabar}}{\H}^2 + 4\lambda_{1}^i \prob(\mathcal{E}_{cov}^c)\left(\norm{f_{i,\overline{\lambda}}}{\H}^2 + \frac{\sigma^2}{2\lambda}\right)\nonumber\\
&\qquad + 16\frac{\lambda_{1}^i}{\lambda\sqrt{p_i}}\sqrt{\prob(\mathcal{E}_{cov}^c)}A_i(\lambdabar)^2\nonumber\\
&\leq\frac{4(\sigma^2 + a_1 \sqrt{p_i}A_i(\lambdabar)^2) S_i(\lambda p_i)}{n} + 4\frac{(\lambdabar - \lambda)^2 p_i}{\lambda}\norm{f_{i,\lambdabar}}{\H}^2  \nonumber\\
&\qquad+  2^{k+2}\lambda_{1}^i \left[\CovErr_{i}(d,\lambda p_i, n)\right]^k\left(\norm{f_{i,\overline{\lambda}}}{\H}^2 + \frac{\sigma^2}{2\lambda}\right) + 2^{\frac{k}{2} + 4}\lambda_{1}^i \left[\CovErr_{i}(d,\lambda p_i, n)\right]^{k/2} \frac{A_{i}(\lambdabar)^2}{\lambda \sqrt{p_i}}
\end{align}
where we have $(a)$ using the same sequence of inequalities employed in Eq. \eqref{eq:biasbinseq}.
\end{proof}

\subsection{Proof of Lemma \ref{lem:expcovbnd}}\label{sec:covbndproof}
\begin{proof}
Using the triangle inequality, we obtain the decomposition
\begin{align}
 \E{\norm{\Sigma_{i,\lambda}^{-1/2}(\Sigmahat_i - \Sigma_i)\Sigma_{i,\lambda}^{-1/2}}{}^k}^{1/k} &\leq \underbrace{\E{\norm{\Sigma_{i,\lambda}^{-1/2}(\Sigmahat_i - \Sigmahat_i^d)\Sigma_{i,\lambda}^{-1/2}}{}^k}^{1/k}}_{T_1} \\\nonumber
 & + \underbrace{\E{\norm{\Sigma_{i,\lambda}^{-1/2}(\Sigmahat_i^d - \Sigma_i^d)\Sigma_{i,\lambda}^{-1/2}}{}^k}^{1/k}}_{T_2} + \underbrace{\norm{\Sigma_{i,\lambda}^{-1/2}(\Sigma_i^d - \Sigma_i)\Sigma_{i,\lambda}^{-1/2}}{}}_{T_3}
\end{align}

\textbf{Bound on $T_1$}. Consider the term $\norm{\Sigma_{i,\lambda}^{-1/2}(\Sigmahat_i - \Sigmahat_i^d)\Sigma_{i,\lambda}^{-1/2}}{}$. Using the definition of $\Sigmahat_i$ and $\Sigmahat_i^d$ from Eqs. \eqref{eq:sigmahatdef} and \eqref{eq:sigmahatddef}, and then applying the triangle inequality, we have
\begin{align}\label{eq:t1first}
 \norm{\Sigma_{i,\lambda}^{-1/2}(\Sigmahat_i - \Sigmahat_i^d)\Sigma_{i,\lambda}^{-1/2}}{}\leq \frac{1}{n}\sum_{j=1}^n \norm{\Sigma_{i,\lambda}^{-1/2}((\phi_{x_j}\otimes \phi_{x_j}) - (P_d\phi_{x_j}\otimes P_d\phi_{x_j}))\Sigma_{i,\lambda}^{-1/2}}{}\indic{x_j\in C_i}
\end{align}
Now, recall that for any $x \in \X$, we let $\Sigma_{i,\lambda}^{-1/2}\phi_x = \phi_x'$ and $P_d^{\perp} = \sum_{j>d} (v_j^i\otimes v_j^i)$. Also, $\phi_x' = P_d \phi_x' + P_d^{\perp} \phi_x'$. Then,
\begin{align}
\norm{\Sigma_{i,\lambda}^{-1/2}((\phi_{x}\otimes \phi_{x}) - (P_d\phi_{x}\otimes P_d\phi_{x}))\Sigma_{i,\lambda}^{-1/2}}{} & = \norm{(\phi_{x}'\otimes \phi_{x}') - (P_d\phi_{x}'\otimes P_d\phi_{x}')}{}\nonumber\\
 & = \norm{(P_d^{\perp}\phi_x'\otimes P_d^{\perp}\phi_x') + (P_d^{\perp}\phi_x'\otimes P_d\phi_x') + (P_d\phi_x'\otimes P_d^{\perp}\phi_x')}{}\nonumber\\
 & = \frac{1}{2}\norm{P_d^{\perp}\phi_x'\otimes(P_d^{\perp}\phi_x' + 2P_d\phi_x') + (P_d^{\perp}\phi_x' + 2P_d\phi_x')\otimes P_d^{\perp}\phi_x'}{}\nonumber\\
 & \overset{(a)}{=} \frac{1}{2}\left(\norm{P_d^{\perp}\phi_x'}{\H}^2 + \norm{P_d^{\perp}\phi_x'}{\H}\norm{P_d^{\perp}\phi_x' + 2P_d\phi_x'}{\H}\right)\nonumber\\
 & \overset{(b)}{\leq} \norm{P_d^{\perp}\phi_x'}{\H}^2 + \norm{P_d^{\perp}\phi_x'}{\H}\norm{P_d\phi_x'}{\H}
\end{align}
where we have $(a)$ using $\norm{u\otimes v  + v\otimes u}{} = \left(\inpdt{v}{u}_{\H} + \norm{u}{\H}\norm{v}{\H}\right)$, and $(b)$ using the triangle inequality.\\

Plugging this back into Eq. \eqref{eq:t1first}, we get
\begin{align}
 \norm{\Sigma_{i,\lambda}^{-1/2}(\Sigmahat_i - \Sigmahat_i^d)\Sigma_{i,\lambda}^{-1/2}}{}\leq \frac{1}{n}\sum_{j=1}^n \left(\norm{P_d^{\perp}\phi_{x_j}'}{\H}^2 + \norm{P_d^{\perp}\phi_{x_j}'}{\H}\norm{P_d\phi_{x_j}'}{\H}\right)\indic{x_j\in C_i}
\end{align}
Taking expectation of the $k^{th}$ power on both sides, and using the triangle inequality again, we get
\begin{align}
 \E{\norm{\Sigma_{i,\lambda}^{-1/2}(\Sigmahat_i - \Sigmahat_i^d)\Sigma_{i,\lambda}^{-1/2}}{}^k}^{1/k}& \leq \frac{1}{n}\sum_{j=1}^{n} \E{\norm{P_d^{\perp}\phi_{x_j}'}{\H}^{2k}\indic{x_j\in C_i} }^{1/k} \nonumber\\
 & + \frac{1}{n}\sum_{j=1}^{n} \E{\norm{P_d^{\perp}\phi_{x_j}'}{\H}^k\norm{P_d\phi_{x_j}'}{\H}^k \indic{x_j\in C_i}}^{1/k}\nonumber\\
 &\overset{(a)}{\leq} \frac{1}{n}\sum_{j=1}^{n} \E{\norm{P_d^{\perp}\phi_{x_j}'}{\H}^{2k}\indic{x_j\in C_i} }^{1/k} \nonumber\\
 & + \frac{1}{n}\sum_{j=1}^{n} \E{\norm{P_d^{\perp}\phi_{x_j}'}{\H}^{2k} \indic{x_j\in C_i}}^{1/2k}\E{\norm{P_d\phi_{x_j}'}{\H}^{2k} \indic{x_j\in C_i}}^{1/2k}
\end{align}
where we have $(a)$ using the Cauchy-Schwarz inequality.\\

Now, as a consequence of the reproducing property of kernels, we note that $\phi_x$, for any $x \in \X$, has the representation:
\begin{align}
\phi_x &= \sum_{j} v_j^i(x) v_j^i
\end{align}
Thus,
\begin{align}\label{eq:phixrep}
\phi_x' &= \Sigma_{i,\lambda}^{-1/2}\phi_x = \sum_{j} \frac{v_j^i(x)}{\sqrt{\lambda_j^i + \lambda}} v_j^i\nonumber\\
\Rightarrow\, P_d^{\perp}\phi_x' &= \sum_{j>d} \frac{v_j^i(x)}{\sqrt{\lambda_j^i + \lambda}} v_j^i\nonumber\\
\Rightarrow\, \norm{P_d^{\perp}\phi_x'}{\H}^2 &= \sum_{j>d} \frac{(v_j^i(x))^2}{\lambda_j^i + \lambda} \nonumber\\
\Rightarrow\, \norm{P_d^{\perp}\phi_x'}{\H}^{2k} &= \left(\sum_{j>d} \frac{(v_j^i(x))^2}{\lambda_j^i + \lambda}\right)^k \nonumber\\
& = \left((\sum_{j>d}\lambda_j^i/(\lambda_j^i + \lambda))\sum_{j>d} \frac{\lambda_j^i/(\lambda_j^i + \lambda)}{(\sum_{j>d}\lambda_j^i/(\lambda_j^i + \lambda))}(v_j^i(x))^2/\lambda_j^i\right)^k\nonumber\\
& \overset{(a)}{\leq} \left(\sum_{j>d}\frac{\lambda_j^i}{\lambda_j^i + \lambda}\right)^k\left(\sum_{j>d} \frac{\lambda_j^i/(\lambda_j^i + \lambda)}{(\sum_{j>d}\lambda_j^i/(\lambda_j^i + \lambda))}\left(\frac{(v_j^i(x))^2}{\lambda_j^i}\right)^k\right)
\end{align}
where we have $(a)$ using Jensen's inequality.\\

Therefore, using Assumption \ref{asmp:eigfn}, we get
\begin{align}
\E{\norm{P_d^{\perp}\phi_x'}{\H}^{2k}\indic{x\in C_i}}^{1/k} &\leq a_1 \left(\sum_{j>d}\frac{\lambda_j^i}{\lambda_j^i + \lambda}\right)
\end{align}
Similarly, we can obtain
\begin{align}
\E{\norm{P_d\phi_x'}{\H}^{2k}\indic{x\in C_i}}^{1/k} &\leq a_1 \left(\sum_{j=1}^{d}\frac{\lambda_j^i}{\lambda_j^i + \lambda}\right)
\end{align}

Combining these bounds gives
\begin{align}
 \E{\norm{\Sigma_{i,\lambda}^{-1/2}(\Sigmahat_i - \Sigmahat_i^d)\Sigma_{i,\lambda}^{-1/2}}{}^k}^{1/k}&\leq a_1\left(\sum_{j>d}\frac{\lambda_j^i}{\lambda_j^i + \lambda} + \sqrt{\sum_{j>d}\frac{\lambda_j^i}{\lambda_j^i + \lambda}}\sqrt{\sum_{j=1}^{d}\frac{\lambda_j^i}{\lambda_j^i + \lambda}}\right)\\\nonumber
& = a_1\left(L_i(d,\lambda) + \sqrt{L_i(d,\lambda) U_i(d,\lambda)}\right)
\end{align}
where $L_i(d,\lambda) = \sum_{j>d}\frac{\lambda_j^i}{\lambda_j^i + \lambda}$ and $U_i(d,\lambda) = \sum_{j=1}^{d}\frac{\lambda_j^i}{\lambda_j^i + \lambda}$.

\textbf{Bound on $T_2$}. We want to bound the quantity $\E{\norm{\Sigma_{i,\lambda}^{-1/2}(\Sigmahat_i^d - \Sigma_i^d)\Sigma_{i,\lambda}^{-1/2}}{}^k}^{1/k}$. Using the definition of $\Sigmahat_i^d$ from Eq. \eqref{eq:sigmahatddef}, we have
\begin{align}
 \Sigma_{i,\lambda}^{-1/2}(\Sigmahat_i^d - \Sigma_i^d)\Sigma_{i,\lambda}^{-1/2} &= \frac{1}{n}\sum_{j=1}^{n}\left(\Sigma_{i,\lambda}^{-1/2}(P_d \phi_{x_j}\otimes P_d \phi_{x_j}\indic{x_j\in C_i})\Sigma_{i,\lambda}^{-1/2} - \Sigma_{i,\lambda}^{-1/2}\Sigma_i^d\Sigma_{i,\lambda}^{-1/2}\right)\nonumber\\
 &= \frac{1}{n}\sum_{j=1}^{n}\left((P_d \phi_{x_j}'\otimes P_d \phi_{x_j}')\indic{x_j\in C_i} - \Sigma_{i,\lambda}^{-1/2}\Sigma_i^d\Sigma_{i,\lambda}^{-1/2}\right)
\end{align}
where $\phi_x' = \Sigma_{i,\lambda}^{-1/2}\phi_x$, for any $x \in \X$. Now, as seen in Eq. \ref{eq:phixrep}, we have the representation:
\begin{align}
 P_d\phi_{x_j}' &= \sum_{m=1}^{d} \frac{v_m^i(x_j)}{\sqrt{\lambda_m^i + \lambda}} v_m^i\\
 \Rightarrow\; P_d\phi_{x_j}'\otimes P_d\phi_{x_j}' &= \sum_{m=1}^{d}\sum_{n=1}^d \frac{v_m^i(x_j) v_n^i(x_j)}{\sqrt{\lambda_m^i + \lambda}\sqrt{\lambda_n^i + \lambda}} (v_m^i\otimes v_n^i)
\end{align}
Also, using the definition of $\Sigma_i^d$ from Eq. \ref{eq:sigmaddef}, we have the relation:
\begin{align}
 \Sigma_{i,\lambda}^{-1/2}\Sigma_i^d\Sigma_{i,\lambda}^{-1/2} &= \sum_{m=1}^{d} \frac{\lambda_m^i}{\lambda_m^i + \lambda} (v_m^i\otimes v_m^i)
\end{align}
Now, let $A_j\in \real^{d \times d}$ be a matrix such that
\begin{align}
 \text{For }m\ne n,\; A_j(m,n) &= v_m^i(x_j)v_n^i(x_j)\indic{x_j\in C_i}/\sqrt{(\lambda_m^i + \lambda)(\lambda_n^i + \lambda)}\\
 A_j(m,m) &= \left(v_m^i(x_j)^2\indic{x_j\in C_i} - \lambda_m^i\right)/(\lambda_m^i + \lambda)
\end{align}
Also, let $B = \sum_{j=1}^n A_j/n$. Then, 
\begin{align}
 \Sigma_{i,\lambda}^{-1/2}(\Sigmahat_i^d - \Sigma_i^d)\Sigma_{i,\lambda}^{-1/2} &= \frac{1}{n}\sum_{j=1}^n\left(\sum_{m=1}^d\sum_{n=1}^d A_j(m,n) (v_m^i\otimes v_n^i)\right)\\
 & = \sum_{m=1}^{d}\sum_{n=1}^d B(m,n) (v_m^i\otimes v_n^i)
\end{align}
So, we get
\begin{align}\label{eq:t2finmat}
 \norm{\Sigma_{i,\lambda}^{-1/2}(\Sigmahat_i^d - \Sigma_i^d)\Sigma_{i,\lambda}^{-1/2}}{} = \norm{\sum_{m=1}^{d}\sum_{n=1}^d B(m,n) (v_m^i\otimes v_n^i)}{} = \norm{B}{2} = \norm{\frac{1}{n}\sum_{j=1}^{n} A_j}{2}
\end{align}
where $\norm{\cdot}{2}$ corresponds to the usual spectral norm for finite dimensional matrices.

Thus to bound $\E{\norm{\Sigma_{i,\lambda}^{-1/2}(\Sigmahat_i^d - \Sigma_i^d)\Sigma_{i,\lambda}^{-1/2}}{}^k}^{1/k}$, we need to bound $\E{\norm{\frac{1}{n}\sum_{j=1}^n A_j}{2}^k}^{1/k}$. To do this, we can use the following result from \cite{chen12} (similar to its use in \cite{zhang13}) which provides a bound on the moment of the spectral norm of a sum of finite dimensional random matrices.
\begin{lemma}{Theorem A.1 \cite{chen12}}\label{lem:maskcovlem}
Let  $q\geq 2$, and fix $r\geq \max\{q, \log d\}$. Consider a finite sequence $\{Y_i\}$ of independent, symmetric, random, self-adjoint matrices with dimension $d\times d$. Then,
 \begin{align}
  \E{\norm{\sum_i Y_i}{2}^q}^{1/q}\leq \sqrt{er}\norm{\sum_i \E{Y_i^2}}{2}^{1/2} + 2er \,\E{\max_i \norm{Y_i}{2}^q}^{1/q}
 \end{align}
\end{lemma}

We apply Lemma \ref{lem:maskcovlem} in our case with the sequence of matrices $\left\{\frac{A_j}{n}\right\}$ to get
\begin{align}
 \E{\norm{\frac{1}{n}\sum_{j=1}^{n}A_j}{2}^k}^{1/k}\leq \frac{\sqrt{e \log d}}{n}\norm{\sum_{j=1}^n \E{A_j^2}}{2}^{1/2} + \frac{2e\log d}{n} \,\E{\max_j \norm{A_j}{2}^k}^{1/k}
\end{align}

Now, we can bound $\norm{\sum_{j=1}^n \E{A_j^2}}{2}$ as:
\begin{align}
 \norm{\sum_{j=1}^n \E{A_j^2}}{2}\overset{(a)}{\leq} \sum_{j=1}^{n}\norm{\E{A_j^2}}{2}& \overset{(b)}{\leq} \sum_{j=1}^{n}\E{\norm{A_j}{2}^2}\nonumber\\
 &\overset{(c)}{\leq} \sum_{j=1}^{n}\E{\Tr{A_j}^2}\nonumber\\
 & = \sum_{j=1}^{n}\E{\left(\sum_{m=1}^d \frac{v_m^i(x_j)^2\indic{x_j\in C_i}}{\lambda_m^i + \lambda}\right)^2} + \sum_{j=1}^{n}\left(\sum_{m=1}^d \frac{\lambda_m^i}{\lambda_m^i + \lambda}\right)^2 \nonumber\\
 & \qquad - \sum_{j=1}^{n}2 \left(\sum_{m=1}^d \frac{\lambda_m^i}{\lambda_m^i + \lambda}\right) \E{\left(\sum_{m=1}^d \frac{v_m^i(x_j)^2\indic{x_j\in C_i}}{\lambda_m^i + \lambda}\right)}\nonumber\\
 &\overset{(d)}{=} \sum_{j=1}^{n}\E{\left(\sum_{m=1}^d \frac{v_m^i(x_j)^2\indic{x_j\in C_i}}{\lambda_m^i + \lambda}\right)^2} - \sum_{j=1}^{n}\left(\sum_{m=1}^d \frac{\lambda_m^i}{\lambda_m^i + \lambda}\right)^2 \nonumber\\
 &\leq \sum_{j=1}^{n}\E{\left(\sum_{m=1}^d \frac{v_m^i(x_j)^2\indic{x_j\in C_i}}{\lambda_m^i + \lambda}\right)^2}\nonumber\\
 &= \sum_{j=1}^{n}\left(\sum_{m=1}^{d} \lambda_m^i/(\lambda_m^i + \lambda)\right)^2 \E{\left(\sum_{m=1}^d \frac{\lambda_m^i/(\lambda_m^i + \lambda)}{\sum_{m=1}^d \lambda_m^i/(\lambda_m^i + \lambda)}\frac{v_m^i(x_j)^2\indic{x_j\in C_i}}{\lambda_m^i}\right)^2}\nonumber\\
 &\overset{(e)}{\leq} \sum_{j=1}^n U_i(d,\lambda)^2\, \E{\sum_{m=1}^d \frac{\lambda_m^i/(\lambda_m^i + \lambda)}{\sum_{m=1}^d \lambda_m^i/(\lambda_m^i + \lambda)}\left(\frac{v_m^i(x_j)^2\indic{x_j\in C_i}}{\lambda_m^i}\right)^2}\nonumber\\
 &\overset{(f)}{\leq} \sum_{j=1}^n U_i(d,\lambda)^2 a_1^2\nonumber\\
 &= n \,U_i(d,\lambda)^2 a_1^2
\end{align}
where we have $(a)$ using the triangle inequality, $(b)$ using Jensen's inequality, $(c)$ since the spectral norm is upper bounded by the trace, $(d)$ using the fact that $\E{v_m^i(x_j)^2\indic{x_j\in C_i}} = \lambda_m^i$ for any $m$, $(e)$ using Jensen's inequality again, and $(f)$ using Assumption \ref{asmp:eigfn}.

We can also bound $\E{\max_j \norm{A_j}{2}^k}$ as:
\begin{align}
 \E{\max_j \norm{A_j}{2}^k} &\leq \sum_{j=1}^{n} \E{\norm{A_j}{2}^{k}} \nonumber\\
 &\overset{(a)}{\leq} \sum_{j=1}^{n}\E{\left(\sum_{m=1}^{d} \frac{v_m^i(x_j)^2\indic{x_j\in C_i}}{\lambda_m^i + \lambda} + \frac{\lambda_1^i}{\lambda_1^i + \lambda}\right)^{k}}\nonumber\\
 &\overset{(b)}{\leq} \sum_{j=1}^{n}\E{2^{k}\left(\sum_{m=1}^{d} \frac{v_m^i(x_j)^2\indic{x_j\in C_i}}{\lambda_m^i + \lambda}\right)^{k} + 2^{k}\left(\frac{\lambda_1^i}{\lambda_1^i + \lambda}\right)^{k}} \nonumber\\
 &\overset{(c)}{\leq} n\, 2^{k}\left(U_i(d,\lambda)^{k} a_1^{k} +  \left(\frac{\lambda_1^i}{\lambda_1^i + \lambda}\right)^{k}\right)
\end{align}
where we have $(a)$ using the triangle inequality for the spectral norm and the fact that $A_j = v v^T - D$ with $v = \left\{v_m^i(x_j)\indic{x_j\in C_i}/\sqrt{\lambda_m^i + \lambda}\right\}_{m=1}^{d}$ and $D = \text{diag}\left(\{\lambda_m^i/(\lambda_m^i + \lambda)\}_{m=1}^{d}\right)$, $(b)$ using the inequality $(a+b)^k\leq 2^k (a^k + b^k)$, and $(c)$ using Jensen's inequality and Assumption \ref{asmp:eigfn}.

Thus,
\begin{align}
\E{\max_j \norm{A_j}{2}^k}^{1/k} \leq 2 n^{1/k} \left(U_i(d,\lambda) a_1 +  \left(\frac{\lambda_1^i}{\lambda_1^i + \lambda}\right)\right)
\end{align}

Plugging these bounds into Eq. \eqref{eq:t2finmat}, we finally have
\begin{align}
 \E{\norm{\Sigma_{i,\lambda}^{-1/2}(\Sigmahat_i^d - \Sigma_i^d)\Sigma_{i,\lambda}^{-1/2}}{}^k}^{1/k} &= \E{\norm{\frac{1}{n}\sum_{j=1}^{n} A_j}{2}^k}^{1/k}\nonumber\\
 &\leq a_1 \sqrt{\frac{e \log d}{n}}\, U_i(d,\lambda) + \frac{4e \log d}{n^{1-1/k}}\left(a_1 U_i(d,\lambda) + \frac{\lambda_1^i}{\lambda_1^i + \lambda}\right)
\end{align}

\textbf{Bound on $T_3$}. We wish to bound $\norm{\Sigma_{i,\lambda}^{-1/2}(\Sigma_i^d - \Sigma_i)\Sigma_{i,\lambda}^{-1/2}}{}$. Using the definition of $\Sigma_i^d$ from Eq. \eqref{eq:sigmaddef}, we can get
\begin{align}
 \Sigma_{i,\lambda}^{-1/2}(\Sigma_i^d - \Sigma_i)\Sigma_{i,\lambda}^{-1/2} = - \sum_{j>d}\frac{\lambda_j^i}{\lambda_j^i + \lambda}(v_j^i\otimes v_j^i)
\end{align}
Thus,
\begin{align}
\norm{\Sigma_{i,\lambda}^{-1/2}(\Sigma_i^d - \Sigma_i)\Sigma_{i,\lambda}^{-1/2}}{} = \frac{\lambda_{d+1}^i}{\lambda_{d+1}^i + \lambda}
\end{align}

\textbf{Overall Bound}. Combining the bounds on the terms $T_1$, $T_2$ and $T_3$, we get the final bound in the lemma.
\end{proof}

\end{document}